\documentclass[10pt,twocolumn,letterpaper]{article}

\usepackage{cvpr}              

\usepackage{graphicx}
\usepackage{amsmath}
\usepackage{amssymb}
\usepackage{booktabs}
\usepackage{tabularx}
\usepackage{makecell}

\usepackage{algorithm}
\usepackage{algorithmic}
\usepackage{scalerel}

\usepackage{amsfonts}
\usepackage{mathtools}
\usepackage{bbm}
\usepackage{multirow}
\usepackage{dsfont}
\usepackage{wrapfig}
\usepackage{enumitem}
\usepackage{comment}
\usepackage{subcaption}
\usepackage{array}
\usepackage{amsthm}
\usepackage{xcolor} 
\makeatletter
\@namedef{ver@everyshi.sty}{}
\makeatother
\usepackage{tikz}
\usetikzlibrary{fit,calc}
\usepackage{minitoc}

\usepackage[pagebackref,breaklinks,colorlinks]{hyperref}

\usepackage[capitalize]{cleveref}
\crefname{section}{Sec.}{Secs.}
\Crefname{section}{Section}{Sections}
\Crefname{table}{Table}{Tables}
\crefname{table}{Tab.}{Tabs.}

\colorlet{orange}{orange!80}
\colorlet{blue}{cyan!80}

\def\cvprPaperID{7645}
\def\confName{CVPR}
\def\confYear{2023}

\begin{document}
\doparttoc 
\faketableofcontents 

\newcommand{\algname}{{LoGo}}  
\newtheorem{definition}{Definition}
\newtheorem{observation}{Observation}
\newtheorem{remark}{Remark}
\doparttoc 
\faketableofcontents 

\title{Re-thinking Federated Active Learning based on Inter-class Diversity}

\author{
SangMook Kim${}^{1}$\thanks{equal contribution} \qquad Sangmin Bae${}^{1}$\footnotemark[1] \qquad Hwanjun Song${}^{2}$\thanks{corresponding authors} \qquad Se-Young Yun${}^{1}$\footnotemark[2] \vspace{2.5pt}\\
${}^{1}$KAIST AI \qquad ${}^{2}$NAVER AI LAB\\
{\tt\small \{sangmook.kim, bsmn0223, yunseyoung\}@kaist.ac.kr \qquad ghkswns91@gmail.com}
}

\maketitle

\begin{abstract}
\vspace{-2pt}
Although federated learning has made awe-inspiring advances, most studies have assumed that the client's data are fully labeled.
However, in a real-world scenario, every client may have a significant amount of unlabeled instances.
Among the various approaches to utilizing unlabeled data, a federated active learning framework has emerged as a promising solution. 
In the decentralized setting, there are two types of available query selector models, namely `global' and `local-only' models, but little literature discusses their performance dominance and its causes.
In this work, we first demonstrate that the superiority of two selector models depends on the global and local inter-class diversity.
Furthermore, we observe that the global and local-only models are the keys to resolving the imbalance of each side.
Based on our findings, we propose \algname{}, a FAL sampling strategy robust to varying local heterogeneity levels and global imbalance ratio, that integrates both models by two steps of active selection scheme.
\algname{} consistently outperforms six active learning strategies in the total number of 38 experimental settings. The code is available at: \url{https://github.com/raymin0223/LoGo}.
\end{abstract}
\vspace{-2pt}


\vspace{-9pt}
\section{Introduction}

\begin{figure*}[t!]
\vspace{-2pt}
\centering
\begin{subfigure}[b]{0.435\linewidth}
\centering
\includegraphics[width=\linewidth]{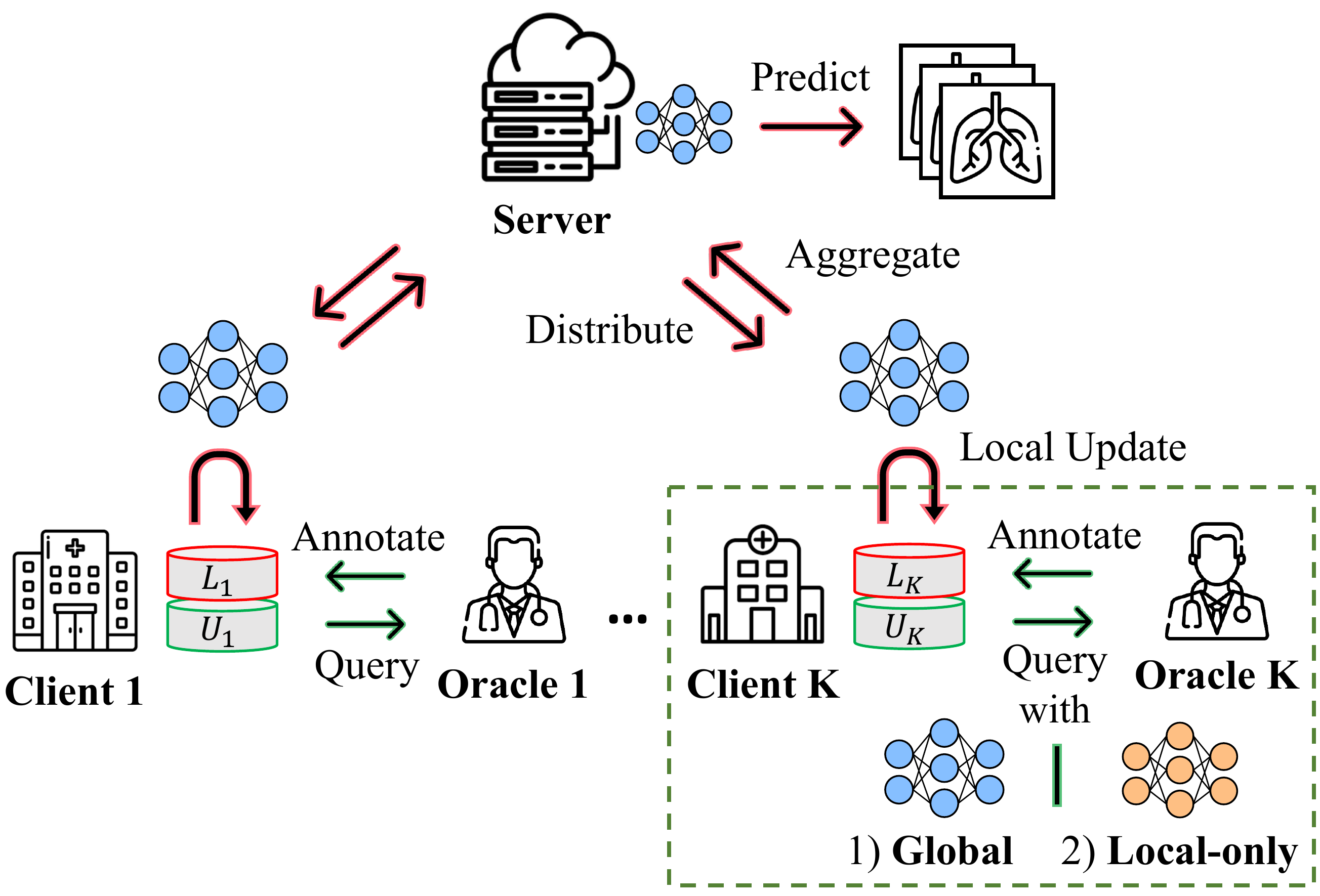}
\caption{\small {Federated Active Learning framework.}}
\label{fig:overview_left}
\end{subfigure}
\hspace{10pt}
\begin{subfigure}[b]{0.435\linewidth}
\centering
\includegraphics[width=\linewidth]{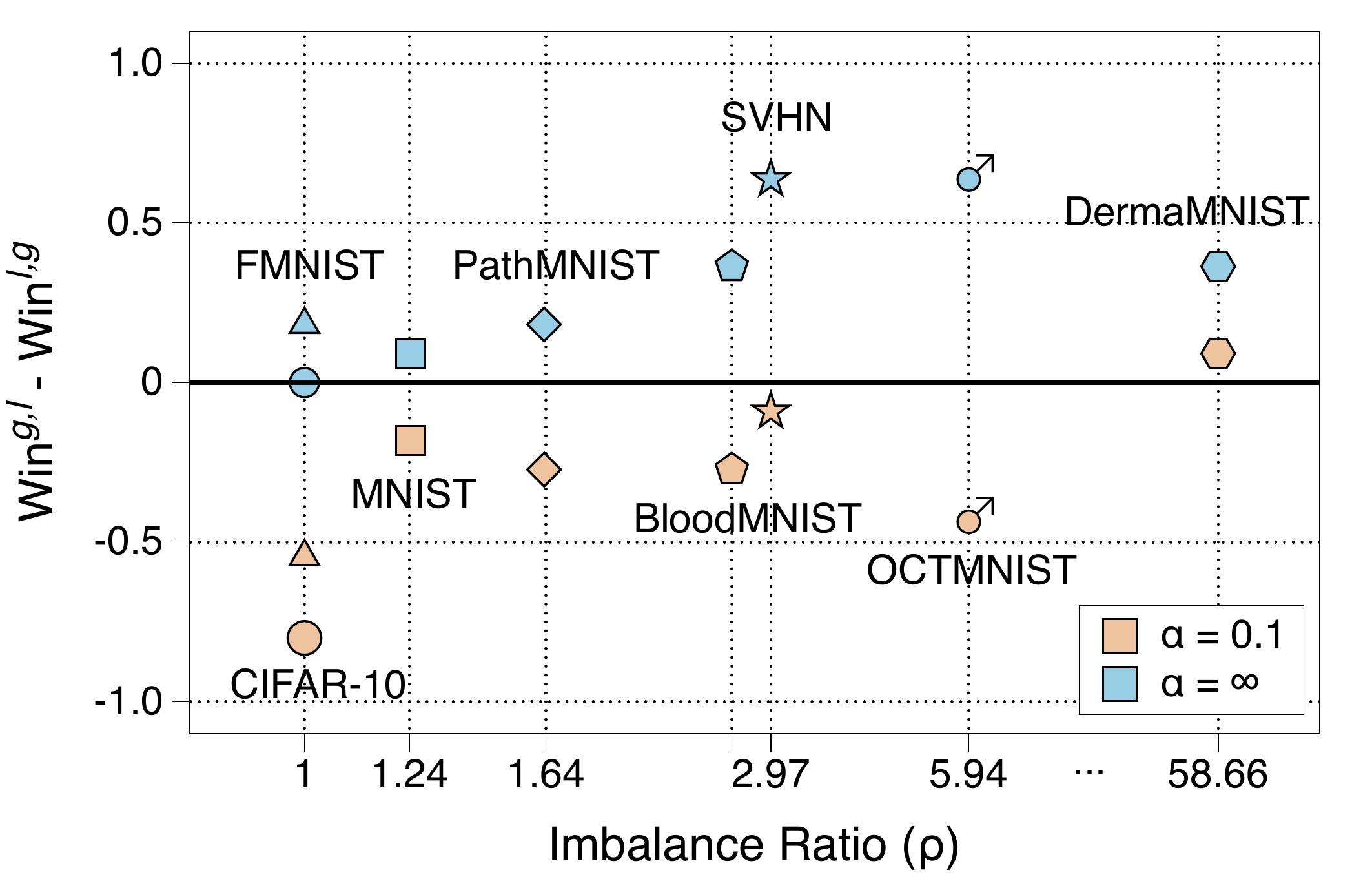}
\caption{\small {Superiority change of query selector.}} 
\label{fig:overview_right}
\end{subfigure}
\vspace*{-0.15cm}
\caption{Motivation: (a) The red and green lines correspond to the conventional FL and AL framework. We focus on a sampling strategy for FAL\,(green box) with the considerations of hierarchy structure and two available query-selecting models. (b) For a fixed querying strategy\,(Entropy sampling), the performance gap occurs only by changing the query selector. The y-axis is the gap in the winning rate in total active rounds\,(refer to Section\,\ref{sec:analysis}). Closer to 1 indicates that global models outperforms than local-only models, and -1 is the opposite. }
\label{fig:overview}
\vspace*{-5pt}
\end{figure*}

Federated learning\,(FL) is a distributed framework that allows multiple parties to learn a unified deep learning model cooperatively with preserving the privacy of the local client\cite{fedavg, fedprox, scaffold}.
Typically, FL has been actively studied in a standard supervised learning setting, where all the training instances are labeled, but it is more realistic for each client to contain both labeled and unlabeled data due to the high labeling cost\cite{fedema, orchestra}. 
Here, active learning\,(AL) can be a promising solution to improve the performance of a cooperated model with the pool of unlabeled data.
In practice, federated active learning\,(FAL) framework has recently attempted to bridge two different philosophies in FL and AL\cite{fal_waste_disaster, f_al}.
As illustrated in Figure\,\ref{fig:overview}-(a) FAL framework alternates an FL procedure\,(red line) of training a predictive model collaboratively through local updates and aggregation phases, and an AL procedure\,(green line) of querying and annotating informative instances separately per client.

Although the overall framework just appears to be a straightforward fusion of two research fields, FL factors introduce two major challenges to the AL procedure.
\textbf{First}, the class imbalance of the local dataset originates from heterogeneous distribution across local clients\cite{fedprox, scaffold, feddc}.
Hence, in the FAL framework, the active selection algorithm has to ensure \emph{inter-class} diversity from both local and global perspectives.
\textbf{Second}, there are two available types of query-selecting models, a \emph{global} model, which is globally optimized through the FL pipeline, and a \emph{local-only} model\cite{fedbabu, fedrep}, which can be separately trained only for each client. 
In the query selection phase, the global model can leverage the aggregated knowledge of all clients, while the local-only model is able to detect the most valuable instances for the local updates.

Prior FAL literature\cite{fal_waste_disaster, f_al}, which simply adapt conventional AL strategies, had little discussion on these challenges.
As our first contribution, we found the significant performance gap between two types of query selector\,(see Figure\,\ref{fig:overview}-(b)), and it is the first study to solve a conundrum of dominance trend
by introducing two indicators of inter-class diversity\footnote{We use the term inter-class diversity interchangeably with class balance throughout the paper.}-- \textit{local heterogeneity level}\,($\alpha$) and \textit{global imbalance ratio}\,($\rho$). 
The first indicator $\alpha$ is the concentration parameter of Dirichlet distribution, commonly seen in the FL literature\cite{scaffold, feddyn, feddc}, resulting in more locally imbalanced class distribution at lower values.
Besides, $\rho$ indicator is the ratio of class imbalance for the aggregated global data of all clients\cite{buda2018systematic, johnson2019survey}.
We discovered three meaningful insights on selector dominance:
\textbf{(Obs.\,\ref{obs:obs1})} Interestingly, the superiority of the two selectors varies depending on the two indicators of inter-class diversity, $\alpha$ and $\rho$.
\textbf{(Obs.\,\ref{obs:obs2})} When local heterogeneity is severe\,($\alpha$ is low), a local-only model is preferred for weighing minority instances for each client, and \textbf{(Obs.\,\ref{obs:obs3})} when globally minor classes exist\,($\rho$ is high), the knowledge of the entire data distribution, inherent in a global model, is more essential.\,\,
$\vartriangleright$ See Section\,\ref{sec:analysis}

In a real FAL scenario, the superiority of two query models for a given dataset cannot be known in advance due to privacy preservation. Therefore, as our second contribution, we design a simple yet effective FAL querying approach, \textbf{\algname{}}, that simultaneously leverages local-only and global models, to be robust to varying heterogeneity levels and global imbalance ratios.
\algname{} is a clutering-based sampling strategy, which is composed of {macro} and {micro} step exploiting local-only and global models, respectively. 
The rationale behind our method is that the optimal querying policy needs to evaluate the informativeness of instances with \emph{both} models, which implicitly learn local and global data distribution, respectively.
In a \emph{macro step}, to improve the local inter-class diversity first, we perform $k$-means clustering\cite{k_means} in hallucinated gradient space generated from local-only models. 
Then, in a \emph{micro step}, the final query set is determined via one step of the EM algorithm\cite{dempster1977maximum}, making cluster boundaries using instances from macro step\,(E-step) and cluster-wise sampling with the global model\,(M-step).
The proposed cluster-wise sampling conservatively guarantees the diversity information of the macro step, \ie, the local inter-class diversity obtained by local-only models, while also considering the global minority classes via the global model.\,\,
$\vartriangleright$ See Section\,\ref{subsec:lg_fal}

As our third contribution, we conduct \emph{a total number of 38 experiments} on five datasets using seven AL strategies including our \algname{} algorithm.
To verify the superiority of our method in real-world scenarios, we build comprehensive combinations of six categories, including query selector types (local-only vs. global models), local heterogeneity levels ($\alpha \in [0.1, \infty)$), global imbalance ratios ($\rho \in [1, 58])$, model architectures, budget sizes, and model initialization scheme. As a result, the experimental results empirically prove our three observations\,(Obs.\,\ref{obs:obs1}--\ref{obs:obs3}). Besides, our method outperforms all other AL baselines and na\"ive implementations for an ensemble of two query selectors in extensive experimental settings.\,\,
$\vartriangleright$ See Section\,\ref{sec:evaluation}
\section{Related Work}

\noindent\textbf{Federated Learning} is a collaborative learning framework with multiple clients while maintaining the privacy of each client dataset.
In this decentralized framework, FedAvg\cite{fedavg} is considered a de facto algorithm in which a server and clients communicate only model parameters efficiently.
One of the main bottlenecks for FL is the statistical heterogeneity problem across the clients' dataset, as the weight divergence from heterogeneous distributions hinders convergence during the aggregation scheme.
Hence, several algorithms have tackled the local heterogeneity through the alignment between the local-updated gradient and the aggregated gradient, as a form of correction term\cite{scaffold, feddc}, regularization loss\cite{fedprox, feddyn}, or distillation-based loss\cite{lee2021preservation, li2021model}.
Furthermore, there are existing literature to alleviate the class imbalance on the local side and they eventually achieve the global balance via rebalancing datasets~\cite{astraea} or a weighted loss function with a monitoring scheme~\cite{wang2021addressing}.

\smallskip\smallskip
\noindent\textbf{Active Learning} minimizes the labeling effort by querying the most informative instances from the unlabeled data pool. 
There are three major types of active learning strategies, namely uncertainty-based sampling, diversity-based sampling, and hybrid strategy.
Uncertainty-based sampling queries the most uncertain instances that lie on the current decision boundary\cite{margin_sampling, confidence_sampling, deep_bayesian_al, deepfool}, while diversity-based sampling selects a set of unlabeled instances that represents the entire unlabeled data distribution\cite{coreset, ff_active}. 
Recently, hybrid strategies simultaneously consider both uncertainty and diversity.
BADGE\cite{badge} utilized the gradient embedding as the uncertainty measure and selected the diverse query set by a \textit{k}-means++ initialization scheme\cite{kmeans}.
Several hybrid AL methods are based on a submodular data subset selection\cite{fass}, pairwise contextual diversity\cite{cdal}, or feature mixing\cite{alfa_mix}.
In addition to the three categories, model-based strategies have also been recently proposed to train additional networks for query selection, such as a VAE and discriminator\cite{vaal} or a sequential graph neural network\cite{gcnal}. 

\smallskip\smallskip
\noindent\textbf{Federated Active Learning} has been recently studied to address a more realistic scenario, where clients have lots of the unlabeled instances\cite{fedema, fedu, fedmatch}. 
However, previous works\cite{fal_waste_disaster, f_al} have not deeply discussed the challenges of FAL framework and na\"ively applied existing AL strategies. Ahn \etal\cite{f_al} have even considered the local-only query selector, but concluded that the global model is superior to the local-only model with limited experimental settings, three benchmarks and one heterogeneity level.
In this work, we observe the counterparts in varying benchmarks and heterogeneity levels, which motivates us to design a novel sampling strategy by exploiting both global and local-only models. Our extensive analysis and experimental results encourage future research on the FAL problem.

\section{Preliminary}
\label{sec:notation}

\begin{table}[!t]
    \small
    \centering
    \resizebox{\linewidth}{!}{
    \begin{tabular}{ll}
        \toprule
        {\textbf{Indices}: }  \\
        $c$ & Index for a class ($c \in \{ 1, \dots, C \}$) \\
        $r$ & Index for AL round ($r \in \{ 1, \dots, R\} = [R]$) \\
        $k$ & Index for a client ($k \in \{ 1, \dots, K\} = [K]$) \\
        \midrule
        {\textbf{Parameters}:} \\
        $B$ & Labeling budget for each AL round $r$ \\
        $\alpha$ & Local heterogeneity level\\
        $\rho$  & Global imbalance ratio\\ 
        \midrule
        {\textbf{Data}:} \\
        $U_k^r$ & Pool of unlabeled instances for a client $k$ at round $r$ \\
        $L_k^r$ & A queried instance set from $U_k^r$ at round $r$ \\
        $D_k^r$ & An available labeled set at round $r$ \\ 
        \midrule
        {\textbf{Weights}:} \\
        $\Theta^{r*}$ & Aggregated weights via FL phases on $D^r$  (\textit{global} model)  \\
        $\Theta_{k*}^{r}$ & Separately optimized weights on $D_k^r$ (\textit{local-only} model)  \\
        \bottomrule
    \end{tabular}}
    \vspace*{-0.2cm}
    \caption{Summary of notations throughout the paper.}
    \label{tab:notation}
    \vspace*{-5pt}
\end{table}

\smallskip
\noindent\textbf{AL Procedure.}
For the ease of understanding, we summarize notations in Table\,\ref{tab:notation}.
At the first AL round\,(\ie, $r=1$), each client $k$ randomly selects $B$ instances, ${L}_k^1 = \{x_1, \dots, x_B\}$, from ${U}_k^1$, and oracles annotate them to obtain the initial labeled set ${D}_k^1 = \{(x_1, y_1), \dots, (x_B, y_B)\}$. 
For the next round ($r\geq 2$), based on the given querying strategy $\mathcal{A}(\cdot)$ and the model parameters $\Theta$, the query set of the $k$-th client at round $r$ is sampled by
\begin{equation}
{L}_k^{r} = \mathcal{A}({U}_k^{r}, \Theta, B), ~~{\rm where}~~ {U}_k^{r} = {U}_k^{r-1} \setminus {L}_k^{r-1}.
\label{eq:al_general_form}
\end{equation}

The querying function $\mathcal{A}(\cdot)$ in Eq.\,\eqref{eq:al_general_form} depends on which AL algorithm is used. 
For example, Entropy sampling\cite{confidence_sampling} queries the instances with the highest uncertainty like:
\begin{equation} 
\mathcal{A}({U}, \Theta, B) = \underset{ x_i \in {L},\,|{L}\vert=B,\,{L} \subseteq {U}}{\arg \max} H(p(y | x_i; \Theta))\\
\label{eq:querying}
\end{equation}
where $H(p)\!=\!-\!\sum_{c=1}^C p_c \, \ln p_c$, and $p$ is the predictive probability.
The query set is annotated by the oracle and assembled to expand the available labeled set, \ie, ${D}_k^{r} = {D}_k^{r-1} \cup \{ (x_i, y_i)\,|\,x_i\in{L}_k^{r} \}$.

\noindent\\
\textbf{FL Procedure.}
After each AL round, we perform the FL procedure of which the objective is to obtain the optimal parameter $\Theta^{r*}$ such that it minimizes the target loss on the given labeled set for all clients, $D^{r} = \cup_{k=1}^{K} D_{k}^{r}$,
\begin{equation}
    \Theta^{r*} = \underset{\Theta}{\arg\min} \, f(\Theta^r) = \underset{\Theta}{\arg\min}\,\frac{1}{|D^{r}|} \sum_{i=1}^{|D^{r}|} f_i(\Theta^r)
    \label{eq:fl_global_loss}
\end{equation}
where $f_i(\Theta) = \ell(x_i, y_i; \Theta)$ and $\ell(\cdot)$ is the loss function determined by the network parameter $\Theta$.
However, due to data privacy, the global model is optimized based on the reformulated update rule on the partitioned data over clients:
\begin{gather}
    f(\Theta^r) = \sum_{k=1}^K \frac{|D_k^{r}|}{|D^{r}|} \, F(\Theta_k^r), \nonumber \\
    \text{ where } F(\Theta_k^r) = \frac{1}{|D_k^{r}|} \sum_{(x_i, y_i) \in D_k^{r}} \ell(x_i, y_i; \Theta_k^r).
    \label{eq:reform_fl_loss}
\end{gather}
The model $\Theta_k^r$ is updated locally on the client side for its local data $D_k^{r}$, and then it is aggregated globally to generate a global model $\Theta^r$. The local update and model aggregation steps are alternated until the global model converges; this corresponds to the most popular FL training pipeline, FedAvg proposed by \cite{fedavg}.

The previous studies\cite{fal_waste_disaster, f_al} have typically used the converged global model $\Theta^{r*}$ in the next AL round as the query selector of Eq.\,\eqref{eq:querying}.
However, considering the hierarchy structure in the FAL framework, it is also possible to use a separately optimized model on local partitioned data; replacing $D^r$ with $D_k^r$ in Eq.\,\eqref{eq:fl_global_loss}. It is often referred to as the local-only model\cite{fedbabu, fedrep}, and we denote it as $\Theta_{k}^{r*}$. 
In the following section, we investigate what these models are specialized in and when they are beneficial to use.

\section{Observation and Analysis}
\label{sec:analysis}

In this section, we analyze the performance trend between global and local-only models as the query selector, with respect to the degree of class imbalance in local and global data distribution.
We synthetically adjust two indicators of inter-class diversity, $\alpha \in \{0.1, 1.0, \infty\}$ and $\rho \in \{1, 5, 10, 20\}$, on CIFAR-10 benchmark.
As $\alpha$ and $\rho$ get lower and higher, the levels of local heterogeneity and global imbalance increase, respectively (refer to Appendix\,\ref{sec:dataset_summary} for the detailed data distribution).
For both query selectors, we use Entropy sampling\cite{confidence_sampling} as an active learning algorithm, and the training set is progressively labeled with the query ratio of 10\% in each AL round.

\noindent\\
{\textbf{Comparison Metric.}}
We evaluate the superiority over AL rounds through pairwise comparison\cite{badge, alfa_mix}, widely used in conventional AL literature.
We repeat each experimental setup, a pair of $\alpha$ and $\rho$, with four different seeds and obtain a set of four accuracy results $a_{r}=\{a_{r,1},..., a_{r,4}\}$ at each round $r$. Then, we conduct a two-sided t-test, where $t$-score is defined by Definition\,\ref{def:t_score} for two given strategies $i$ and $j$. Note that the strategy denotes the combination of the sampling strategy and the type of query selector. 

\begin{definition}\!\!\!{\normalfont\cite{semenick1990tests}}\, Let $a_{r}^{i}$ and $a_{r}^{j}$ be the set of accuracies for two different FAL strategies $i$ and $j$. Then, $t$-score at AL round $r$ is formulated as:
\begin{equation}
\begin{gathered}
t_{r}^{ij} = \frac{\sqrt{4} \mu^{ij}_{r}}{\sigma^{ij}_{r}},\,\,\, \text{\normalfont where } \mu^{ij}_{r} \!= \!\frac{1}{4}\sum_{l=1}^{4} \big(a_{r,l}^{i} - a_{r,l}^{j}\big) \!\!\! \\ \text{\normalfont and}\,\,\, \sigma^{ij}_{r}=\sqrt{\frac{1}{3}\sum_{l=1}^{4} \Big(\big(a_{r,l}^{i} - a_{r,l}^{j}\big) - \mu^{ij}_{r}\Big)}.
\label{eq:t_score}
\end{gathered}
\end{equation}
\label{def:t_score}
\end{definition}
\vspace*{-0.4cm}
\noindent Here, the strategy $i$ is considered to beat the strategy $j$ if $t_r^{ij} >$ 2.776. Therefore, the \textit{winning rate} for all AL rounds is formulated as follows:
\begin{equation}
{\sf win}^{ij} =  \sum_{r=1}^{R} \frac{1}{R} \mathds{1}_{t_{r}^{ij} > \text{2.776}}.
\label{eq:cell_value}
\end{equation}
The value of winning rate becomes 1 if the strategy $i$ beats the strategy $j$ over all AL rounds.

\begin{figure}[t!]
\centering
\includegraphics[width=0.92\linewidth]{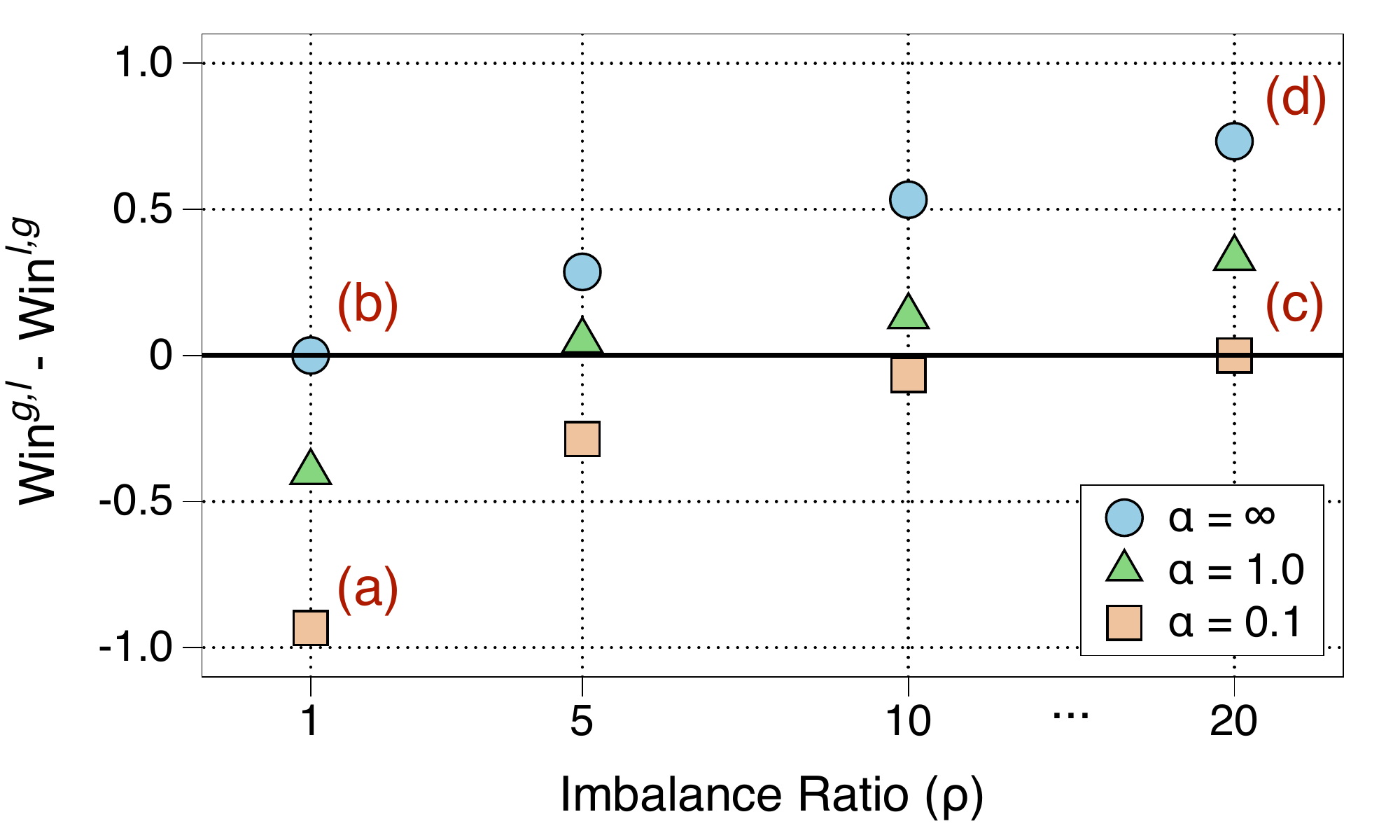}
 \vspace*{-7pt}
\caption{Gap between the winning rate of global and local-only models by varying global imbalance ratio ($\rho$) and local heterogeneity ($\alpha$) on CIFAR-10 benchmark. The experimental setups for (a)-(d) are also compatible with Figure\,\ref{fig:obs2_cnt_acc} and Table\,\ref{tab:emd}.}
\vspace{-16pt}
\label{fig:cifar_lt}
\end{figure}

\begin{observation} 
The superiority of local-only and global query-selecting models varies according to the degree of local heterogeneity and global imbalance ratio.
\label{obs:obs1}
\end{observation}

In Figure\,\ref{fig:cifar_lt}, we summarize the performance gap between two query models depending on the local heterogeneity level\,(indicated by different shapes) and the global imbalance ratio\,(increased along x-axis).
The y-axis represents the gap of the winning rate in Eq.\,\eqref{eq:cell_value} between global and local-only models; thus, the value becomes positive up to +1 if the global model beats the local-only model, otherwise negative up to -1. 
At a glance, there is a clear and consistent superiority of two query models according to $\alpha$ and $\rho$, where the dominance has intensified toward both extremes (\eg, upper right and lower left).
This observation contradicts the previous findings that the global model has always outperformed the local-only model as the query selector in a FAL framework\cite{f_al}. 
We provide more in-depth analysis in following Obs.\,\ref{obs:obs2} and Obs.\,\ref{obs:obs3}. \qed

\begin{figure*}[t!]
    \centering
    \begin{subfigure}[b]{0.47\linewidth}
    \centering
\includegraphics[width=\linewidth]{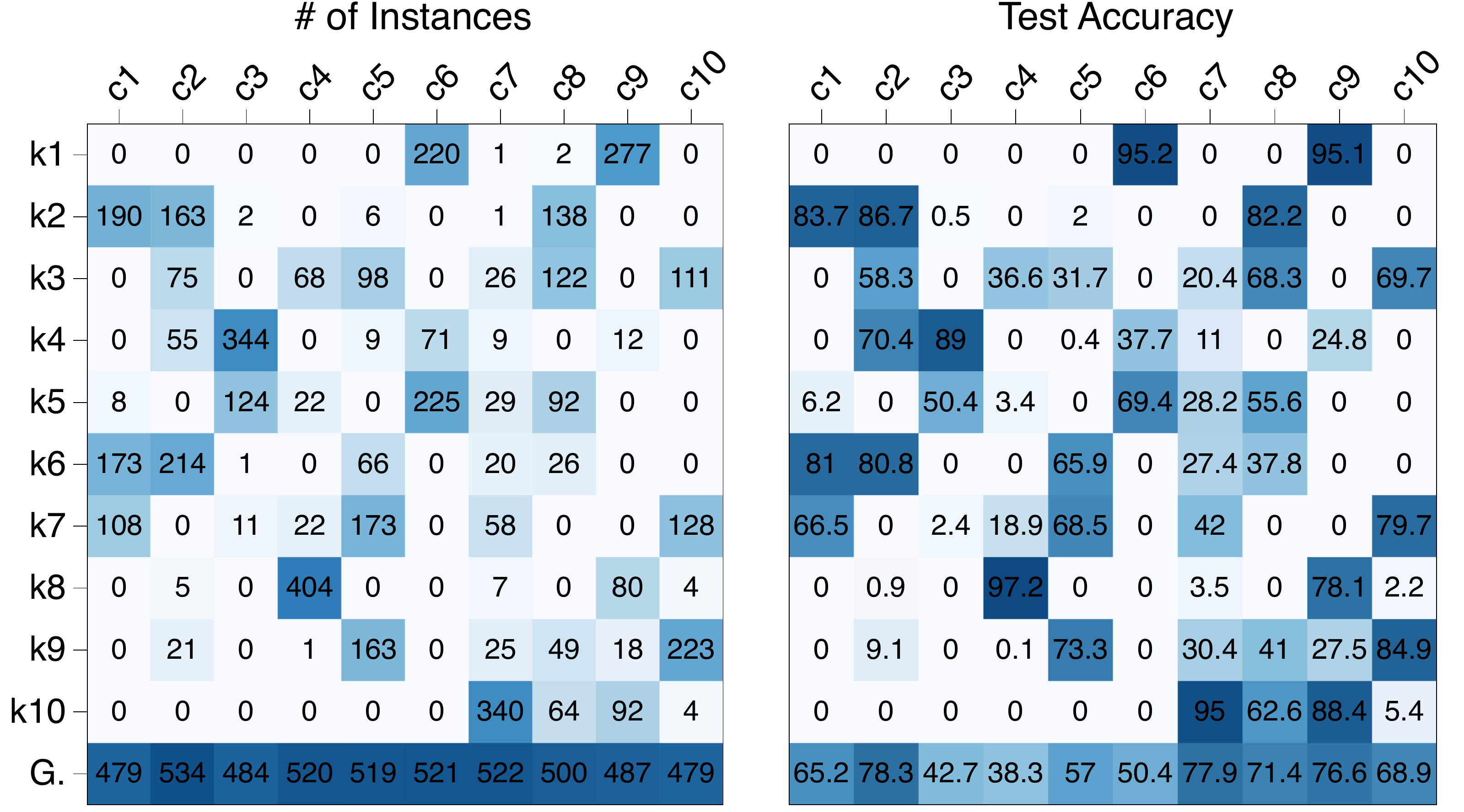}
    \caption{Low global imbalance ($\rho=1$) and high heterogeneity ($\alpha=0.1$).}     
    \end{subfigure}
    \hspace{15pt}
    \centering
    \begin{subfigure}[b]{0.47\linewidth}
    \centering
    \includegraphics[width=\linewidth]{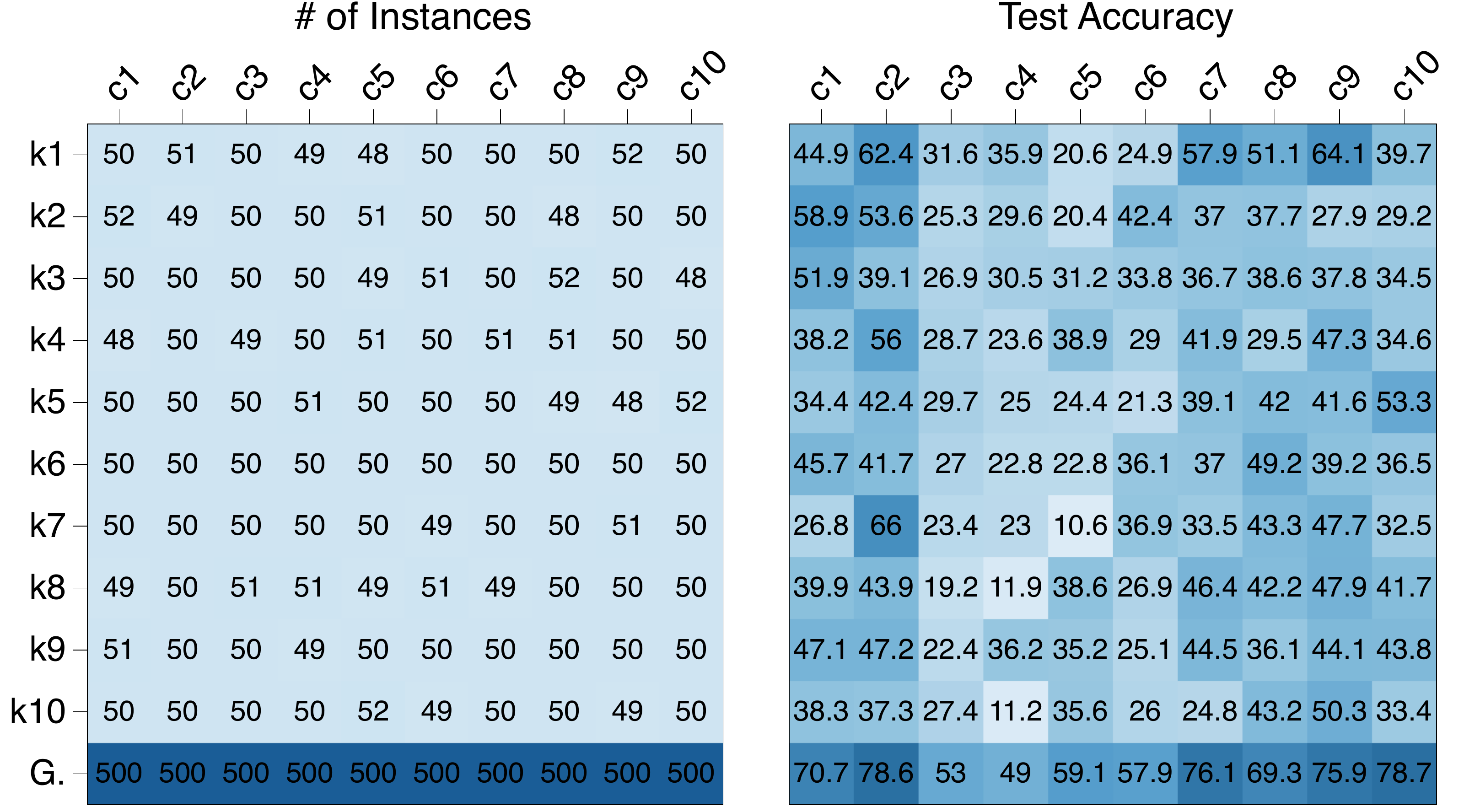}
    \caption{Low global imbalance ($\rho=1$) and low heterogeneity ($\alpha=\infty$).}       
    \end{subfigure}
    \begin{subfigure}[b]{0.47\linewidth}
    \centering
    \includegraphics[width=\linewidth]{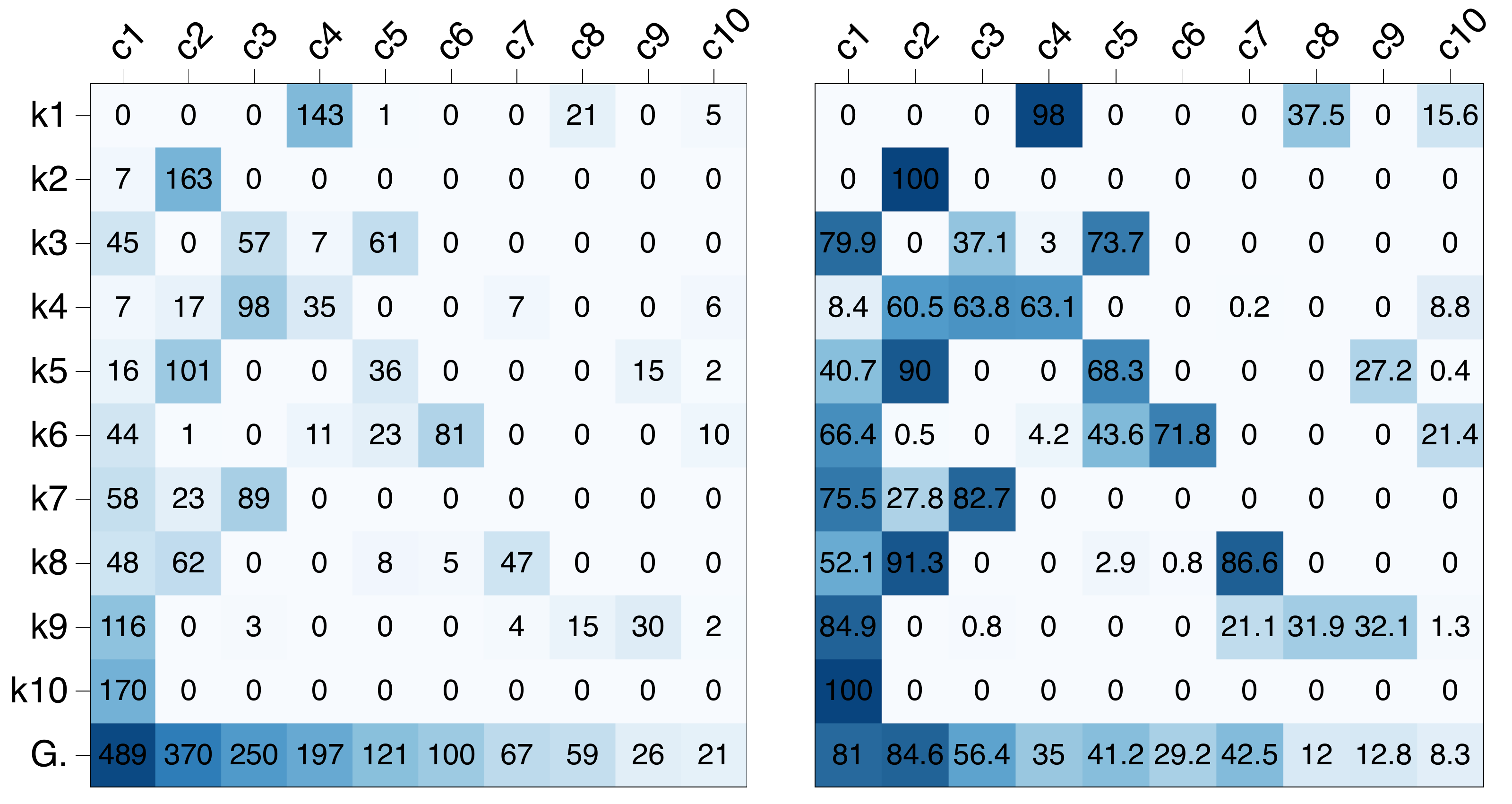}
    \caption{High global imbalance ($\rho=20$) and high heterogeneity ($\alpha=0.1$).}
    \end{subfigure}
    \hspace{15pt}
    \begin{subfigure}[b]{0.47\linewidth}
    \centering
    \includegraphics[width=\linewidth]{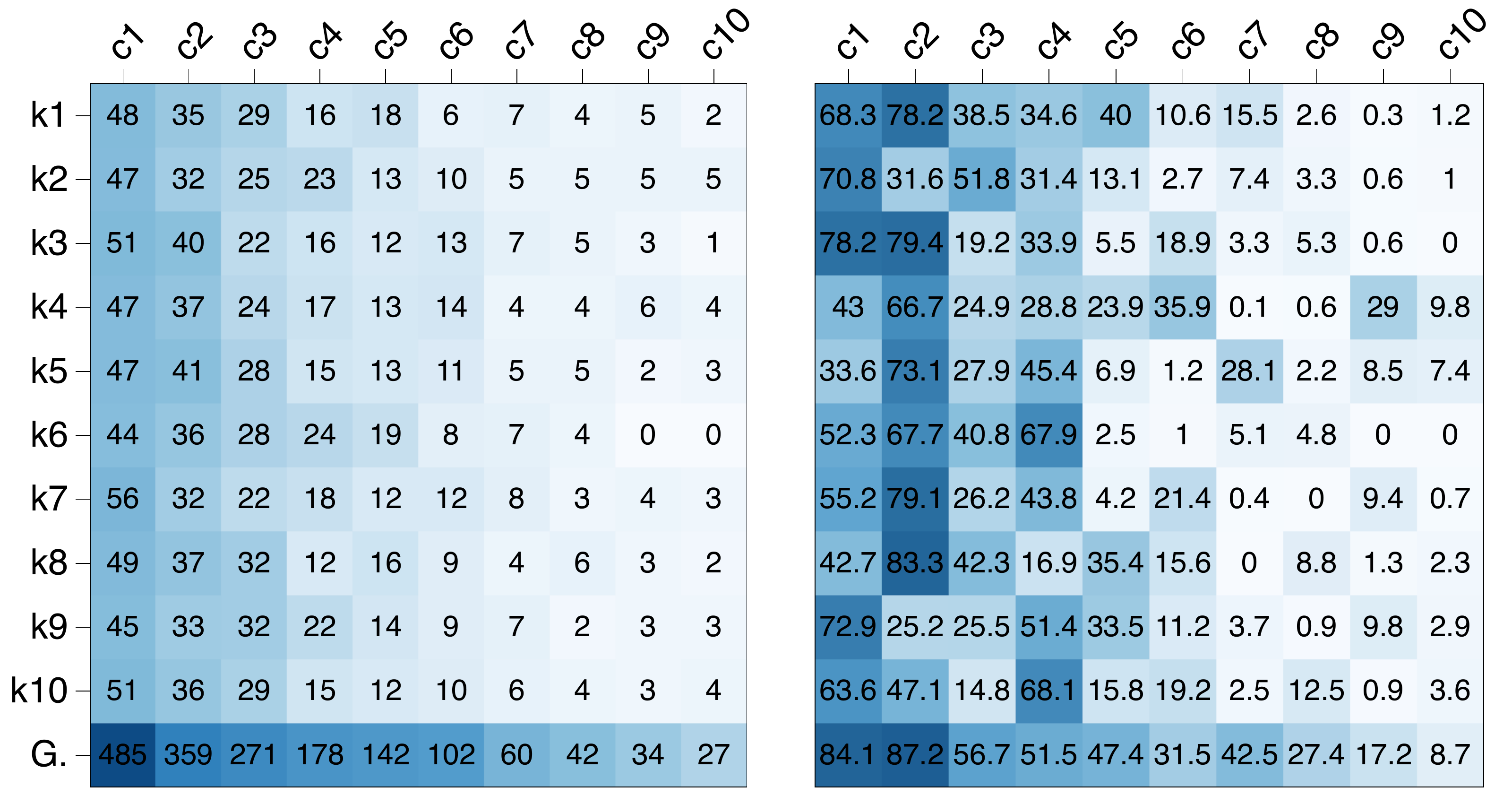}
    \caption{High global imbalance ($\rho=20$) and low heterogeneity ($\alpha=\infty$).}  
    \end{subfigure}
    \vspace*{-0.2cm}
    \caption{Matrices of data count\,(left) and class-wise accuracy\,(right) for CIFAR-10, with four combinations of $\rho=\{1, 20\}$ and $\alpha=\{0.1, \infty\}$. k1--10 denote ten clients, c1--10 are ten classes of CIFAR-10. 
    There are two types of data counts, \# of per-client local instances (1-10th rows) and aggregated instances over all clients (the last row). Similarly, the test accuracy is measured with local-only models (1-10th rows) and a global model (the last row). 
    (a)--(d) setups correspond to those of Figure\,\ref{fig:cifar_lt}. See Appendix\,\ref{sec:detail_analysis_matrices} for the more cases.
    }
    \vspace*{-0.2cm}
    \label{fig:obs2_cnt_acc}
\end{figure*}

\noindent \\
\vspace{-15pt}
\begin{observation} 
As local heterogeneity increases\,($\alpha \downarrow$), a local-only query selector is preferred due to the increased significance of local inter-class diversity. 
\label{obs:obs2}
\end{observation}

As the collapse of the local inter-class balance (\ie, lower $\alpha$) incurs severe performance degradation due to a weight divergence\cite{astraea, fedavg}, addressing local imbalance can improve the learning stability and performance.
Since the local-only models are separately trained on each client, in general, the local-only model has shown higher confidence for its own data distribution than the global model\cite{fedbabu, fedrep}.
In Figure\,\ref{fig:obs2_cnt_acc}, we visualized the number of class instances and the class-wise test accuracy in the first AL round (see the caption for details).
Specifically, we confirmed a high correlation between counts and accuracy in Figure\,\ref{fig:obs2_cnt_acc}-(a), such that the local-only models have higher accuracy than the global model for major classes of their data distribution.
Thus, by the nature of favoring low-confident instances in AL, the local-only model tends to select the instances with local minority classes as the query.

More precisely, we verify that the local-only model indeed queries the locally balanced set using earth mover's distance\,(EMD)\cite{emd_measure}. 
In Table\,\ref{tab:emd}, local EMD measures the mean of distance between class distribution of local query sets and a uniform distribution. The lower the value, the more balanced the locally queried instances.
As shown in Table\,\ref{tab:emd}-(a) with high local heterogeneity, the local EMD of the local-only model\,(L) is lower than that of the global model\,(G). 
That is, the local-only model queries more diverse instances than the global model with respect to the local inter-class diversity. 

Meanwhile, in the case of (b), the global model, trained with more samples, has higher accuracy over the classes due to little distribution discrepancy.
Although the more accurate model is likely to have the higher prediction confidence, it does not mean that it is better at identifying the required instances based on the current local dataset, which the global model had not directly learned. 
In practice, the local-only model still chose the more locally balanced query set (Table\,\ref{tab:emd}-(b)), and we supposed this contradiction makes no sizeable winning gap of the case (b) in Figure\,\ref{fig:cifar_lt}. \qed

\begin{table}[t!]
    \small
    \renewcommand*{\arraystretch}{1}
    \addtolength{\tabcolsep}{-1pt}
    \resizebox{\linewidth}{!}{
    \begin{tabular}{c|c|cccc|cccc}
         \toprule
        & & \multicolumn{4}{c|}{\bf Obs.\,\ref{obs:obs2}: Local EMD $(\downarrow)$} & \multicolumn{4}{c}{\bf Obs.\,\ref{obs:obs3}: Global EMD $(\downarrow)$} \\
        \cmidrule(l{2pt}r{2pt}){3-6} \cmidrule(l{2pt}r{2pt}){7-10}
        \multirow{-2.5}{*}{\!\!Case} & \multirow{-2.5}{*}{\!Model\,\!\!} & 10\% &  20\% & 30\% & 40\% &  10\% & 20\% & 30\% & 40\% \\
        \midrule
        \multirow{2}{*}{\!\!(a)} & G & 0.632 & 0.638 & 0.641 & 0.643 & 0.019 & 0.064 & 0.086 & 0.095  \\
        & L & 0.632 &  0.597 &  0.592 &  0.595 & 0.019 &  0.050 &  0.050 &  0.046  \\
        \hline
        \multirow{2}{*}{\!\!(b)} & G & 0.049 & 0.077 & 0.070 & 0.084 & 0.014 & 0.070 & 0.066 & 0.063  \\
        & L & 0.049 &  0.042 &  0.054 &  0.059 & 0.014 &  0.025 &  0.044 &  0.053  \\
        \hline
       \multirow{2}{*}{\!\!(c)}  & G & 0.692 & 0.680 & 0.676 & 0.674 & 0.377 &  0.300 &  0.294 &  0.294  \\
        & L & 0.692 &  0.641 &  0.633 &  0.636 & 0.377 & 0.334 & 0.326 & 0.321 \\\hline
        \multirow{2}{*}{\!\!(d)}  & G & 0.371 &  0.298 &  0.284 &  0.274 & 0.368 &  0.294 &  0.282 &  0.272 \\
        & L & 0.371 & 0.313 & 0.293 & 0.290 & 0.368 & 0.309 & 0.287 & 0.288  \\ \bottomrule
    \end{tabular}}
    \vspace*{-0.2cm}
    \caption{Local EMD and global EMD on CIFAR-10. We summarize the results of four AL rounds with the labeling budget of 10\% per round. (a)--(d) setups correspond to those of Figure \ref{fig:cifar_lt}. See Appendix\,\ref{sec:detail_analysis_emd} for EMDs of more cases.}
    \label{tab:emd}
    \vspace{-12pt}
\end{table}

\begin{observation}
As the degree of global class imbalance increases ($\rho \uparrow$), it is more advantageous to exploit a global model that alleviates the global class imbalance.
\label{obs:obs3}
\end{observation}

Based on Obs.\,\ref{obs:obs2}, the local-only model should outperform the global model in the case (c), but there is no clear superiority between them with respect to the winning rate in the case (c) of Figure \ref{fig:cifar_lt}.
The only answer for this conundrum is the presence of global minority classes due to the high global imbalance ratio.
The local heterogeneity is obviously a crucial factor by Obs.\,\ref{obs:obs2}, but global class imbalance is also another factor that significantly degrades the classification performance in the FAL framework.
Here, the major challenge is that neither the central server nor local clients cannot access any information of aggregated data due to privacy preservation.
The only way to address this problem, we should utilize the global model that implicitly learns the knowledge of the entire data distribution through the aggregation phase in Eq.\,\eqref{eq:reform_fl_loss}.

We introduce an additional global EMD, the indicator of measuring the inter-class diversity of the aggregated queried set over all clients. 
As can be seen in Table \ref{tab:emd}-(c), where the global imbalance ratio is high, we confirm that the global query selector\,(G) favors to query global minority classes. 
The global EMD of the global model is lower than that of the local-only model, \ie the more globally balanced query set, but the local EMD is the opposite. 

Meanwhile, in the case of (d), the minority classes are always the same from the global and local perspectives.
It is different from the case (b), where the minority classes with respect to the accuracy differ in each client depending on the informativeness of instances despite the same instance number.
Therefore, in this scenario, the global model has high confidence even in local datasets, leading to significantly overwhelming the local-only model in the case (d) of Figure \ref{fig:cifar_lt}.
In conclusion, the global inter-class diversity is an essential factor when global imbalance exists. \qed

\section{Method}
\label{subsec:lg_fal}

From three observations in Section \ref{sec:analysis}, we confirmed that FAL framework requires delicate consideration of local and global inter-class diversity. 
However, since clients are reluctant to share their data information, we should simultaneously utilize local-only and global models to ensure both sides of inter-class diversity.
To this end, we propose a novel query sample strategy, named \algname{}, that composed of \textit{macro} and \textit{micro} steps.
Pseudo algorithm for \algname{} is provided in Appendix\,\ref{sec:pseudo_algorithm}.
Before describing details of each step, let us assume a scenario where $k$-th client queries $B$ unlabeled instances at the round of $r$.

\smallskip \smallskip
\noindent 
\textbf{Macro Step: Clustering with Local-Only Model.} \\
\indent The ultimate goal of macro step is to satisfy local inter-class diversity by primarily sampling the informative instances by the \emph{local-only} model.
In detail, we introduce $k$-means clustering\cite{k_means} on the hypothetical gradient embedding.
Let $z$ be the embedding vector before forwarding to the last layer $W$ for $x \in U_k^r$.
Here, we utilize the gradient of negative cross-entropy loss, induced by a pseudo label, with respect to the last layer of encoder as follows:
\vspace{-1pt}
\begin{equation}
g^x_{c} = -\frac{\partial}{\partial W_{c}} \ell_{CE}(x, \hat{y}; \Theta_{k*}^r) = z\!\cdot\!(\mathbbm{1}_{[\hat{y}=c]} - p_{c}),
\label{eq:gradient}
\vspace{-1pt}
\end{equation}
where $\hat{y} = \arg\max_{c \in [C]} \,p_c$ and $W_{c}$ is the weights connected to $c$-th neuron of logits. 
Gradient embedding is widely used in conventional AL algorithm\cite{badge, venkatesh2020ask}, and we consider only the gradients corresponding to the pseudo label (\ie, $g^x_{\hat{y}}$) for computation efficiency.

Then, we calculate $B$ number of centroids on the hallucinated gradient space via EM algorithm\cite{dempster1977maximum} of $k$-means clustering by minimizing
\vspace{-1pt}
\begin{equation}
    J = \sum_{i=1}^{N} \sum_{b=1}^{B} w_{ib} \lVert g^{x_i}_{\hat{y}} - \mu_b  \rVert^2,
\label{eq:kmeans}
\vspace{-1pt}
\end{equation}
where $w_{ib}$ is an indicator function whether $g^{x_i}_{\hat{y}}$ is assigned to $\mu_b$ for E-step.
Eq.\,\eqref{eq:gradient} shows that the gradient embedding is a just scaling of feature embedding $z$, especially with the scale of uncertainty. 
In other words, if an instance is uncertain to predict\,(low value of $p_{\hat{y}}$), its gradient will be much highly scaled.
In this sense, Eq.\,\eqref{eq:kmeans} can be regarded as a weighted $k$-means clustering\cite{duda2006pattern, spath1980cluster} on the feature embedding space by using the function of softmax response\cite{geifman2017selective}. 
Hence, the macro step of \algname{} enable the query set to contain both diversity in the embedding space and uncertainty from the perspective of local-only model.

\smallskip \smallskip
\noindent 
\textbf{Micro Step: Cluster-wise Sampling with Global Model.}

\indent
In micro step, the \emph{global model} selects the final instances that results in the higher global inter-class diversity.
Given $B$ selected instances from the macro step, the most uncertain instance is selected for each cluster:
\begin{equation}
L^{r}_k = \{\mathcal{A}(\mathcal{C}_1, \Theta^{r*}, 1), ..., \mathcal{A}(\mathcal{C}_B, \Theta^{r*}, 1)\}
\label{eq:micro_step}
\end{equation}
$C_b$ denotes $b$-th cluster, generated by the query set from the macro step. We simply used Entropy sampling for $\mathcal{A}$, but we should note that $\mathcal{A}$ can be any AL sampling strategy. 

The cluster-wise sampling in the micro step is a simple yet effective strategy to leverage the advantages of both query selector models, ensuring local and global inter-class diversity. 
Here, we discuss how \algname{} can be a promising sampling strategy for the FAL framework.

\begin{remark}
The micro step is same with one step of EM algorithm. We first update cluster assignments with $B$ number of centroids from the macro step\,(E-step).
Then, within each cluster, we select the most informative instance based on the uncertainty score. This can be regarded as one M-step of the weighted $k$-means clustering, using infinitely scaled weight function of uncertainty measure.
\label{remark1}
\end{remark}

\noindent
Then, let $c_b$ be a centroid of $C_b$ and we define $M$ as follows:
\begin{equation}
    M = \sum_{b=1}^B \| c_b - \Tilde{x}_b \|^2,\,\, \text{where  } \Tilde{x}_b = \arg\min_{x} \| c_b - x \|^2
    \label{eq:remark_2}
\end{equation}
where $x \in L_k^r$ (Eq.\,\eqref{eq:micro_step}) and $\Tilde{x}$ is one-to-one mapping with the minimal transport cost.
The lower value of $M$ means that the final query set much consider the diversity in the macro step.
\begin{remark}
EM-based sampling in the micro step guarantees the diversity information of the macro step.
Through sampling one instance per cluster, each $\Tilde{x}_b$ is disjointly assigned to one cluster. Therefore, \algname{} conservatively ensures the local inter-class diversity from the macro step (\ie, lower $M$ value), comparing to any other strategy that query at least two instances for one cluster.
\label{remark2}
\end{remark}
\begin{figure*}[t]
\centering
\vspace*{-0.5cm}
\includegraphics[width=0.98\linewidth]{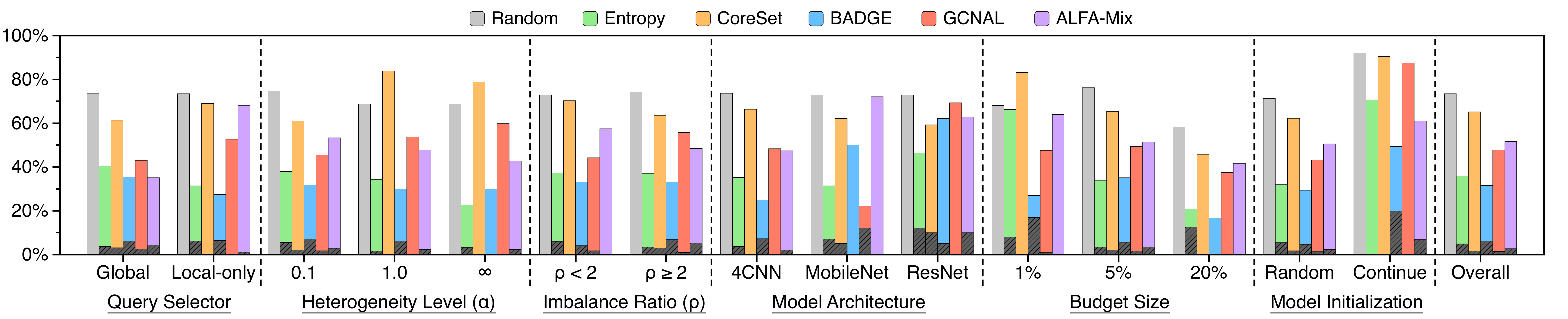}
\vspace{-5pt}
\caption{Winning percentage across six categories. We also added defeat percentage, the black hatched bar that represents the percentage at which \algname{} has been defeated by each baseline. Among total experiments, only statistically reliable values ($t$-score $>$ 2.776) are considered.
Thus, the lower value of the colored bar and the higher value of the black bar indicate a more comparable baseline.}
\label{fig:bar_graph}
\vspace*{-3pt}
\end{figure*}

\vspace{-0.2cm}
\section{Evaluation}
\label{sec:evaluation}

\subsection{Experimental Configuration}
\label{app:implementation_detail}

\paragraph{Training Settings.}
In a FAL framework, a central server should carefully consider the fairness of labeling and training costs between clients. Therefore, we assume that ten clients have the same size of unlabeled data pool and query the same number of instances per every AL round.
Moreover, we focus on a cross-silo FL setting\cite{survey_fl} where every client participates in every FL round.

We compared \algname{} with six active learning strategies on five benchmark dataset (CIFAR-10\cite{cifar}, SVHN\cite{svhn}, PathMNIST, DermaMNIST, and OrganAMNIST\cite{medmnist}).
We considered 38 comprehensive experimental settings combined into six categories (see Figure\,\ref{fig:bar_graph}).  
Except for learning ablations of the architectures, labeling budgets, and initialization schemes, in default, we implemented four layers of CNN and trained the encoder from scratch with the labeling budget of 5\% for every AL round. 
We repeat all experiments four times and report their averaged values. Refer to Appendix \ref{sec:exp_settings} for detailed experimental settings.

\vspace{-0.4cm}
\paragraph{Baselines.}

\begin{figure}[t!]
 \centering
 \includegraphics[width=0.85\linewidth]{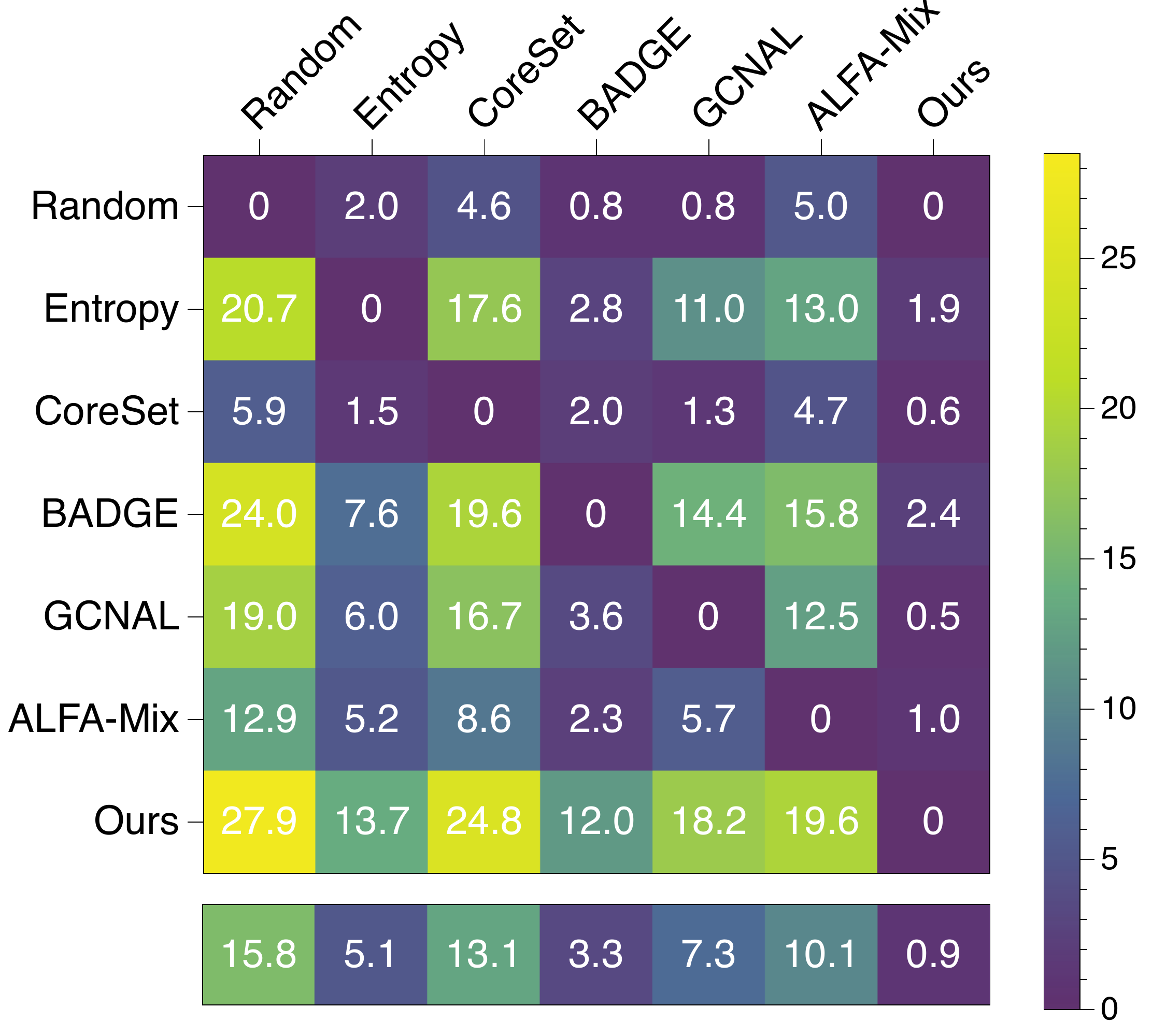}
 \vspace*{-0.1cm}
\caption{Pairwise penalty matrix over 38 experimental settings. The value $P_{ij}$ indicates the number of times that the $i$-th strategy outperforms the $j$-th strategy (\ie, sum of ${\sf win}^{ij}$ in Eq.\,\eqref{eq:cell_value} over 38 settings).
The last row is the average number of times the j-th strategy is defeated by the rest strategies; the lower, the better.
}
 \vspace*{-0.3cm}
\label{fig:comparision_matrix}
\end{figure}

We considered six standard AL strategies.
\emph{Random} randomly samples the instances from the unlabeled pool. 
\emph{Entropy} selects instances with the largest entropy\cite{confidence_sampling}.
\emph{CoreSet} chooses the small subset that can represent the whole unlabeled set\cite{coreset}.
\emph{BADGE} selects the diverse points with high magnitude in a hallucinated gradient space\cite{badge}.
\emph{GCNAL} adapts CoreSet on a sequential graph convolution network to measure the relation between labeled and unlabeled instances\cite{gcnal}.
\emph{ALFA-Mix} identifies the valuable instances by seeking inconsistencies in the prediction of mixing features\cite{alfa_mix}.
We adopt these sampling strategies with either the global model or local-only model in FedAvg pipeline.
Refer to Appendix\,\ref{sec:computational_cost} for the comparison of query selection cost. Besides, we summarized the results with various FL methods in Appendix\,\ref{sec:various_fl_algo}.

\subsection{Overall Comparison}
\label{exp:baselines}
\subsubsection{Pairwise Penalty Matrix}
We summarized the overall comparison results as a pairwise penalty matrix $P$ in Figure\,\ref{fig:comparision_matrix}, following the recent work\cite{badge, alfa_mix}.
Each cell of the pairwise penalty matrix represents the summation of the winning rate of Eq.\,\eqref{eq:cell_value} calculated above for each of the 38 experimental sets.
As the rows of the matrix $P_i$ indicate how many times $i$-th algorithm outperforms the other algorithms, the brighter color means the better algorithm (conversely, the darker columns are the better).
Note that we only considered statistically reliable results throughout the two-sided t-test.

\algname{} defeats all baselines in general (see 7-th row values) while under-performed only 0.9 out of 38 times on average (see 7-th column values in the last row).
In particular, \algname{} outperforms BADGE and Entropy sampling, which are top-2 baselines, with 13.7 and 12.0 out of 38 times, while \algname{} only loses 2.4 and 1.9 out of 38 times, respectively. 
This results demonstrate the robust effectiveness of our LoGo algorithm in various experimental settings.

\vspace{-8pt}
\subsubsection{{Winning Rate Bar Plots}}
\vspace{-1.5pt}

Figure \ref{fig:bar_graph} summarizes the comparison results according to the six categories, which is the breakdown of Figure\,\ref{fig:comparision_matrix} on a more systematic categorization.
For example, the colored bar of the overall category is calculated by dividing the total setting number of 38 in the 7-th row vector of Figure\,\ref{fig:comparision_matrix} (the black bar is from the 7-th column vector).
The colored bar represents the average percentage at which \algname{} defeats each baseline algorithm, and the higher bar means that \algname{} has won more.
Appendix\,\ref{sec:detail_comp_matrix} provides the detailed comparison penalty values according to each category.

Overall, \algname{} consistently shows an overwhelming winning percentage toward any baseline against the defeat percentage.
Interestingly, ALFA-Mix, the latest AL strategy, shows performance considerably varying depending on which query selector is selected (see the query selector in Figure\,\ref{fig:bar_graph}).
The reason would be that a local-only model, which is separately trained on the highly heterogeneous dataset, is not appropriate for the feature mixing or the sensitivity to lots of hyperparameters.
Moreover, Coreset, which does not consider uncertainty, cannot resolve the local and global imbalance, showing similar performance as Random sampling.
Thus, \algname{} is a superior algorithm for a FAL framework in that it is robust to the type of query selector or most combinations of experimental settings. 

\begin{table*}[t!]
    \small
    \centering
    \renewcommand*{\arraystretch}{1.1}
    \addtolength{\tabcolsep}{-1pt}
    \resizebox{\textwidth}{!}{

    \begin{tabular}{l|c|cccc|cccc|cccc|cccc}
         \toprule
         & & \multicolumn{4}{c|}{CIFAR-10} & \multicolumn{4}{c|}{SVHN} & \multicolumn{4}{c|}{PathMNIST} & \multicolumn{4}{c}{DermaMNIST} \\
         \cmidrule(l{2pt}r{2pt}){3-6} \cmidrule(l{2pt}r{2pt}){7-10} \cmidrule(l{2pt}r{2pt}){11-14} \cmidrule(l{2pt}r{2pt}){15-18}
         \multirow{-2.5}{*}{Method} & \multirow{-2.5}{*}{Model} & 20\% & 40\% & 60\% & 80\% & 20\% & 30\% & 40\% & 50\% & 20\% & 30\% & 40\% & 50\% & 20\% & 30\% & 40\% & 50\% \\
         \midrule
        Random & - & 64.19&69.07&71.63&72.81&80.90&83.07&84.22&84.77&68.41&72.70&73.76&75.49&71.70&72.57&72.66&72.86 \\
        \hline
         & \textit{G} & 64.02&69.12&71.87&73.33&82.08&84.61&85.88&86.31&71.54&74.39&75.91&76.65&72.49&72.63&73.02&73.20 \\
        \multirow{-2}{*}{Entropy \cite{confidence_sampling}} & \textit{}L & 66.29&\underline{71.45}&\underline{73.51}&74.02&82.09&84.58&85.69&86.18&\textbf{76.52}&\underline{78.29}&\underline{78.71}&\underline{79.10}&71.38&72.04&72.22&72.65 \\
        \hline
         & \textit{G} & 64.66&69.43&71.75&73.1&80.94&82.74&83.81&84.46&74.84&76.24&76.85&76.80&72.02&72.16&72.34&72.74 \\
        \multirow{-2}{*}{Coreset \cite{coreset}} & \textit{L} & 64.06&68.79&71.49&73.28&80.94&82.92&83.78&84.48&72.53&76.06&76.28&76.86&71.13&71.48&72.15&72.38 \\
        \hline
         & \textit{G} & 65.12&69.57&72.11&73.53&82.81&84.82&85.89&86.2&72.21&74.38&75.53&76.97&\underline{72.59}&73.09&\underline{73.23}&\underline{73.45} \\
        \multirow{-2}{*}{BADGE \cite{badge}} & \textit{L} & \underline{66.32}&71.28&73.41&\underline{74.28}&82.69&84.67&85.61&86.1&76.48&78.51&78.42&78.68&71.35&72.13&72.25&72.99 \\
        \hline
         & \textit{G} & 65.40&70.05&72.41&73.42&82.05&84.07&85.09&85.61&75.51&77.79&78.13&78.81&72.01&72.60&73.07&73.17 \\
        \multirow{-2}{*}{GCNAL \cite{gcnal}} & \textit{L} &65.62&70.18&72.36&73.42&81.92&83.58&84.55&85.10&74.85&76.46&77.18&77.45&71.95&72.91&72.91&73.29 \\
        \hline
         & \textit{G} &65.45&69.87&72.24&73.29&\underline{83.02}&\underline{84.99}&\textbf{86.05}&\underline{86.33}&73.34&74.83&76.31&77.43&72.39&\underline{73.14}&73.27&73.10 \\
        \multirow{-2}{*}{ALFA-Mix \cite{alfa_mix}} & \textit{L} &64.14&68.79&71.03&72.6&81.08&82.55&83.62&84.33&71.10&75.01&75.81&76.70&71.51&72.18&72.94&73.28 \\
        \hline
        \textbf{LoGo (ours)} & \textit{G}, \textit{L} & \textbf{66.50}&\textbf{71.70}&\textbf{73.80}&\textbf{74.49}&\textbf{83.46}&\textbf{85.31}&\underline{86.02}&\textbf{86.38}&\underline{76.32}&\textbf{78.72}&\textbf{79.51}&\textbf{79.58}&\textbf{72.61}&\textbf{73.18}&\textbf{73.33}&\textbf{73.77} \\
    \bottomrule
    \end{tabular}}
    \vspace*{-0.2cm}
    \caption{Comparison of test accuracy on four benchmarks with $\alpha$ = 0.1. We reported the results with four random seeds. The baselines, except for Random sampling, are combined with two query selector models, $G$ and $L$ that stands for a global or local-only model, respectively. \textbf{Bold} and \underline{underline} mean Top-1 and Top-2, respectively.}
    \label{tab:acc_comparision}

\vspace{-3pt}
\end{table*}

\vspace*{-0.15cm}
\subsubsection{Under Query Selector Category}
Table\,\ref{tab:acc_comparision} shows the test accuracy according to the increased labeling budgets over rounds on four datasets.
Even with the same active learning strategy, a gap in test accuracy occurs depending on which query selector is used because the global imbalance varies by $1.0$--$58.7$ across datasets.
For example, in general, global models outperform local-only models for SVHN and DermaMNIST, while the opposite trend is observed for CIFAR-10 and PathMNIST.
However, irrespective of the benchmarks and querying model types, our \algname{} shows the best performance in most cases.
Two-step selection strategy enables \algname{} to be robust by utilizing both the benefits of global and local-only models.
Because we cannot know the degree of local and global imbalance in advance, our proposed method has a strong advantage in providing data-agnostic performance improvements over all baselines.
Appendix\,\ref{sec:detail_comp_performance} provides a detailed performance comparison between \algname{} and baselines under various experimental settings.

\subsection{\algname{} vs. Simple Ensemble}
\label{sec:simple_ensemble}
To prove that \algname{} is an effective way to leverage the advantages of both models, we compared \algname{} with three na\"ive implementation of ensemble methods for global and local-only models: (1) the average of logits\,(for Entropy sampling) or gradient embedding\,(for BADGE) from two models, (2) weighing the instances based on the selection ranks\,(\ie, more weights if it is selected from both models), and (3) fine-tuning the global model on local datasets.

In Table\,\ref{tab:compare_ens}, \algname{} consistently shows better classification accuracy over increasing labeling budgets than three counterparts. 
Compared with the results in both Tables\,\ref{tab:acc_comparision} and \ref{tab:compare_ens}, all three ensemble methods show lower performance than using a single superior query selector.
That is, the na\"ive ensemble suffers from a performance trade-off between two query selector models and, therefore, their results fall somewhere in the middle of using global and local models.

\begin{table}[t!]
    \small
    \centering
    \renewcommand*{\arraystretch}{1.1}
    \addtolength{\tabcolsep}{-1pt}
    \resizebox{\linewidth}{!}{
    \begin{tabular}{l|c|cccc|ccc}
        \toprule
         &  & \multicolumn{4}{c|}{CIFAR-10} & \multicolumn{3}{c}{SVHN} \\
         \cmidrule(l{2pt}r{2pt}){3-6} \cmidrule(l{2pt}r{2pt}){7-9}
        \multirow{-2.5}{*}{Method} & \multirow{-2.5}{*}{Strategy} & 20\% & 40\% & 60\% & 80\% & 20\% & 30\% & 40\% \\
        \midrule
         & +Entropy &64.53&70.36&73.02& \underline{74.28} & 81.81 &84.64&85.87  \\
         \multirow{-2}{*}{Ens.\,Logit} & +BADGE  &65.55&70.31&72.83&73.97&82.77&84.76&85.90  \\
         \hline
         & +Entropy &65.90&70.92& \underline{73.34} &74.20&82.15&84.38&85.64  \\
         \multirow{-2}{*}{Ens.\,Rank} & +BADGE  & \underline{66.21} & \underline{70.98} &73.15&74.01& \underline{83.02} & \underline{85.05} &85.86  \\
         \hline
         & +Entropy &65.10&70.75&73.21&74.23&82.53& \underline{85.05} & \underline{86.01}  \\
         \multirow{-2}{*}{Fine-tuning} & +BADGE  &65.82&70.95&72.94&74.12&82.59&84.89&85.82  \\
         \hline
         \textbf{LoGo\,(ours)} & - & \textbf{66.50} & \textbf{71.70} & \textbf{73.80} & \textbf{74.49} & \textbf{83.46} & \textbf{85.31} & \textbf{86.02}  \\
         \bottomrule
    \end{tabular}}
    \vspace*{-0.2cm}
    \caption{Comparison of test accuracy on two benchmarks\,($\alpha$=0.1) with baselines using both global and local information.}
    \label{tab:compare_ens}
\vspace{-0.35cm}
\end{table}

\section{Conclusion}

We discovered the superiority of the two query selectors according to the local heterogeneity level and global imbalance ratio. Based on our findings, the local-only and global models are both crucial since global and local inter-class diversity affects their performance dominance.
To this end, we propose LoGo algorithm that incorporates local-only and global models into macro and micro steps of cluster-based active learning.
LoGo preferentially selects samples that simultaneously alleviate local and global imbalance as a query, making it robust to local and global imbalance.
Our experiments verified that \algname{} consistently outperforms six baselines under comprehensive settings of 38 combinations using six categories.

\vspace{-12pt}
\paragraph{Acknowledgement.}
This work was supported by Institute of Information \& communications Technology Planning \& Evaluation (IITP) grant funded by the Korea government(MSIT) (No.2019-0-00075, Artificial Intelligence Graduate School Program(KAIST) and No. 2022-0-00871, Development of AI Autonomy and Knowledge Enhancement for AI Agent Collaboration).

{\small
\bibliographystyle{ieee_fullname}
\bibliography{egbib}
}

\clearpage
\appendix
\onecolumn
\addcontentsline{toc}{section}{Appendices}

\clearpage

\section{Detailed Local Data Distribution}
\label{sec:dataset_summary}

\begin{figure}[h!]
\centering
\begin{minipage}{0.94\linewidth}
    \begin{subfigure}[b]{0.32\linewidth}
    \includegraphics[width=\linewidth]{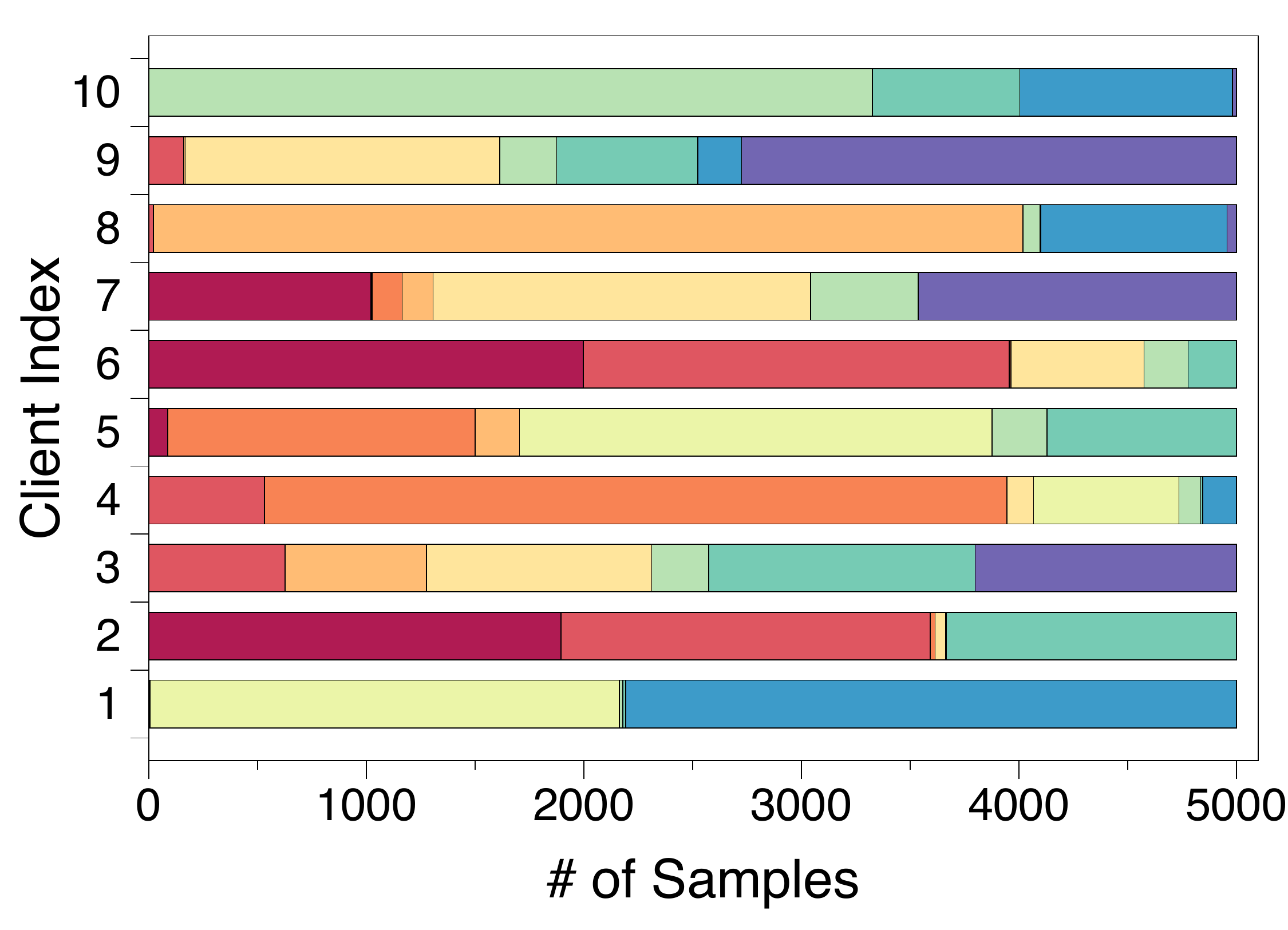}
    \caption{\small {$\rho$ = 1 and $\alpha$ = 0.1}}
    \end{subfigure}
    \hfill
    \begin{subfigure}[b]{0.32\linewidth}
    \includegraphics[width=\linewidth]{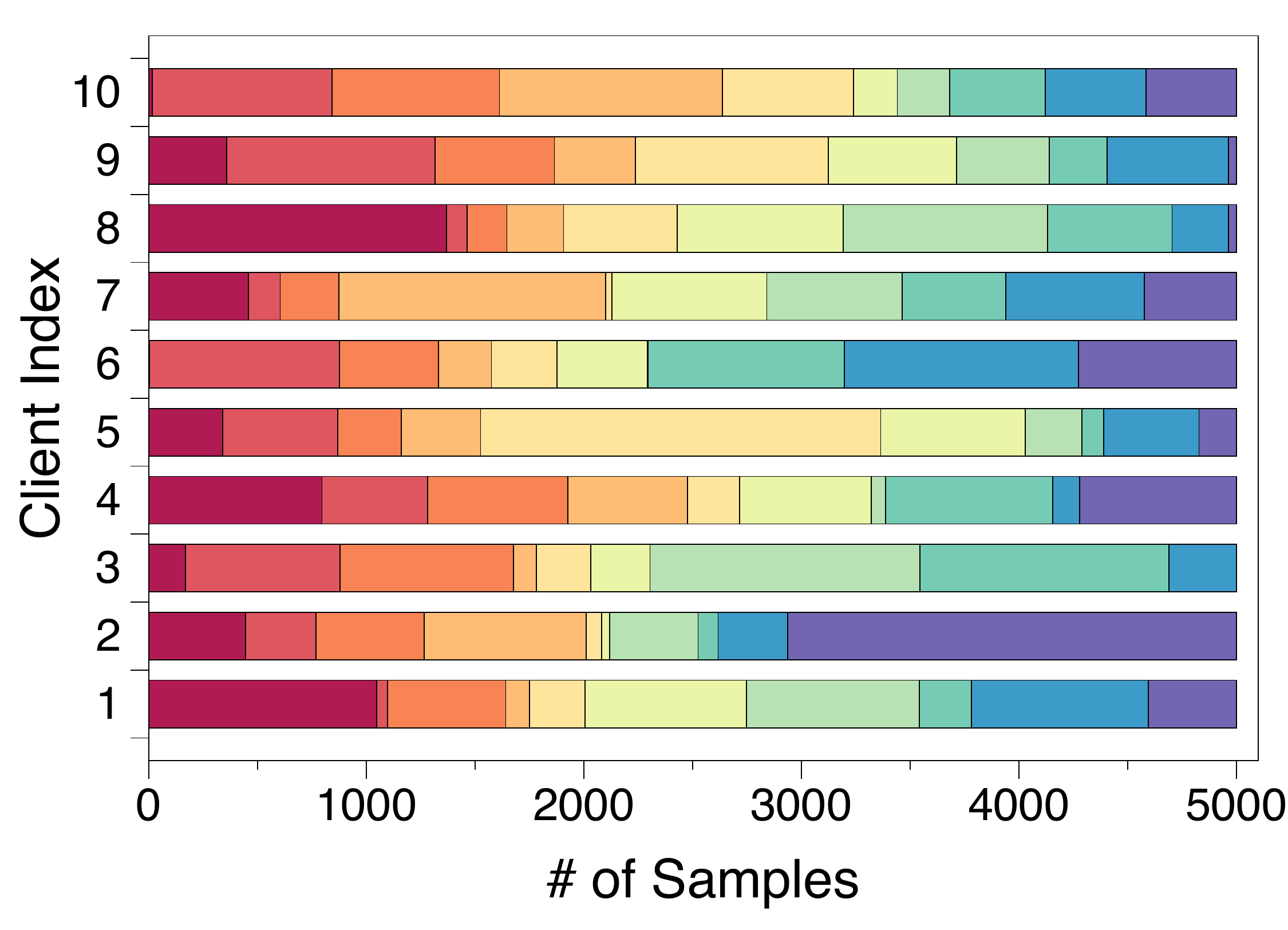}
    \caption{\small {$\rho$ = 1 and $\alpha$ = 1.0}}
    \end{subfigure}
    \hfill
    \begin{subfigure}[b]{0.32\linewidth}
    \includegraphics[width=\linewidth]{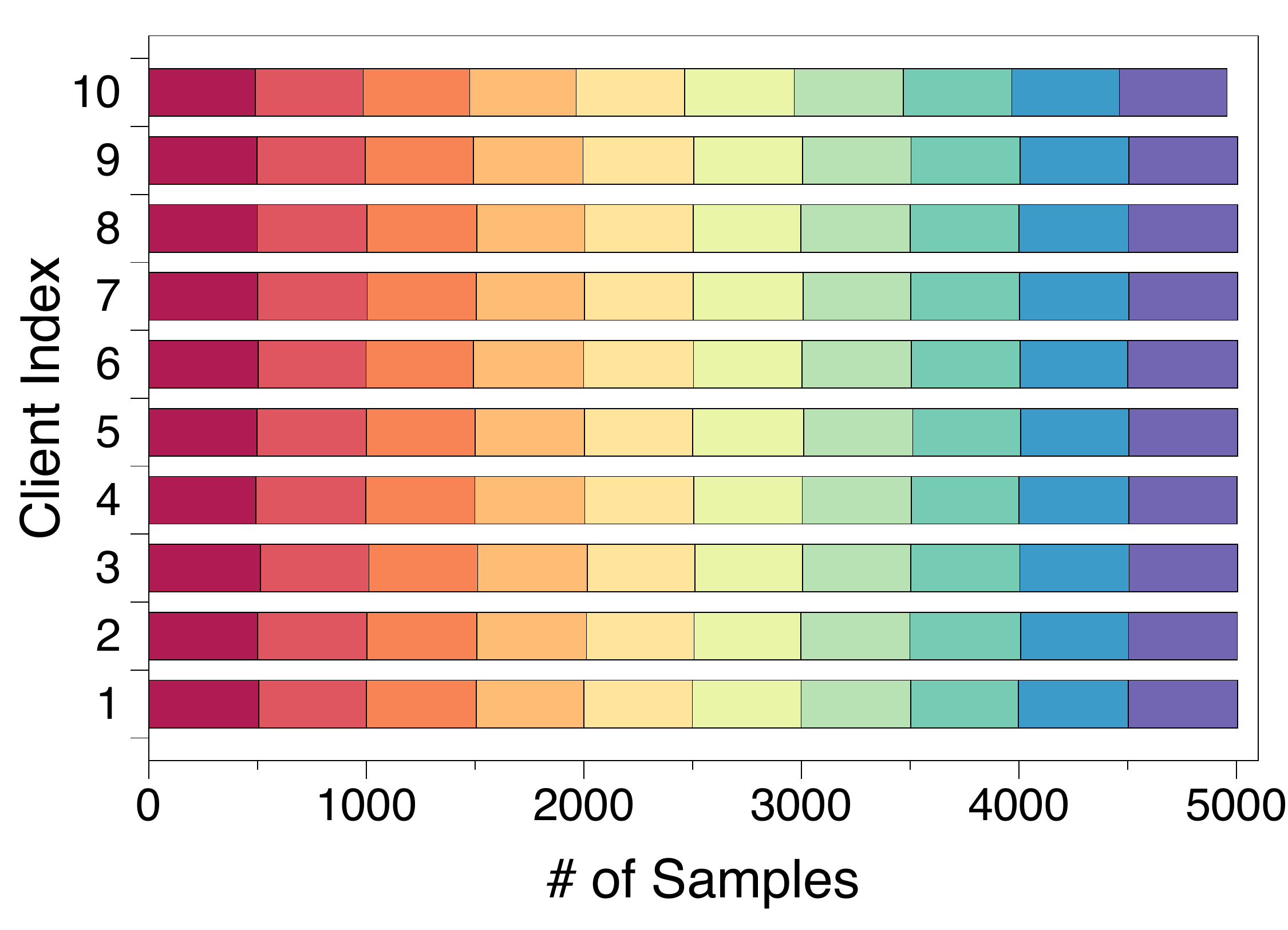}
    \caption{\small {$\rho$ = 1 and $\alpha$ = $\infty$}}
    \end{subfigure}
    \begin{subfigure}[b]{0.32\linewidth}
    \centering
    \includegraphics[width=\linewidth]{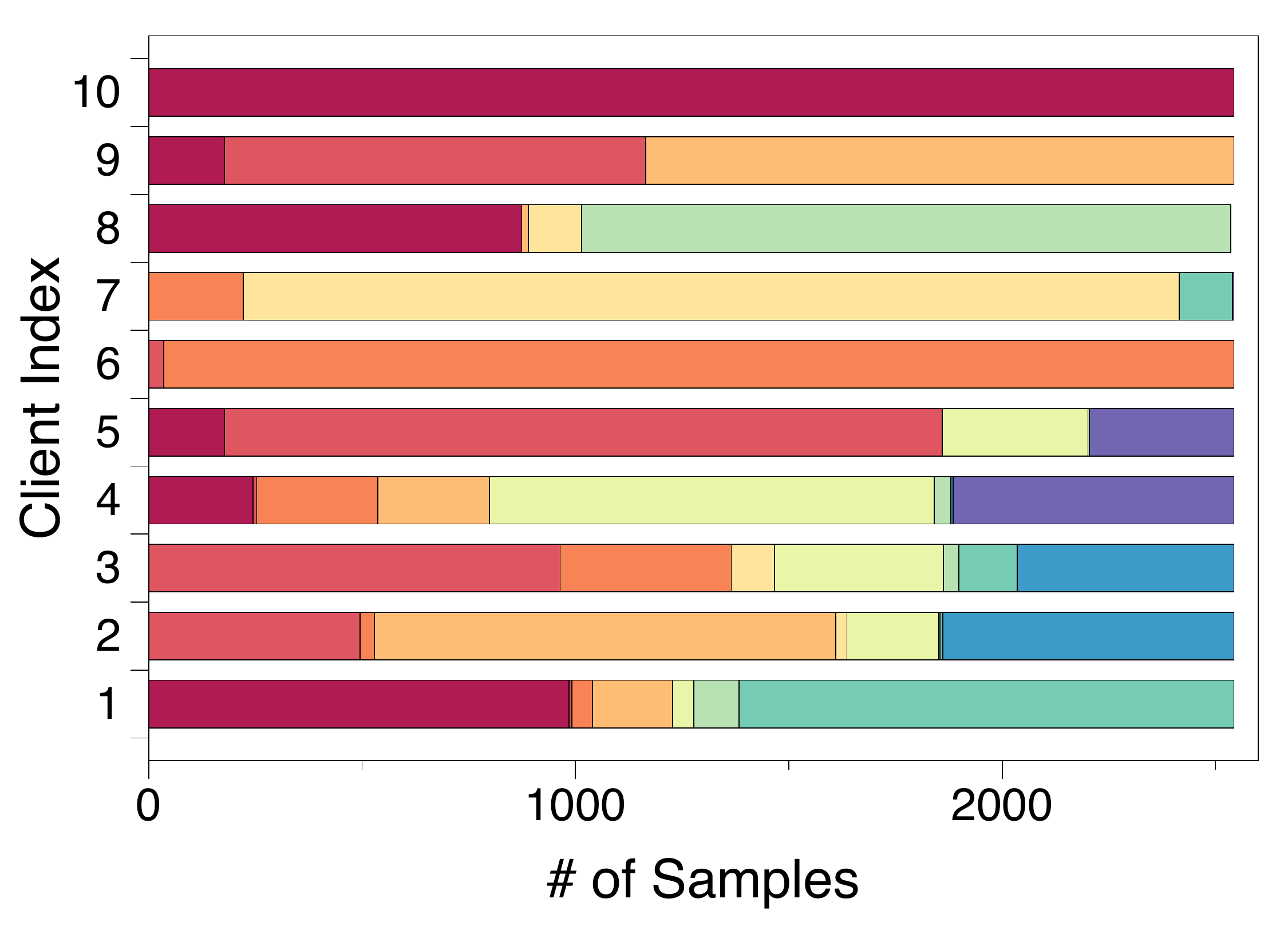}
    \caption{\small {$\rho$ = 5 and $\alpha$ = 0.1}}
    \end{subfigure}
    \hfill
    \begin{subfigure}[b]{0.32\linewidth}
    \centering
    \includegraphics[width=\linewidth]{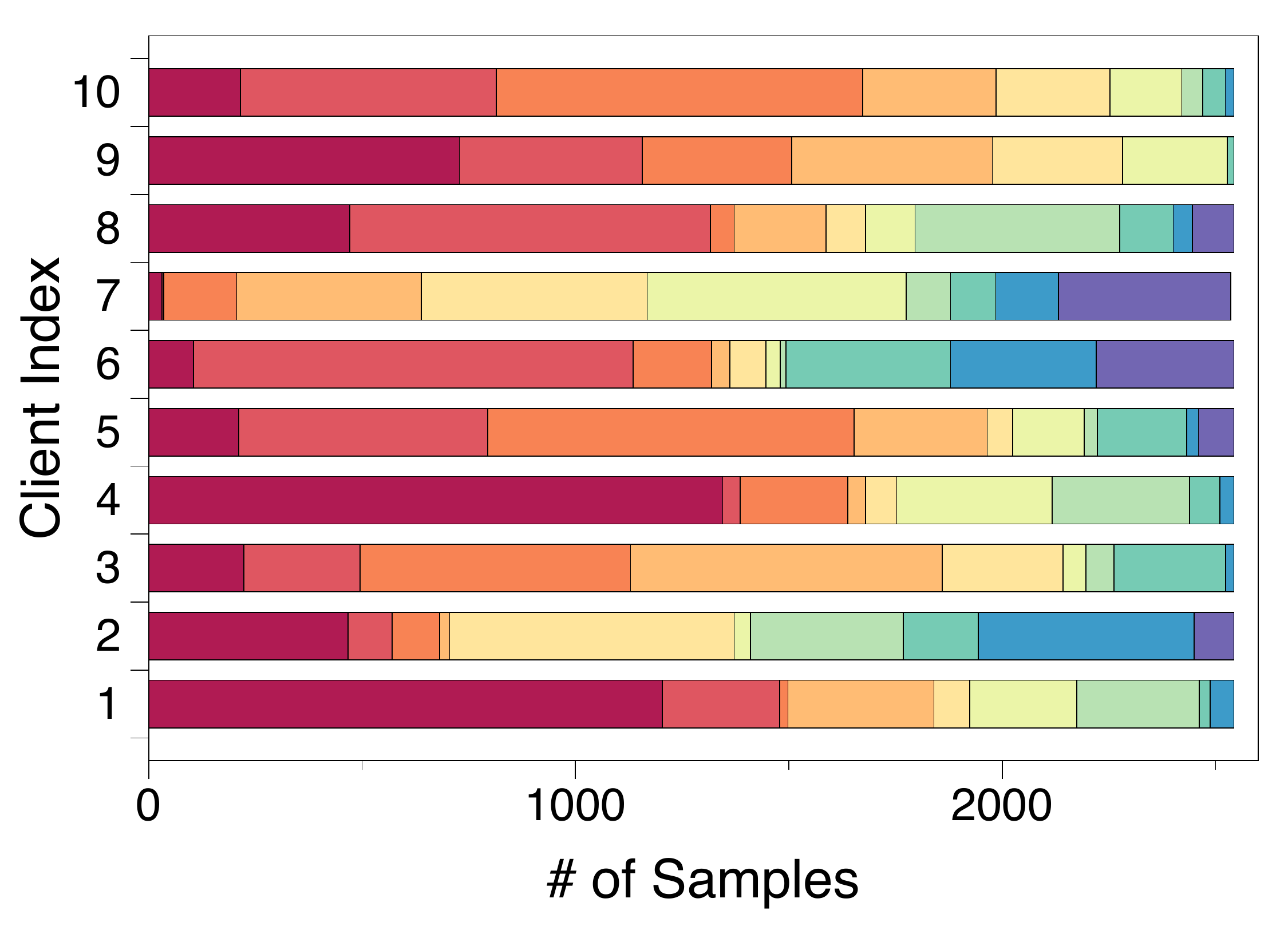}
    \caption{\small {$\rho$ = 5 and $\alpha$ = 1.0}}
    \end{subfigure}
    \hfill
    \begin{subfigure}[b]{0.32\linewidth}
    \centering
    \includegraphics[width=\linewidth]{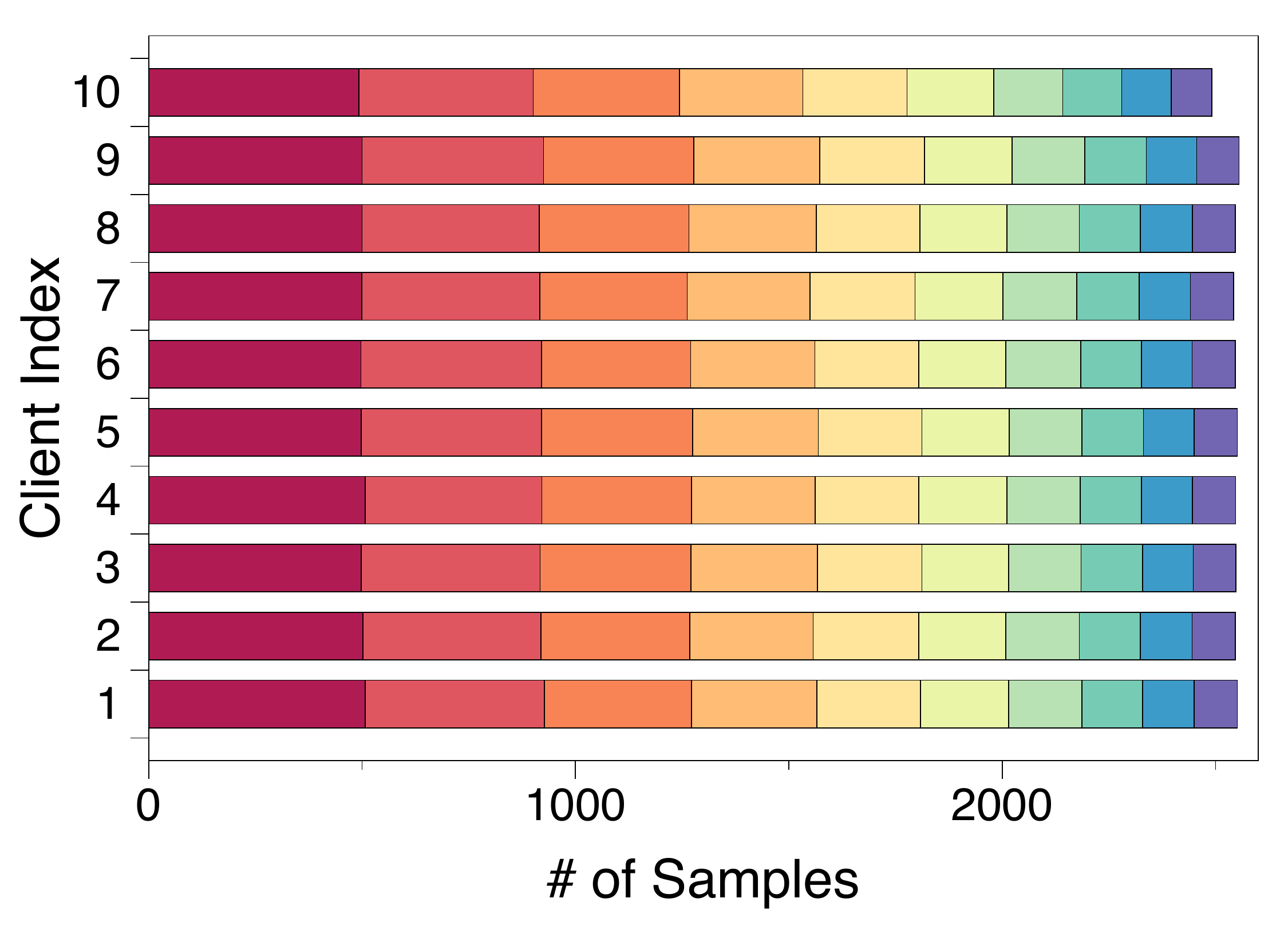}
    \caption{\small {$\rho$ = 5 and $\alpha$ = $\infty$}}
    \end{subfigure}
    \begin{subfigure}[b]{0.32\linewidth}
    \centering
    \includegraphics[width=\linewidth]{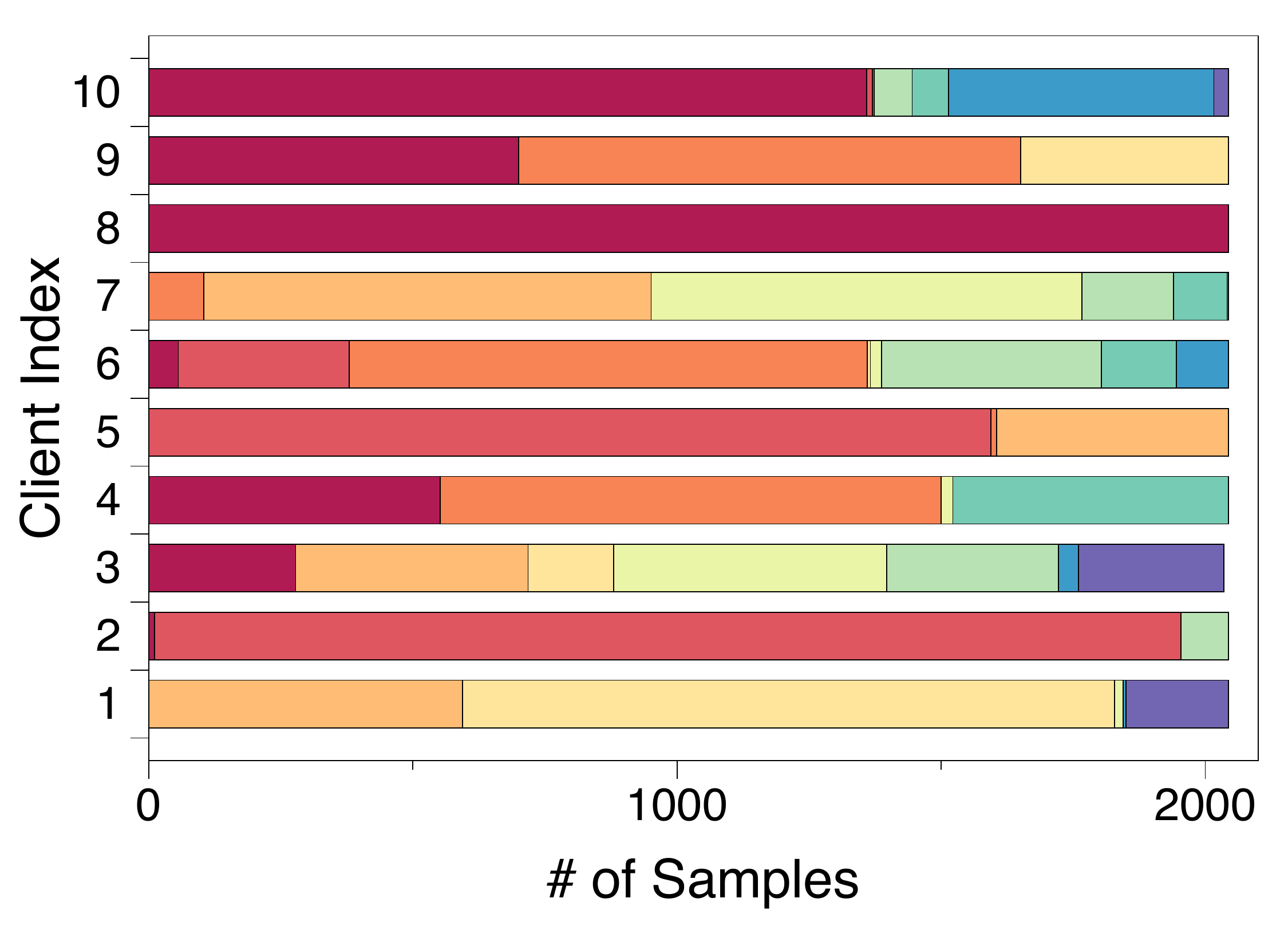}
    \caption{\small {$\rho$ = 10 and $\alpha$ = 0.1}}
    \end{subfigure}
    \hfill
    \begin{subfigure}[b]{0.32\linewidth}
    \centering
    \includegraphics[width=\linewidth]{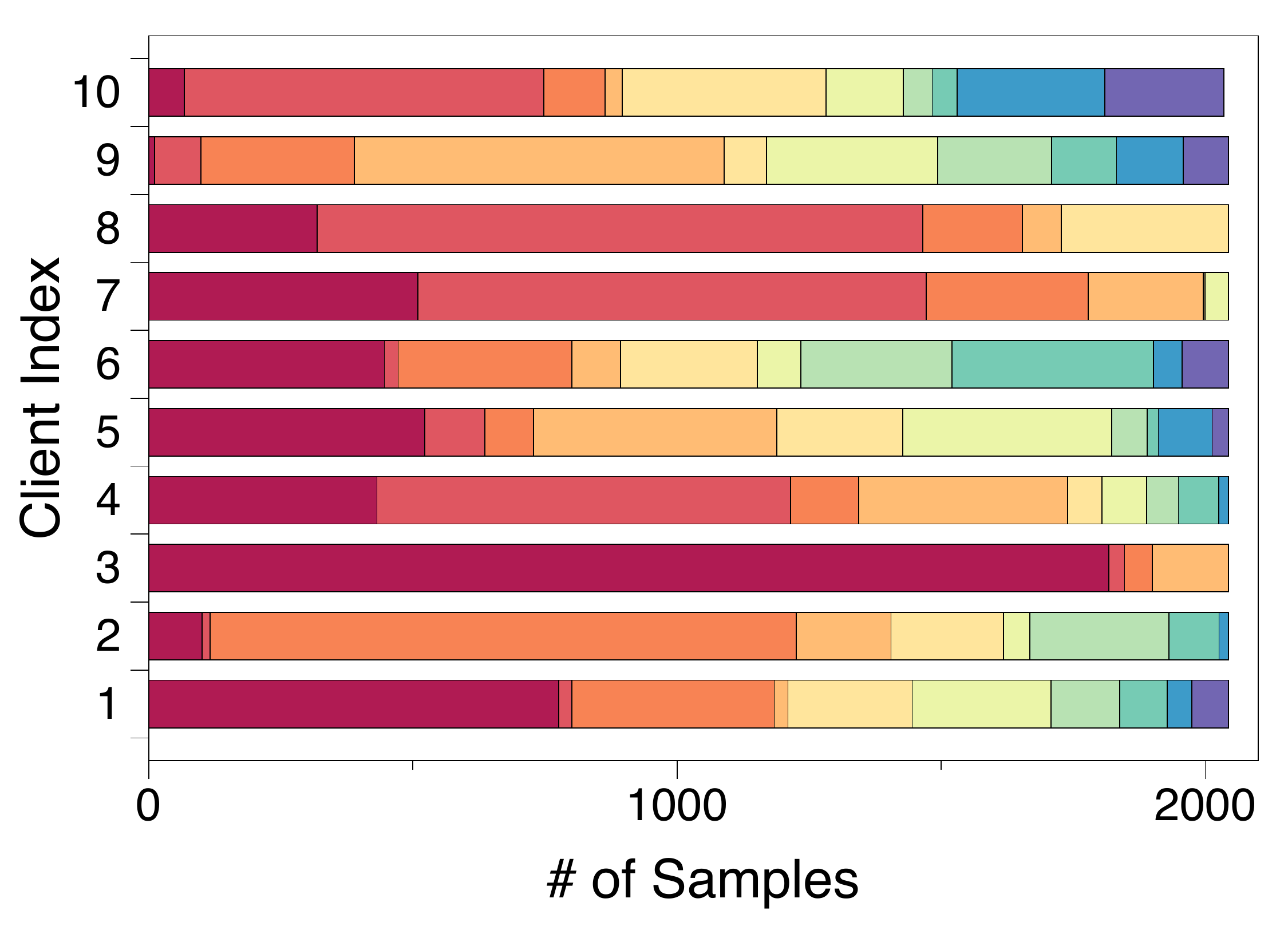}
    \caption{\small {$\rho$ = 10 and $\alpha$ = 1.0}}
    \end{subfigure}
    \hfill
    \begin{subfigure}[b]{0.32\linewidth}
    \centering
    \includegraphics[width=\linewidth]{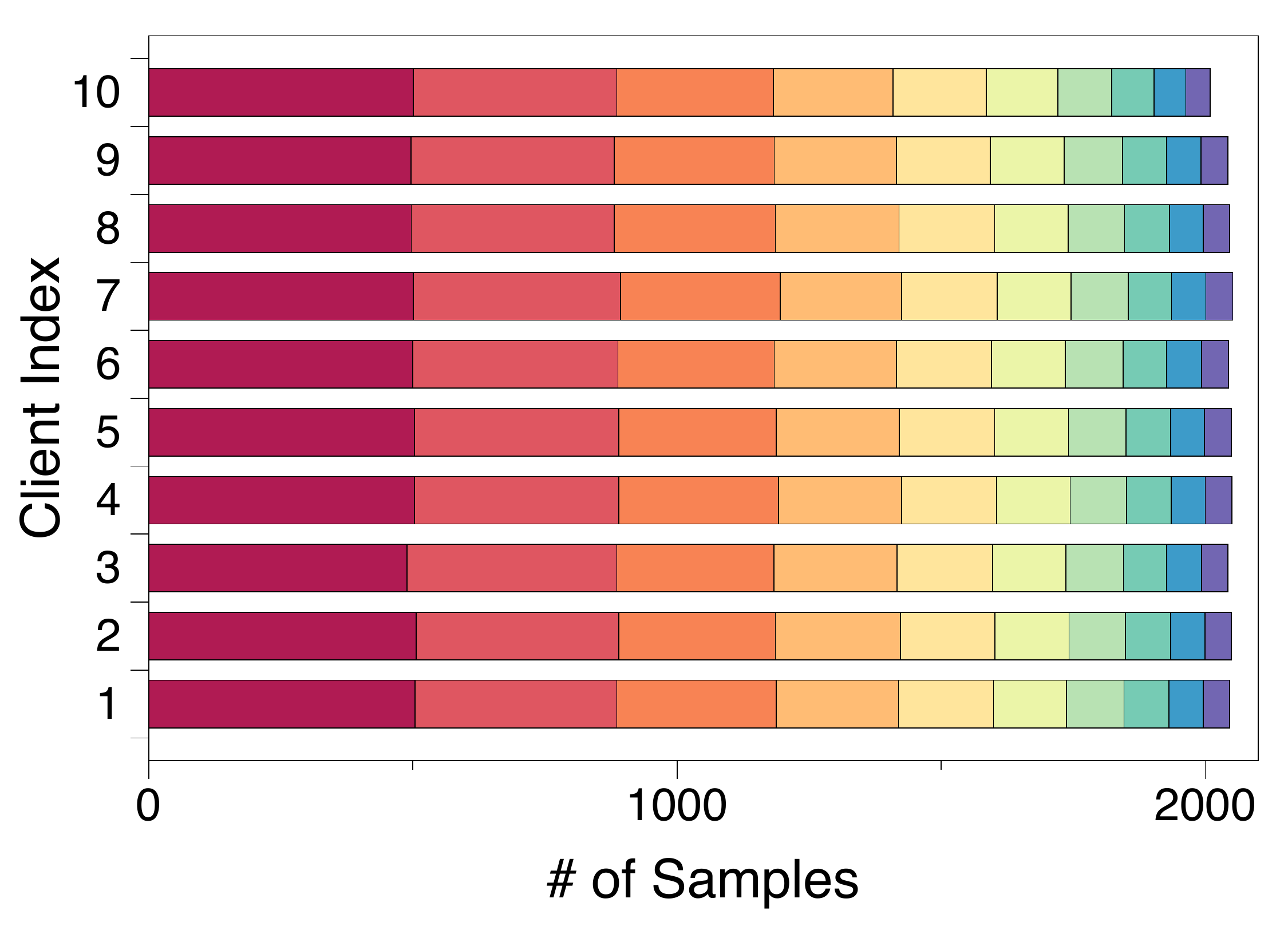}
    \caption{\small {$\rho$ = 10 and $\alpha$ = $\infty$}}
    \end{subfigure}
    \begin{subfigure}[b]{0.32\linewidth}
    \centering
    \includegraphics[width=\linewidth]{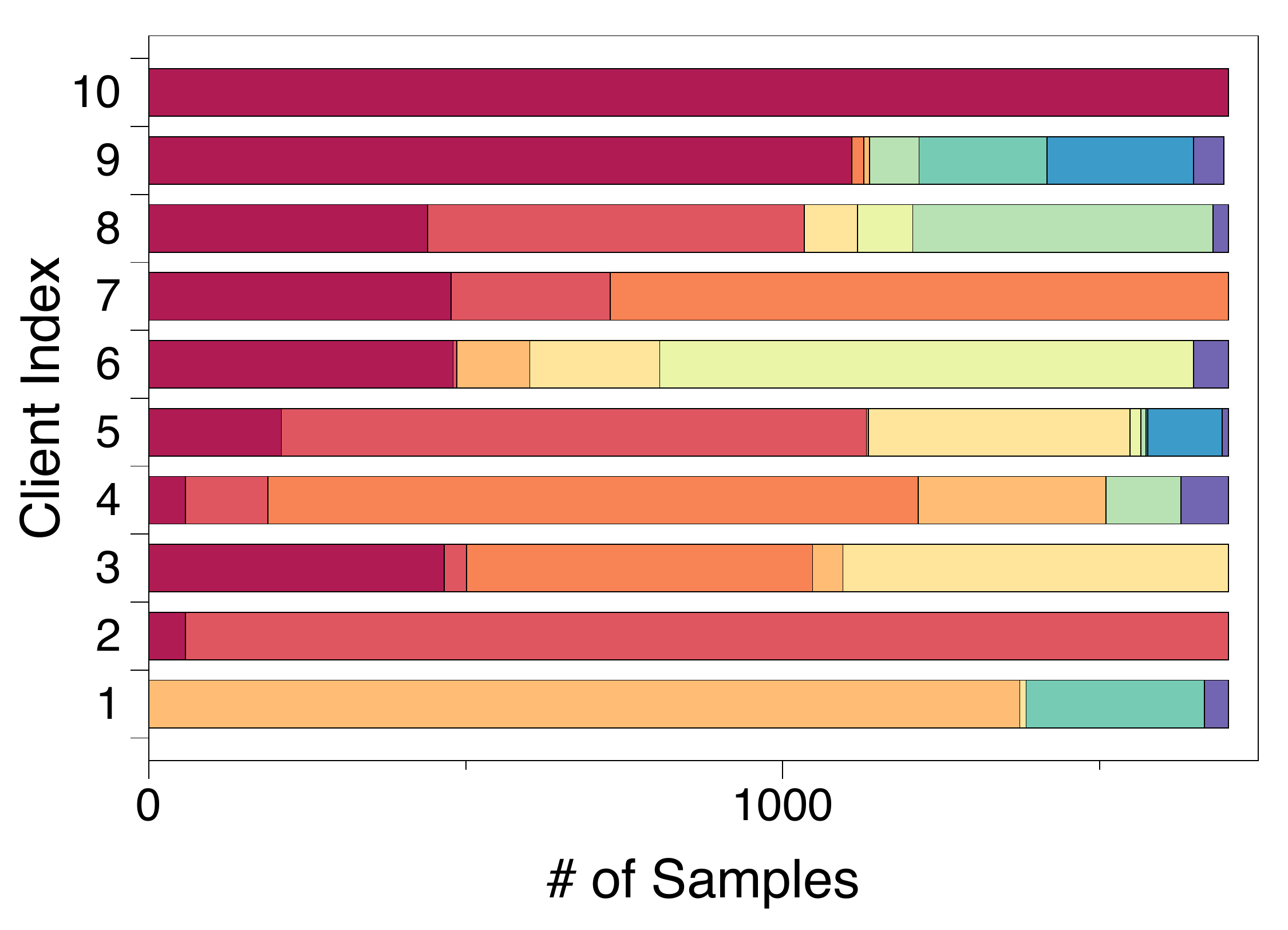}
    \caption{\small {$\rho$ = 20 and $\alpha$ = 0.1}}
    \end{subfigure}
    \hfill
    \begin{subfigure}[b]{0.32\linewidth}
    \centering
    \includegraphics[width=\linewidth]{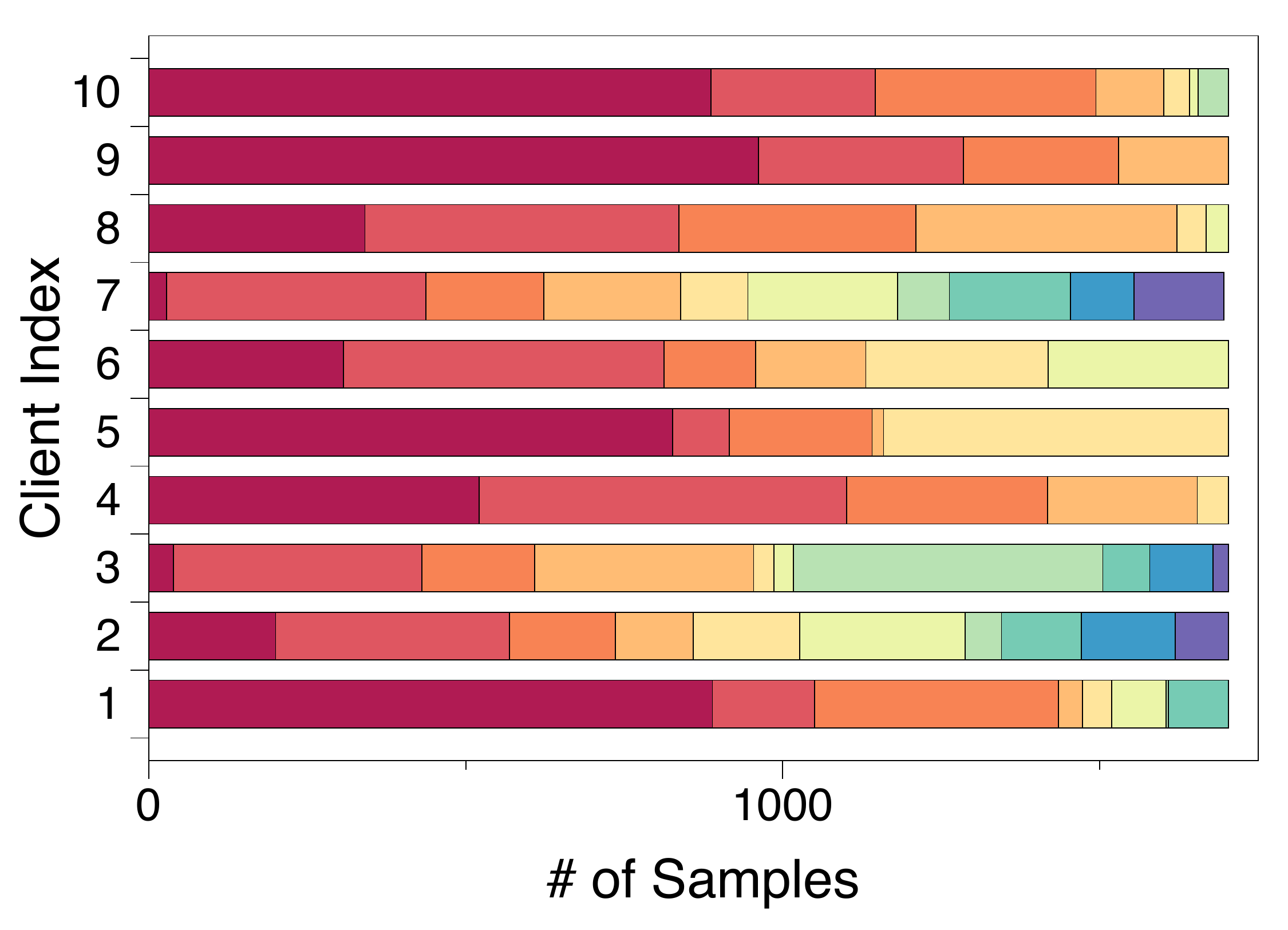}
    \caption{\small {$\rho$ = 20 and $\alpha$ = 1.0}}
    \end{subfigure}
    \hfill
    \begin{subfigure}[b]{0.32\linewidth}
    \centering
    \includegraphics[width=\linewidth]{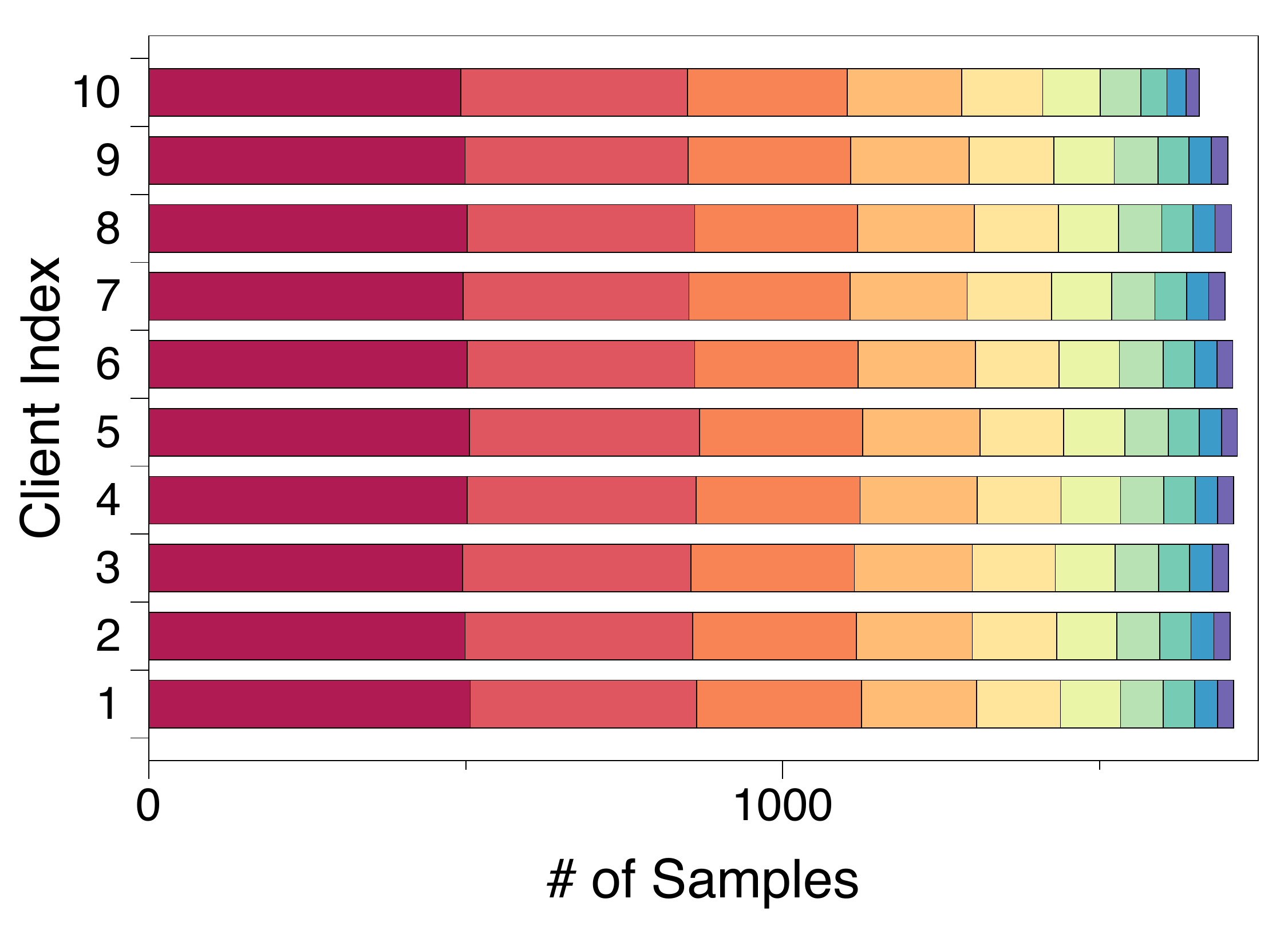}
    \caption{\small {$\rho$ = 20 and $\alpha$ = $\infty$}}
    \end{subfigure}
\end{minipage}
\caption{Visualization of the client data distribution on CIFAR-10. Each color represents a different class. The higher $\rho$ denotes the more global imbalanced distribution. The higher $\alpha$ denotes the more locally balanced data.}
\label{fig:data_dist}
\end{figure}

We adopt a Latent Dirichlet Allocation (LDA) strategy for Non-IID setting \cite{fedma, moon}, where each client $k$ is assigned the partition of classes by sampling $\mathbf{p}_k \sim Dir(\alpha \cdot \mathds{1})$, where $\mathds{1} \in \, \mathbb{R}^C$.
$\alpha$ is a concentration parameter that controls the local heterogeneity level. The smaller $\alpha$, the more heterogeneous data distribution.
Since we consider a fairness issue in the FAL framework, the total number of samples should be equally partitioned for all clients.
Therefore, we made a doubly stochastic matrix $P = [\tilde{\mathbf{p}}_1, \dots, \tilde{\mathbf{p}}_K]^\top$ by scaling $\mathbf{p}_k$ to $\tilde{ \mathbf{p}}_k$, when the number of client and class are same (i.e., $P$ is a square matrix).
Note that we set the sum of columns and rows to the proper values for a non-square matrix.
We visualized the examples of CIFAR-10 when the clients $K=10$ in Figure \ref{fig:data_dist}.

\clearpage

\section{Detailed Analysis Results}
\label{sec:detail_analysis}

\subsection{Detailed Matrices for Data Counts and Accuracy}
\label{sec:detail_analysis_matrices}

We summarized the detailed matrices for the combinations of $\rho$ = $\{$1, 5, 10, 20$\}$ and $\alpha$ = $\{$0.1, 1.0, $\infty \}$.

\begin{figure*}[!h]
\centering
\begin{minipage}{0.8\linewidth}
    \centering
    \begin{subfigure}[b]{0.32\linewidth}
    \centering
    \includegraphics[width=\linewidth]{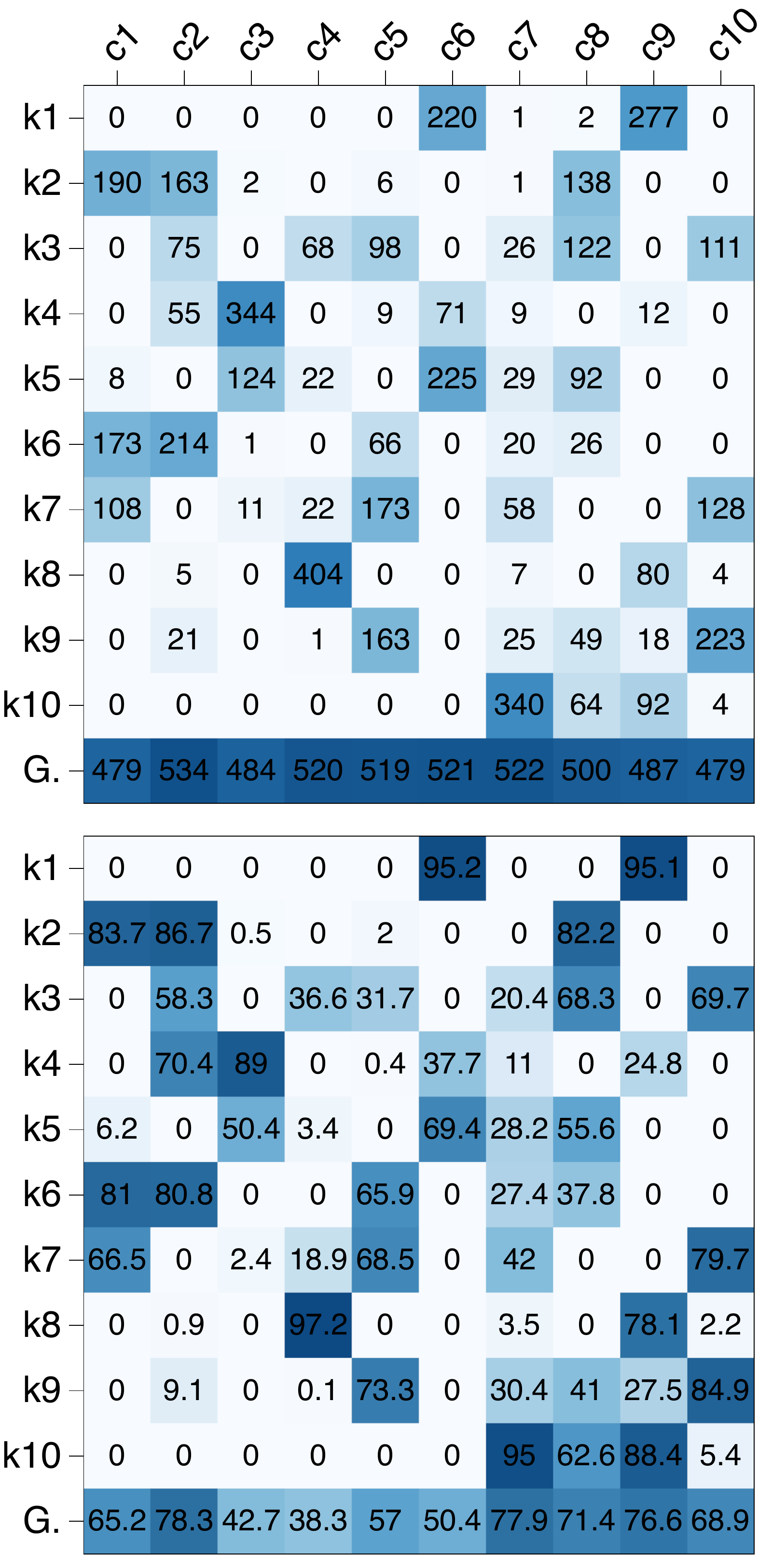}
    \caption{$\rho$ = 1 and $\alpha$ = 0.1}         
    \end{subfigure}
    \hfill
    \centering
    \begin{subfigure}[b]{0.32\linewidth}
    \centering
    \includegraphics[width=\linewidth]{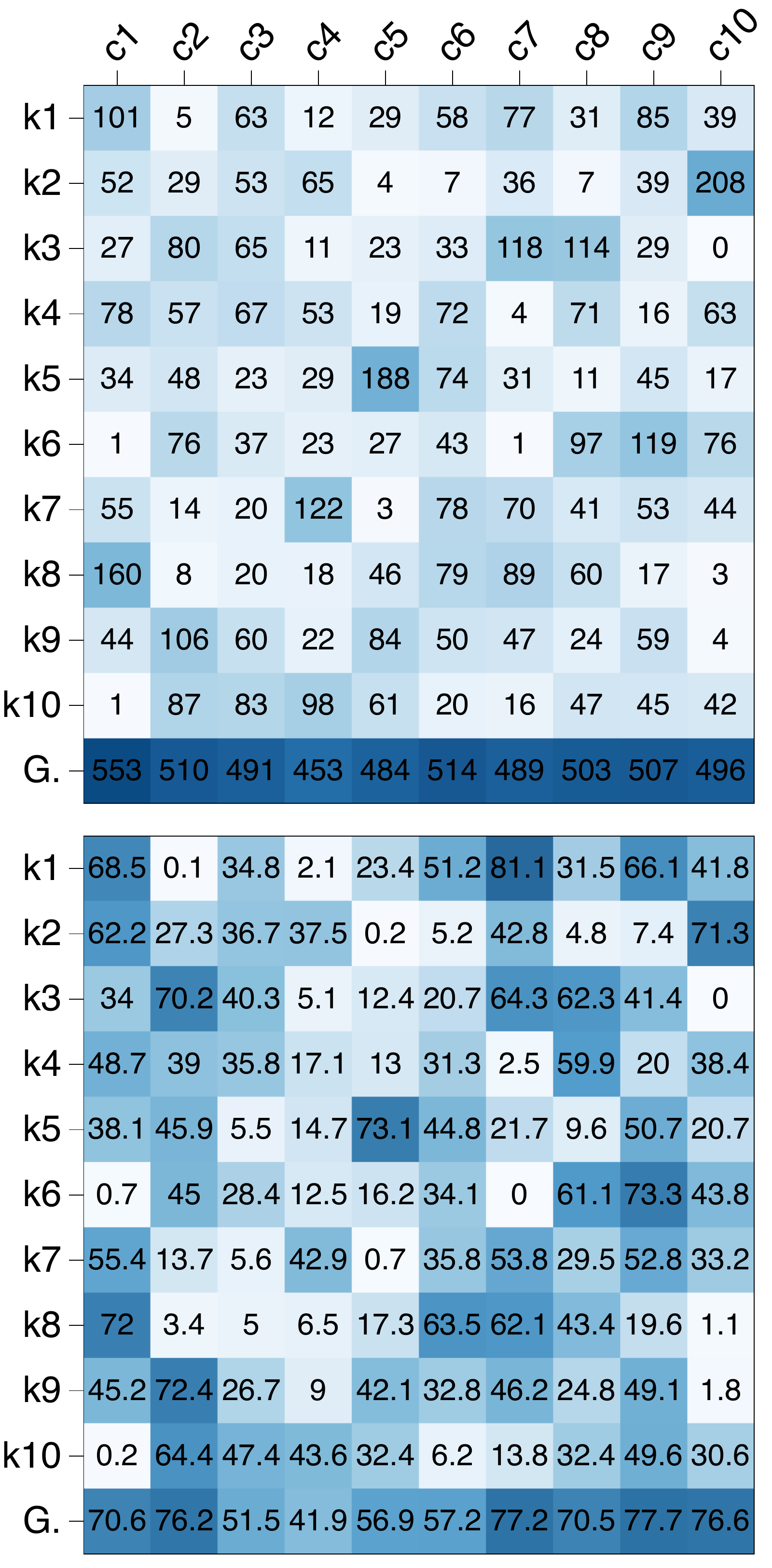}
    \caption{$\alpha$ = 1.0}       
    \end{subfigure}
    \hfill
    \begin{subfigure}[b]{0.32\linewidth}
    \centering
    \includegraphics[width=\linewidth]{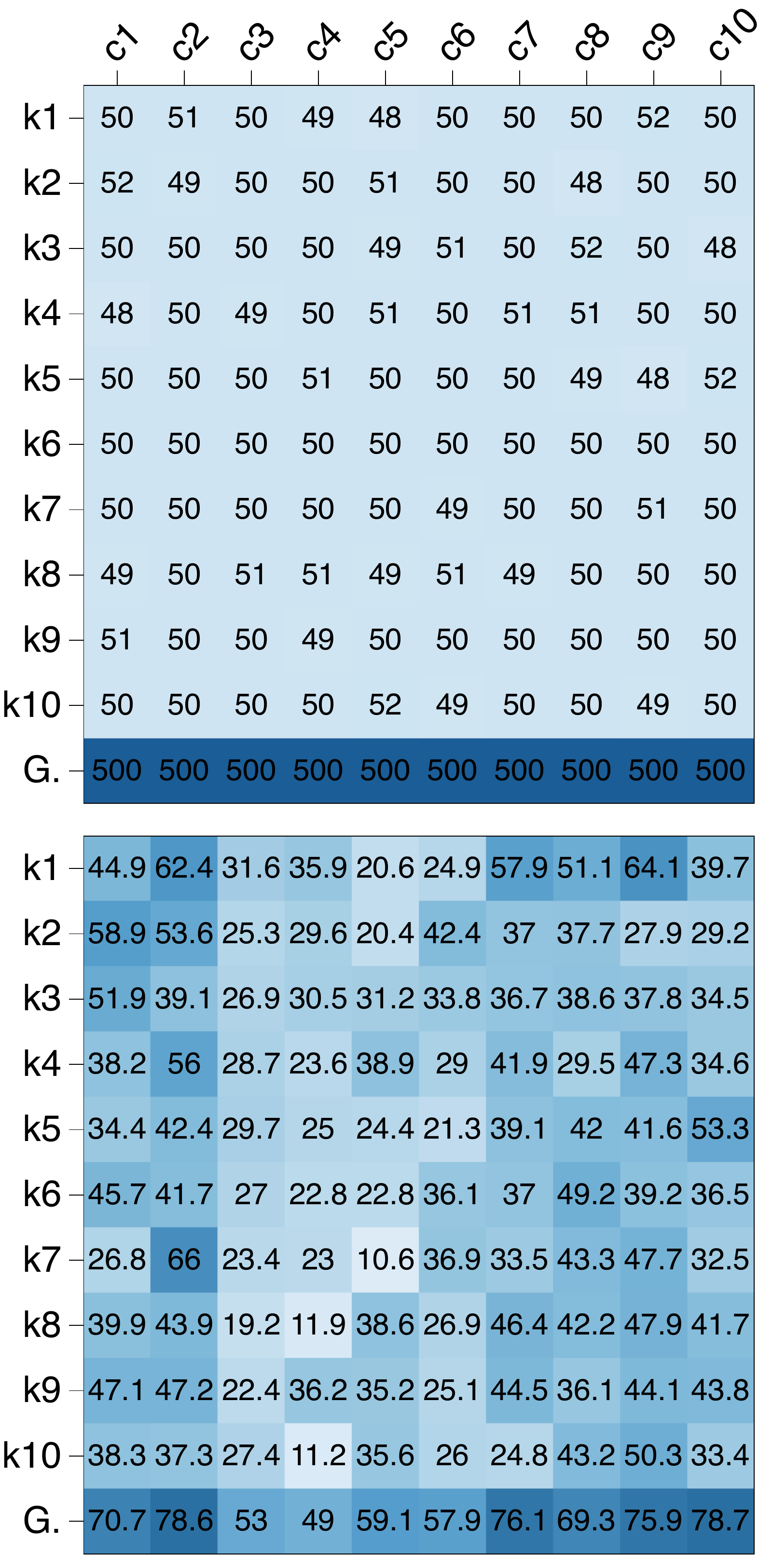}
    \caption{$\alpha$ = $\infty$}     
    \end{subfigure}
    \vspace*{-0.2cm}
\end{minipage}
\caption{Matrices of data count (top) and class-wise accuracy (down) when $\rho$ = 1.}
\vspace*{-0.2cm}
\label{fig:cnt_acc_rho1}
\end{figure*}

\vspace{-10pt}
\begin{figure*}[!h]
\centering
\begin{minipage}{0.8\linewidth}
    \centering
    \begin{subfigure}[b]{0.32\linewidth}
    \centering
    \includegraphics[width=\linewidth]{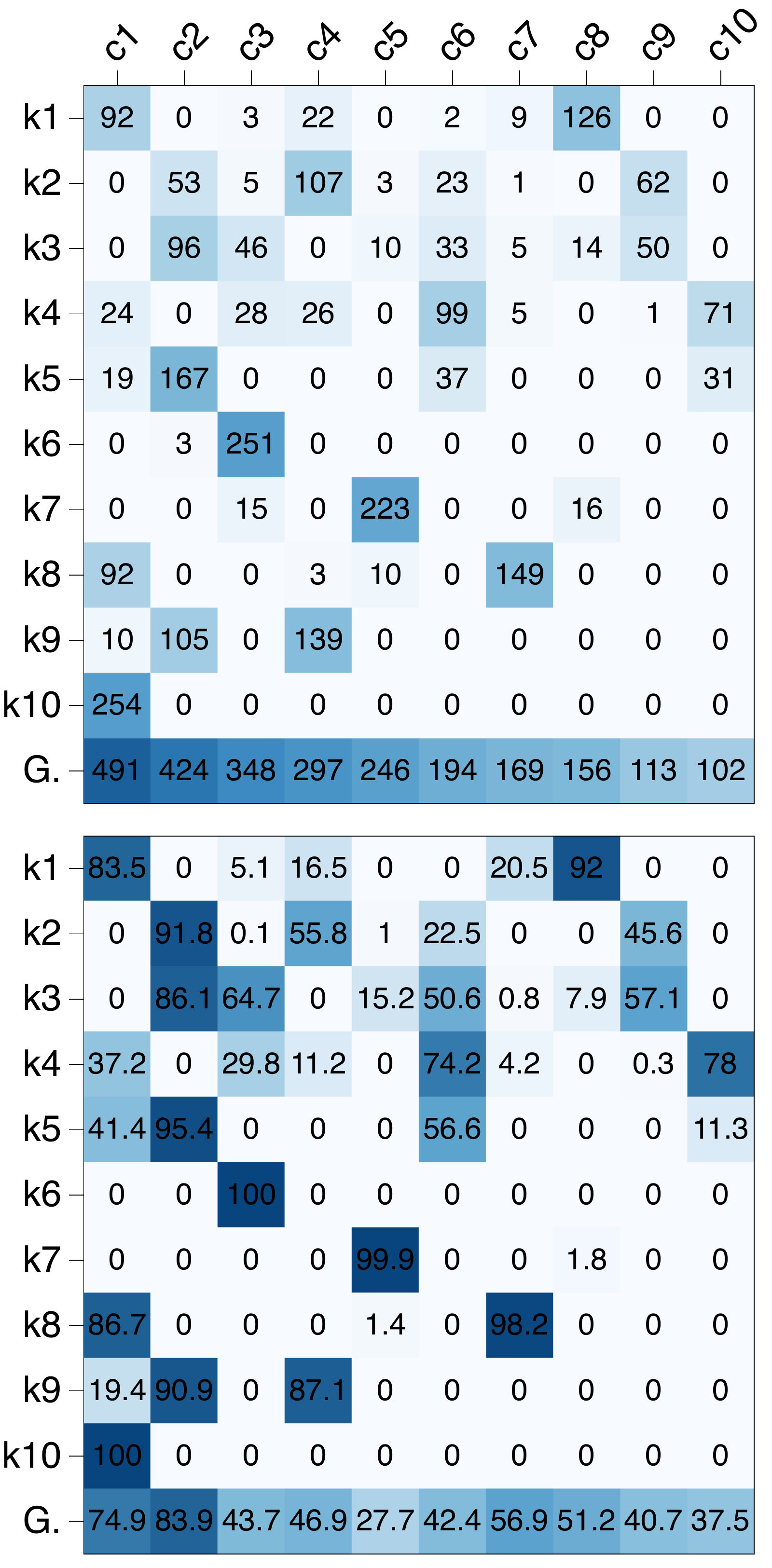}
    \caption{$\alpha$ = 0.1}         
    \end{subfigure}
    \hfill
    \centering
    \begin{subfigure}[b]{0.32\linewidth}
    \centering
    \includegraphics[width=\linewidth]{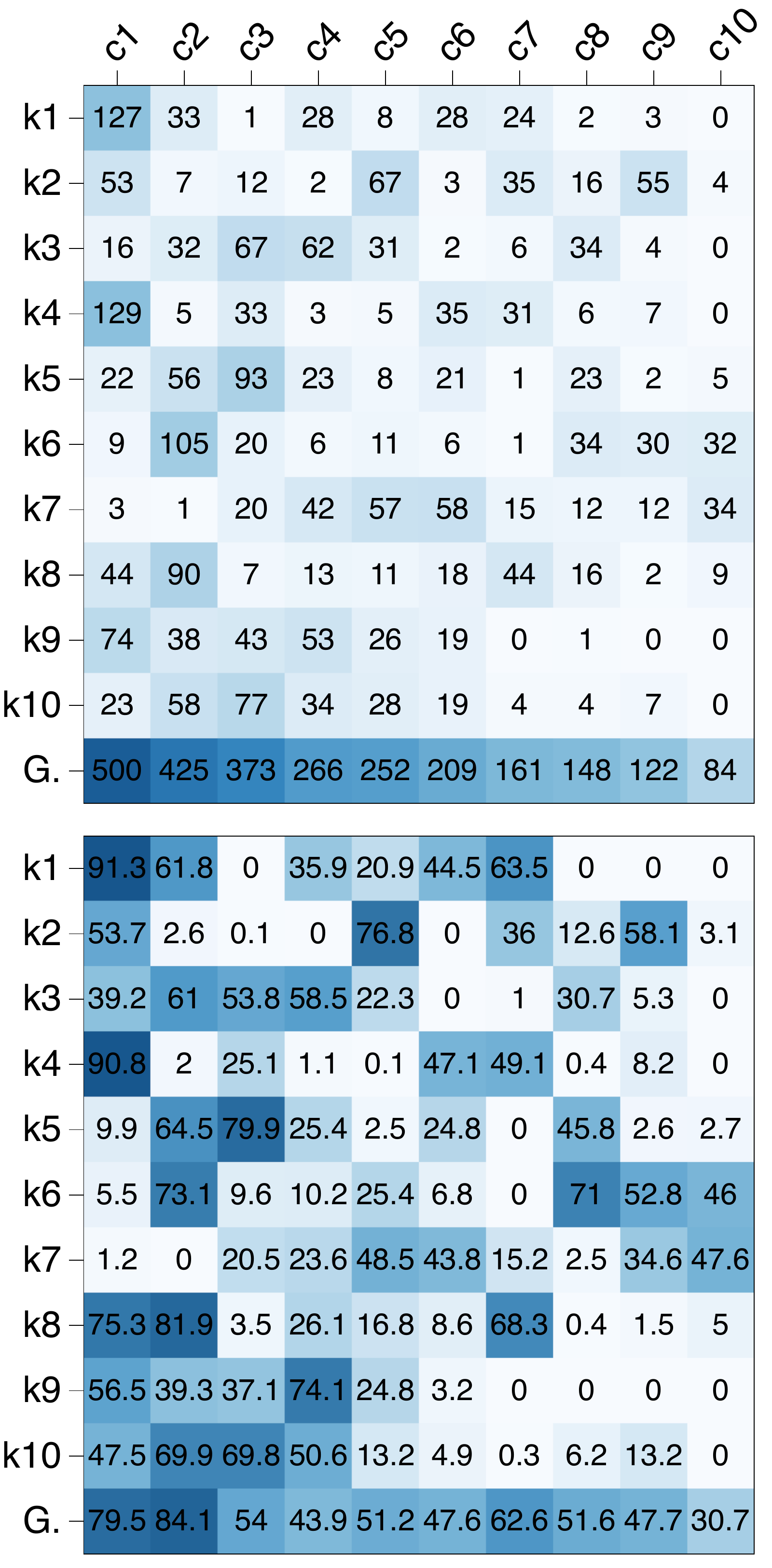}
    \caption{$\alpha$ = 1.0}       
    \end{subfigure}
    \hfill
    \begin{subfigure}[b]{0.32\linewidth}
    \centering
    \includegraphics[width=\linewidth]{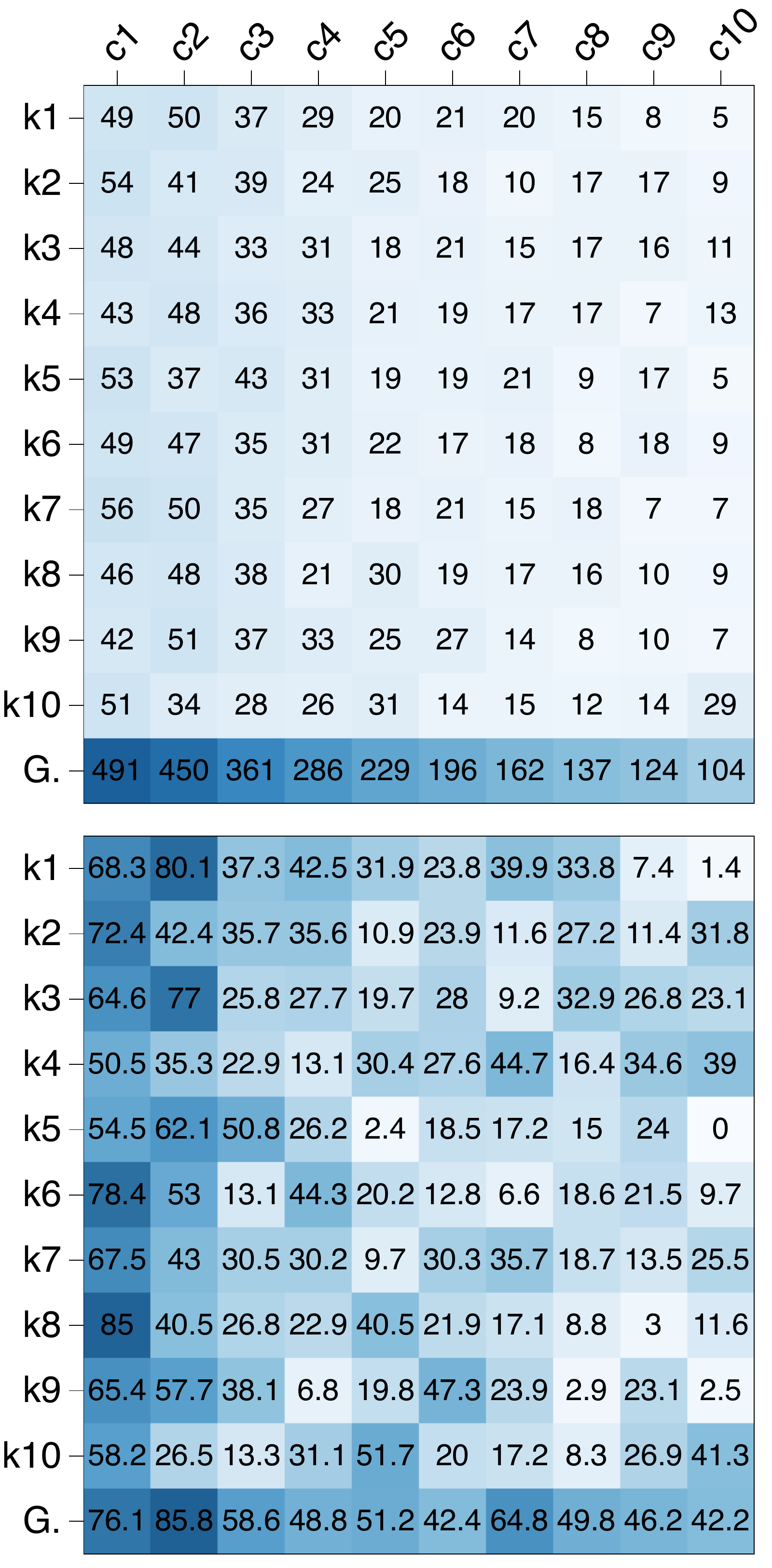}
    \caption{$\alpha$ = $\infty$}     
    \end{subfigure}
    \vspace*{-0.2cm}
\end{minipage}
\caption{Matrices of data count (top) and class-wise accuracy (down) when $\rho$ = 5.}
\vspace*{-0.2cm}
\label{fig:cnt_acc_rho5}
\end{figure*}

\newpage

\begin{figure*}[!h]
\centering
\begin{minipage}{0.8\linewidth}
    \centering
    \begin{subfigure}[b]{0.32\linewidth}
    \centering
    \includegraphics[width=\linewidth]{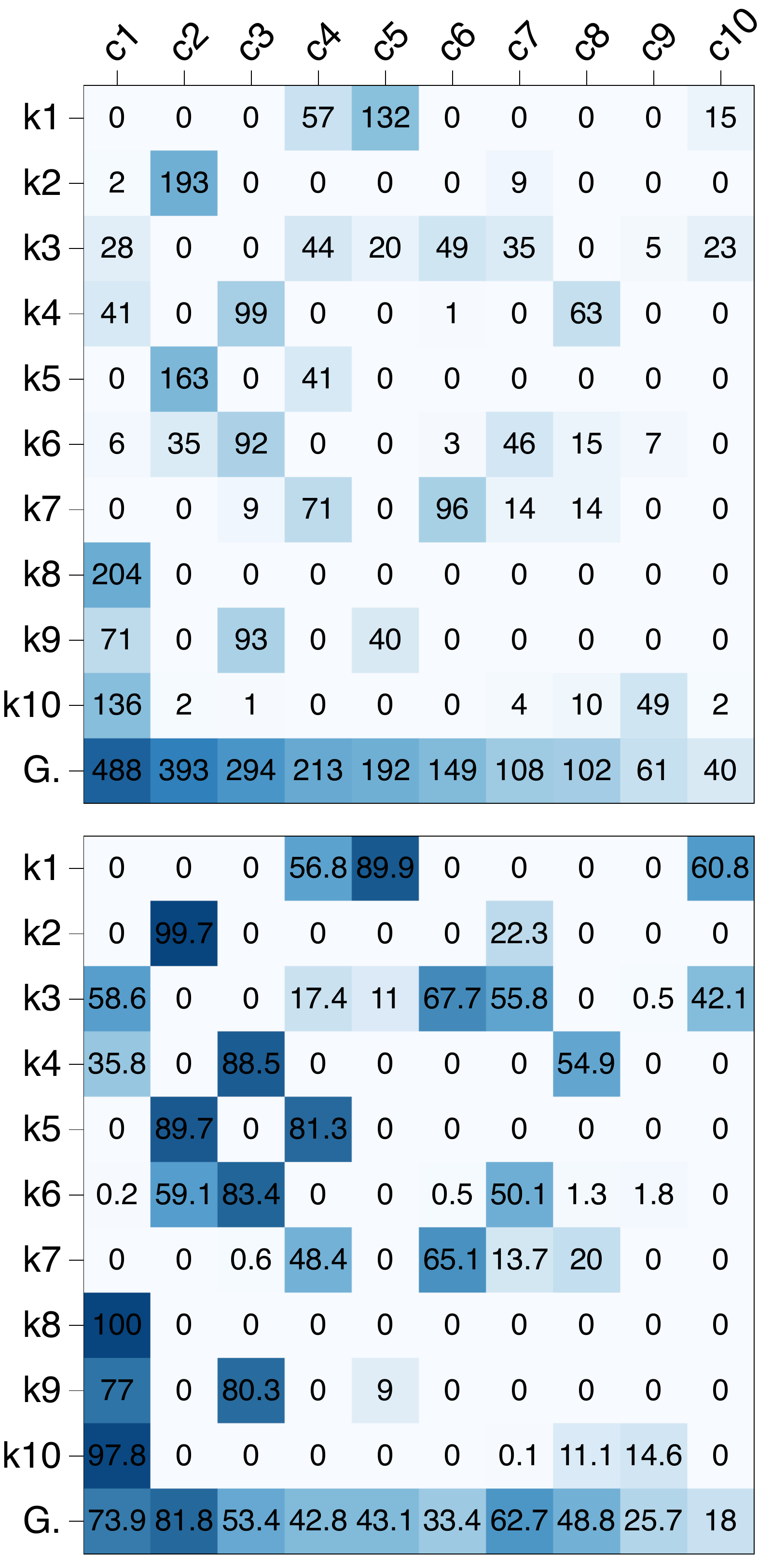}
    \caption{$\alpha$ = 0.1}         
    \end{subfigure}
    \hfill
    \centering
    \begin{subfigure}[b]{0.32\linewidth}
    \centering
    \includegraphics[width=\linewidth]{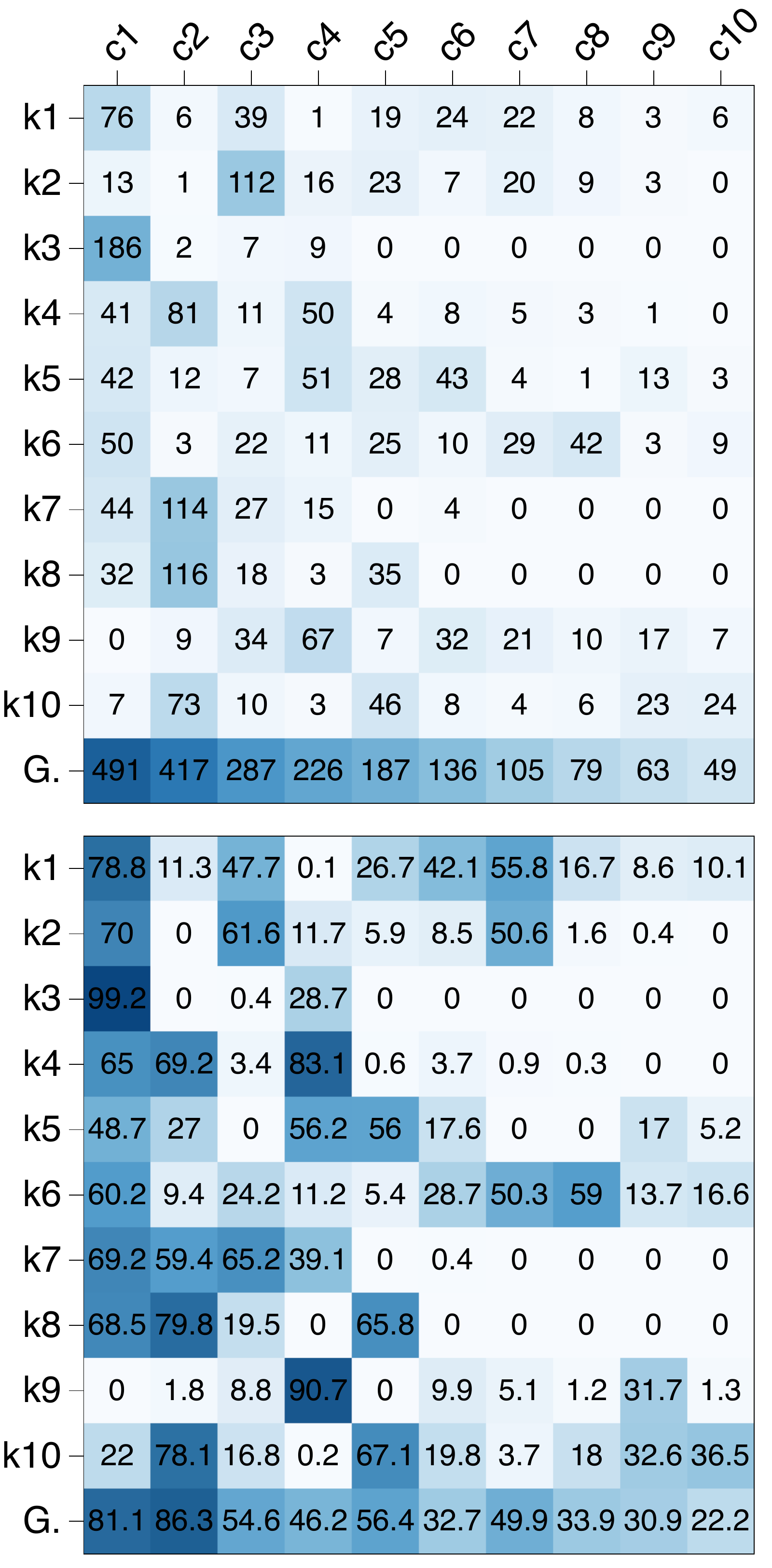}
    \caption{$\alpha$ = 1.0}       
    \end{subfigure}
    \hfill
    \begin{subfigure}[b]{0.32\linewidth}
    \centering
    \includegraphics[width=\linewidth]{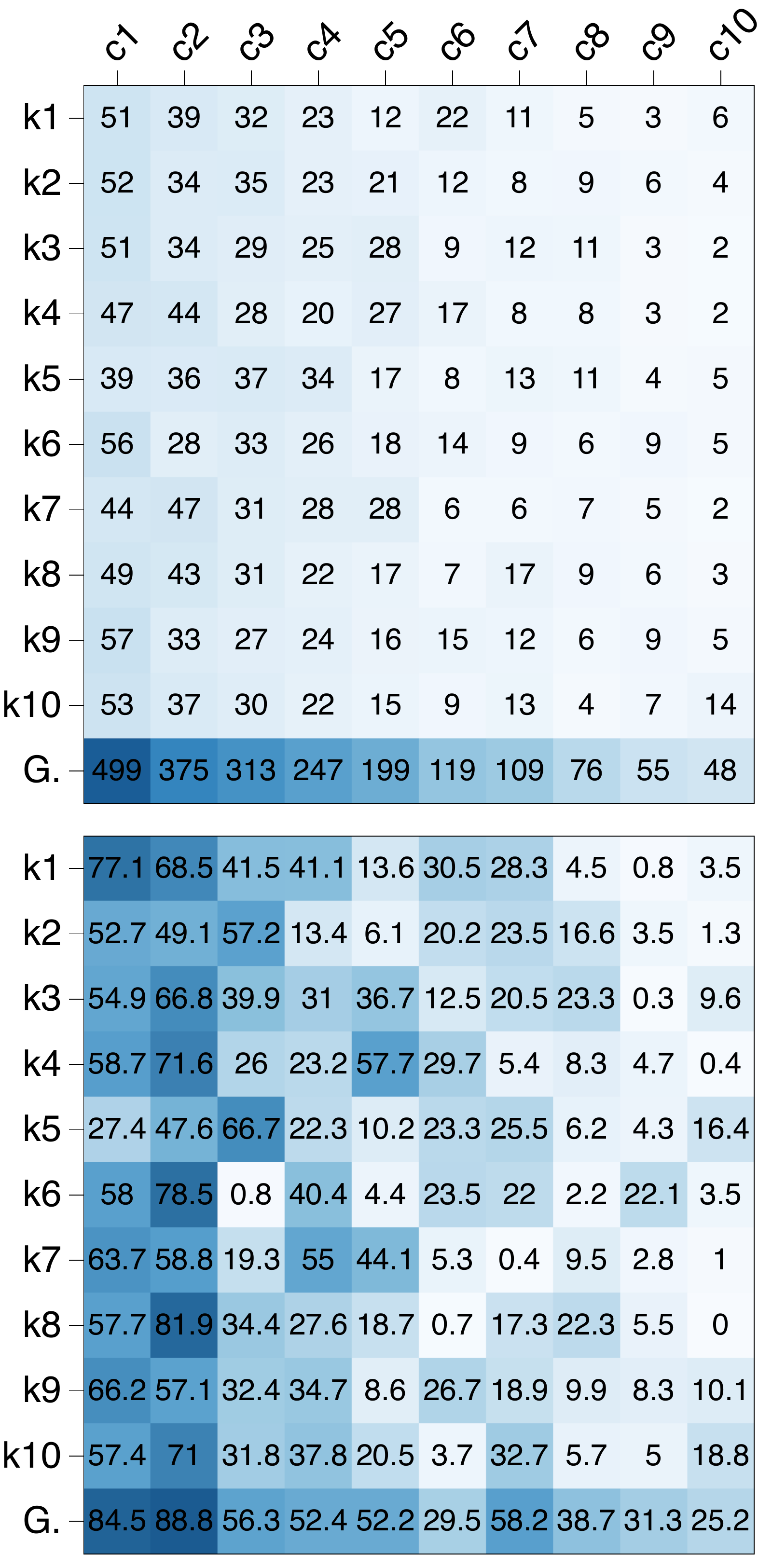}
    \caption{$\alpha$ = $\infty$}     
    \end{subfigure}
    \vspace*{-0.2cm}
\end{minipage}
\caption{Matrices of data count (top) and class-wise accuracy (down) when $\rho$ = 10.}
\vspace*{-0.2cm}
\label{fig:cnt_acc_rho10}
\end{figure*}

\begin{figure*}[!h]
\centering
\begin{minipage}{0.8\linewidth}
    \centering
    \begin{subfigure}[b]{0.32\linewidth}
    \centering
    \includegraphics[width=\linewidth]{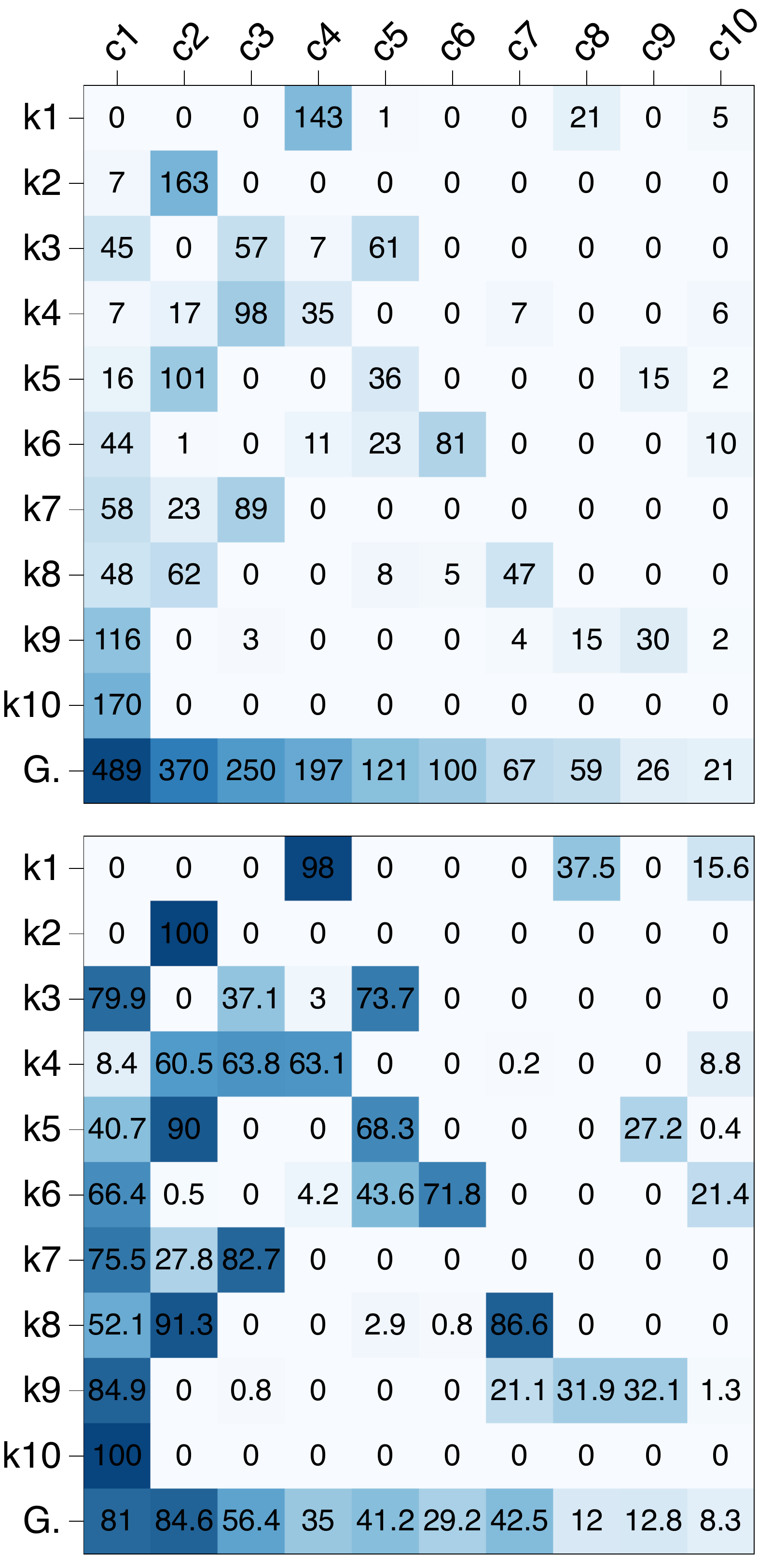}
    \caption{$\alpha$ = 0.1}         
    \end{subfigure}
    \hfill
    \centering
    \begin{subfigure}[b]{0.32\linewidth}
    \centering
    \includegraphics[width=\linewidth]{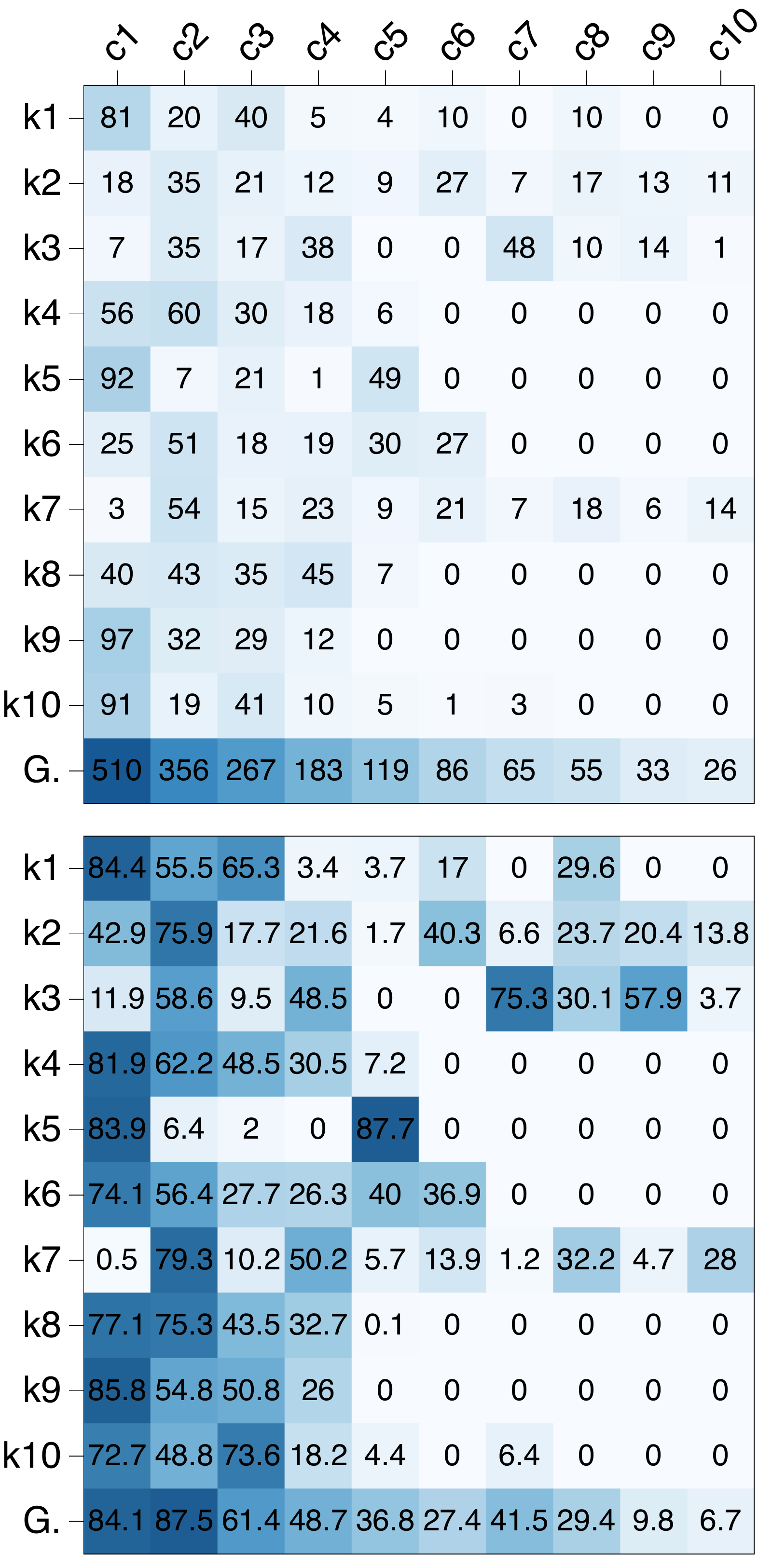}
    \caption{$\alpha$ = 1.0}       
    \end{subfigure}
    \hfill
    \begin{subfigure}[b]{0.32\linewidth}
    \centering
    \includegraphics[width=\linewidth]{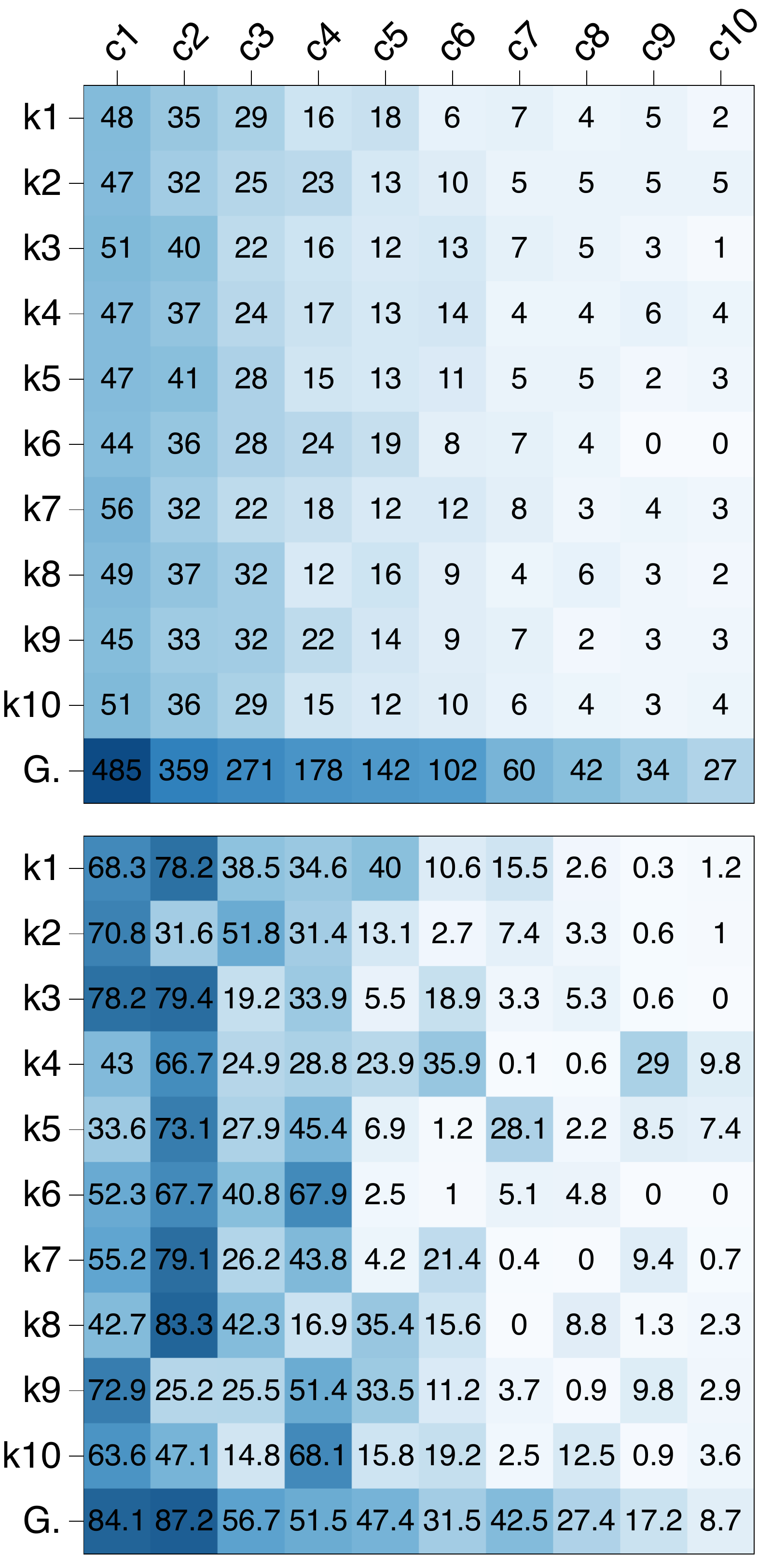}
    \caption{$\alpha$ = $\infty$}     
    \end{subfigure}
    \vspace*{-0.2cm}
\end{minipage}
\caption{Matrices of data count (top) and class-wise accuracy (down) when $\rho$ = 20.}
\vspace*{-0.2cm}
\label{fig:cnt_acc_rho20}
\end{figure*}

\newpage
\subsection{Detailed Earth Mover Distance}
\label{sec:detail_analysis_emd}
Table \ref{tab:detail_emd} \ summarizes the detailed local and global EMD for the combinations of $\rho$ = $\{$1, 5, 10, 20$\}$ and $\alpha$ = $\{$0.1, 1.0, $\infty \}$.

\begin{table}[H]
\centering
\small
\renewcommand*{\arraystretch}{0.85}
\addtolength{\tabcolsep}{1pt}
\resizebox{0.8\linewidth}{!}{
\begin{tabular}{cc|c|ccccc|ccccc}
\toprule
\multirow{2}{*}{$\rho$} &
  \multirow{2}{*}{$\alpha$} &
  \multirow{2}{*}{model} &
  \multicolumn{5}{c|}{Local EMD $(\downarrow)$} &
  \multicolumn{5}{c}{Global EMD $(\downarrow)$} \\
\cmidrule(l{2pt}r{2pt}){4-8} \cmidrule(l{2pt}r{2pt}){8-13}
                   &                           &   & 10\%  & 20\%  & 30\%  & 40\%  & 50\%  & 10\%  & 20\%  & 30\%  & 40\%  & 50\%  \\
                   \midrule
\multirow{2}{*}{1} & \multirow{2}{*}{0.1}      & G & 0.632 & 0.638 & 0.641 & 0.643 & 0.646 & 0.019 & 0.064 & 0.086 & 0.095 & 0.091 \\
                   &                           & L & 0.632 & 0.597 & 0.592 & 0.595 & 0.601 & 0.019 & 0.050 & 0.050 & 0.046 & 0.055 \\ \midrule
\multirow{2}{*}{1} & \multirow{2}{*}{1.0}      & G & 0.297 & 0.297 & 0.300 & 0.300 & 0.300 & 0.017 & 0.066 & 0.079 & 0.084 & 0.083 \\
                   &                           & L & 0.297 & 0.248 & 0.232 & 0.235 & 0.241 & 0.017 & 0.053 & 0.065 & 0.068 & 0.074 \\ \midrule
\multirow{2}{*}{1} & \multirow{2}{*}{$\infty$} & G & 0.049 & 0.077 & 0.070 & 0.065 & 0.061 & 0.014 & 0.070 & 0.066 & 0.063 & 0.060 \\
                   &                           & L & 0.049 & 0.042 & 0.054 & 0.059 & 0.066 & 0.014 & 0.025 & 0.044 & 0.053 & 0.062 \\  \midrule

\multirow{2}{*}{5} & \multirow{2}{*}{0.1}      & G & 0.662 & 0.663 & 0.666 & 0.666 & 0.669 & 0.211 & 0.201 & 0.196 & 0.194 & 0.195 \\ 
                   &                           & L & 0.662 & 0.628 & 0.627 & 0.628 & 0.634 & 0.211 & 0.232 & 0.232 & 0.236 & 0.228 \\ \midrule
\multirow{2}{*}{5} & \multirow{2}{*}{1.0}      & G & 0.402 & 0.391 & 0.387 & 0.388 & 0.389 & 0.206 & 0.188 & 0.180 & 0.173 & 0.169 \\
                   &                           & L & 0.402 & 0.309 & 0.306 & 0.306 & 0.341 & 0.206 & 0.200 & 0.201 & 0.196 & 0.196 \\ \midrule
\multirow{2}{*}{5} & \multirow{2}{*}{$\infty$} & G & 0.213 & 0.190 & 0.178 & 0.168 & 0.165 & 0.206 & 0.185 & 0.174 & 0.162 & 0.163 \\
                   &                           & L & 0.213 & 0.179 & 0.176 & 0.180 & 0.180 & 0.206 & 0.176 & 0.173 & 0.178 & 0.180 \\ \midrule

\multirow{2}{*}{10} & \multirow{2}{*}{0.1}      & G & 0.692 & 0.685 & 0.687 & 0.685 & 0.685 & 0.280 & 0.268 & 0.267 & 0.265 & 0.267 \\
                    &                           & L & 0.692 & 0.652 & 0.650 & 0.654 & 0.660 & 0.280 & 0.270 & 0.277 & 0.282 & 0.281 \\ \midrule
\multirow{2}{*}{10} & \multirow{2}{*}{1.0}      & G & 0.491 & 0.463 & 0.459 & 0.456 & 0.455 & 0.297 & 0.263 & 0.247 & 0.244 & 0.242 \\
                    &                           & L & 0.491 & 0.408 & 0.402 & 0.405 & 0.415 & 0.297 & 0.256 & 0.257 & 0.255 & 0.255 \\ \midrule
\multirow{2}{*}{10} & \multirow{2}{*}{$\infty$} & G & 0.315 & 0.240 & 0.229 & 0.223 & 0.222 & 0.303 & 0.237 & 0.226 & 0.222 & 0.221 \\
                    &                           & L & 0.315 & 0.238 & 0.237 & 0.239 & 0.240 & 0.303 & 0.237 & 0.234 & 0.237 & 0.239 \\ \midrule

\multirow{2}{*}{20} & \multirow{2}{*}{0.1}      & G & 0.692 & 0.680 & 0.676 & 0.674 & 0.677 & 0.377 & 0.300 & 0.294 & 0.294 & 0.298 \\
                    &                           & L & 0.692 & 0.641 & 0.633 & 0.636 & 0.644 & 0.377 & 0.304 & 0.326 & 0.321 & 0.323 \\ \midrule
\multirow{2}{*}{20} & \multirow{2}{*}{1.0}      & G & 0.481 & 0.455 & 0.450 & 0.448 & 0.448 & 0.374 & 0.311 & 0.300 & 0.295 & 0.292 \\
                    &                           & L & 0.481 & 0.448 & 0.437 & 0.431 & 0.437 & 0.374 & 0.354 & 0.342 & 0.303 & 0.304 \\ \midrule
\multirow{2}{*}{20} & \multirow{2}{*}{$\infty$} & G & 0.371 & 0.298 & 0.284 & 0.274 & 0.276 & 0.368 & 0.294 & 0.282 & 0.271 & 0.272 \\
                    &                           & L & 0.371 & 0.313 & 0.293 & 0.290 & 0.289 & 0.368 & 0.309 & 0.287 & 0.288 & 0.289 \\

                   \bottomrule
\end{tabular}}
    \caption{Local and global EMD on CIFAR-10 for 12 combinations of $\rho$ = $\{$1, 5, 10, 20$\}$ and $\alpha$ = $\{$0.1, 1.0, $\infty \}$.}
    \label{tab:detail_emd}
\end{table}

\clearpage
\section{Pseudo Algorithm of LoGo}
\label{sec:pseudo_algorithm}

Algorithm\,\ref{alg:logo} is the overall pipeline of the FAL framework.
Specifically, we summarize the detailed pseudocode of our \algname{} algorithm.

\begin{algorithm}[h]
\small
\caption{FAL framework with \algname{} algorithm}
\label{alg:logo}
\textbf{Input}: initialized parameter $\Theta$; unlabeled data $U^{\scaleto{1}{4pt}}$; sampling strategy $\mathcal{A}$; labeling budget $B$; clients number $K$; AL round $R$; \\
\textbf{Output}: trained parameter $\Theta^{\scaleto{R*}{4pt}}$ \\
\\
\textbf{\# Alternating AL and FL Procedure}
\begin{algorithmic}[1]
\FOR{$k=1, \dots, K$}
\STATE Randomly sample $L_{\scaleto{k}{4pt}}^{\scaleto{1}{4pt}}= \{ x_{\scaleto{1}{4pt}}, \dots, x_{\scaleto{B}{4pt}} \}$ from $U_{\scaleto{k}{4pt}}^{\scaleto{1}{4pt}}$, and $U_k^{2} = U_k^1 \setminus L_k^1$
\STATE Get the labeled set $D_{\scaleto{k}{4pt}}^{\scaleto{1}{4pt}}$\, from the oracles
\ENDFOR
\STATE $\Theta^{\scaleto{1*}{4pt}}=$\,\texttt{FedAvg}\,($\Theta$, $D^{\scaleto{1}{4pt}}, K$) \\
\, \\
\FOR{$r=2, \dots, R$}
\FOR{$k=1, \dots, K$}
\STATE $D^{r}_k, \,U_k^{r+1}=$\,\,\texttt{LoGo}\,($\Theta^{(r-1)*}$, $D^{r-1}_{\scaleto{k}{4pt}}, U_k^r$)
\ENDFOR
\STATE $\Theta^{r*}=$\,\texttt{FedAvg}\,($\Theta$, $D^{r}, K$)
\ENDFOR
\end{algorithmic}
\, \\
\textbf{Function}\,\,\texttt{LoGo}:
\begin{algorithmic}[1] 
\STATE \textbf{\# Macro Step}
\STATE 
Train a local-only model $\Theta^{(r-1)}_{k*}$ from the scratch only using $D_k^{r-1}$
\STATE  For each $x\in U_k^r$, calculate the gradient embedding $g_{\hat{y}}^x$  by Eq.\,\eqref{eq:gradient}
\STATE Cluster $U_k^r$ into $B$ clusters($\mathcal{C}_1,...,\mathcal{C}_B$) by Eq.\,\eqref{eq:kmeans} \\
\, \\
\STATE \textbf{\# Micro Step}
\STATE 
$L_k^r= \emptyset $
\FOR{$\mathcal{C}_{\scaleto{i}{4pt}}=\mathcal{C}_{\scaleto{1}{4pt}}, \dots, \mathcal{C}_{\scaleto{B}{4pt}}$}
\STATE $L_k^r = L_k^r \cup \{ \mathcal{A}(\mathcal{C}_{\scaleto{i}{4pt}}, \Theta^{(r-1)*}, 1) \}$
\STATE $D_k^r = D_k^{r-1} \cup D_k^r$\, and\, $U_{\scaleto{k}{4pt}}^{\scaleto{r+1}{4pt}} = U_{\scaleto{k}{4pt}}^{\scaleto{r}{3pt}} \setminus L_k^r$
\ENDFOR 
\STATE \textbf{return}  $D_k^r$, \,$U_k^{r+1}$
\end{algorithmic}
\, \\
\textbf{Function}\,\,\texttt{FedAvg}:
\begin{algorithmic}[1]
\FOR{$\,FL\,\,round$\,}
\STATE Distribute $\Theta$ to the all client
\FOR{$k = 1, \dots, K$}
\STATE Train $\Theta_k$ on $D_k^r$ by minimizing $\mathbb{E}_{D_k^r}[\ell(x,y; \Theta_k)]$
\ENDFOR
\STATE $\Theta = (\sum_k \Theta_k) / K$
\ENDFOR
\STATE \textbf{return} $\Theta$
\end{algorithmic}

\end{algorithm}

\clearpage
\section{Experimental Settings}
\label{sec:exp_settings}

\subsection{Datasets}
We mainly experimented on two natural image datasets (CIFAR-10\footnote{https://www.cs.toronto.edu/~kriz/cifar.html}, SVHN\footnote{http://ufldl.stanford.edu/housenumbers}) and three medical image datasets\footnote{https://medmnist.com/} (PathMNIST, DermaMNIST, OrganAMNIST). 
Table \ref{tab:dataset_summray} provides a summary of the five datasets.
For the details of partitioning data to each client, please refer to Appendix\,\ref{sec:dataset_summary}.

\newcolumntype{L}[1]{>{\raggedright\let\newline\\\arraybackslash\hspace{0pt}}m{#1}}
\newcolumntype{X}[1]{>{\centering\let\newline\\\arraybackslash\hspace{0pt}}p{#1}}
\begin{table}[h!]
\centering
\footnotesize
\begin{tabular}{c|c|c|c|c|c}
\toprule
                        & Dataset     & \# of Train & \# of Test & \# of Classes & $\rho$  \\ 
\midrule
\multirow{2}{*}{Natural} & CIFAR-10    & 50,000      & 10,000     & 10  & 1.0            \\ 
                         & SVHN        & 73,257      & 26,032     & 10  & 2.97            \\ 
                         
\midrule
\multirow{3}{*}{Medical} & \scriptsize PathMNIST   & 89,996      & 7,180      & 9  &      1.63     \\ 
                         &  \scriptsize DermaMNIST  & 7,007       & 2,005      & 7  &     58.66      \\ 
                         & \scriptsize OrganAMNIST & 34,581      & 17,778     & 11  &     4.54       \\

\bottomrule
\end{tabular}
\caption{Summary of benchmark datasets.}
\label{tab:dataset_summray}
\end{table}

\subsection{Implementation Details}
For the FL training pipeline, we set the number of FL rounds to 100 and local update epochs to 5.
We used a SGD optimizer with the initial learning rate of 0.01 and the momentum of 0.9.
The learning rate was decayed by 0.1 at half and three-quarters of federated learning rounds to ensure convergence, and we used a random horizontal flipping as data augmentation.
For training local-only models, we trained the model using the aforementioned settings for 50 epochs. However, the training was terminated if the training accuracy reached 99\%.
It should be noted that we averaged the classification accuracy of the last 5 epochs in each round and repeated all experiments with four different seeds.
All algorithms were implemented using PyTorch 1.11.0 and executed using NVIDIA RTX 3080 GPUs.

\subsection{Experimental Categories}
A total of six categories were considered in the evaluation:
\begin{enumerate}
\item {`Query selector'} of whether to use a local-only or global model with the six compared strategies. 
\item  {`Heterogeneity level'} of varying degree of class imbalance. We adopt a Latent Dirichlet Allocation (LDA) \cite{moon} strategy. 
For example, the smaller $\alpha$, the more heterogeneous the data distribution. 
\item  {`Imbalance ratio'} of used datasets. 
We classified five datasets for evaluation based on the imbalance ratio $\rho$.
CIFAR-10 and PathMNIST belong to a low imbalance ratio ($\rho<2$), and SVHN, DermaMNIST, and OrganAMNIST belong to a high imbalance ratio ($\rho\geq 2$).
\item  {`Model architecture.'} We employed four layers of convolution neural network for a base architecture and also experimented with ResNet-18 \cite{resnet} and MobileNet \cite{mobilenet}. 
\item  `{Budget size}' for labeling. We tested small (1\%), medium (5\%), and large (20\%) budget sizes for each round. 
\item  {`Model initialization'} of either learning from scratch (random) or from the checkpoint of the previous AL round (continue). \looseness=-1
\end{enumerate}

\newpage
\subsection{Combination of experimental settings}
We compared our algorithms and baselines in 38 comprehensive experimental settings, which are the combinations of the aforementioned six categories.
All the experimental combinations we performed are summarized in Table \ref{tab:setting_summary}.

\begin{table*}[h!]
\centering
\small
\renewcommand*{\arraystretch}{1}
\addtolength{\tabcolsep}{1pt}
\resizebox{0.75\linewidth}{!}{
\begin{tabular}{cccccc}
\toprule
Query Selelctor   & Dir($\alpha$) & Data Type     & Model Arch.   & Budget Size & Model Init. \\ 
\midrule
Global          & 0.1 & CIFAR-10     & 4CNN   & 5\%         & Random     \\ 
Global          & 0.1 & SVHN        & 4CNN   & 5\%         & Random     \\ 
Global          & 0.1 & PathMNIST   & 4CNN   & 5\%         & Random     \\ 
Global        & 0.1  & OrganAMNIST & 4CNN & 5\%         & Random     \\ 
Global        & 0.1  & DermaMNIST & 4CNN   & 5\%         & Random     \\ 
Global        & 1   & CIFAR-10      & 4CNN   & 5\%         & Random     \\ 
Global        & 1   & SVHN         & 4CNN    & 5\%         & Random     \\ 
Global        & $\infty$   & CIFAR-10     & 4CNN   & 5\%         & Random     \\ 
Global        & $\infty$   & SVHN        & 4CNN   & 5\%         & Random     \\ 
Global        & 0.1  & CIFAR-10      & 4CNN  & 5\%         & Continue   \\ 
Global        & 0.1  & SVHN         & 4CNN   & 5\%         & Continue   \\ 
Global        & 0.1  & CIFAR-10      & ResNet-18 & 5\%         & Random     \\ 
Global      & 0.1    & SVHN       & ResNet-18  & 5\%         & Random     \\ 
Global        & 0.1  & CIFAR-10      & MobileNet  & 5\%         & Random     \\ 
Global        & 0.1  & SVHN         & MobileNet  & 5\%         & Random     \\ 
Global        & 0.1  & CIFAR-10      & 4CNN   & 1\%         & Random     \\ 
Global        & 0.1  & SVHN         & 4CNN   & 1\%         & Random     \\ 
Global        & 0.1  & CIFAR-10      & 4CNN   & 20\%        & Random     \\ 
Global        & 0.1  & SVHN         & 4CNN   & 20\%        & Random     \\ 
Local-only    & 0.1  & CIFAR-10      & 4CNN  & 5\%         & Random     \\ 
Local-only    & 0.1  & SVHN         & 4CNN   & 5\%         & Random     \\ 
Local-only    & 0.1  & PathMNIST    & 4CNN   & 5\%         & Random     \\ 
Local-only    & 0.1  & OrganAMNIST  & 4CNN   & 5\%         & Random     \\ 
Local-only   & 0.1   & DermaMNIST   & 4CNN & 5\%         & Random     \\ 
Local-only      & 1  & CIFAR-10    & 4CNN    & 5\%         & Random     \\ 
Local-only     & 1   & SVHN       & 4CNN   & 5\%         & Random     \\ 
Local-only     & $\infty$  & CIFAR-10     & 4CNN   & 5\%         & Random     \\ 
Local-only     & $\infty$  & SVHN       & 4CNN  & 5\%         & Random     \\ 
Local-only      & 0.1 & CIFAR-10    & 4CNN   & 5\%         & Continue   \\ 
Local-only     & 0.1  & SVHN        & 4CNN  & 5\%         & Continue   \\ 
Local-only      & 0.1 & CIFAR-10    & ResNet-18    & 5\%         & Random     \\ 
Local-only     & 0.1  & SVHN        & ResNet-18 & 5\%         & Random     \\ 
Local-only     & 0.1  & CIFAR-10     & MobileNet  & 5\%         & Random     \\ 
Local-only     & 0.1  & SVHN         & MobileNet  & 5\%         & Random     \\ 
Local-only     & 0.1  & CIFAR-10      & 4CNN  & 1\%         & Random     \\ 
Local-only     & 0.1  & SVHN         & 4CNN   & 1\%         & Random     \\ 
Local-only     & 0.1  & CIFAR-10      & 4CNN   & 20\%        & Random     \\ 
Local-only     & 0.1  & SVHN        & 4CNN   & 20\%        & Random     \\ 
\bottomrule
\end{tabular}}
\caption{Summary of the entire experimental combinations.}
\label{tab:setting_summary}
\end{table*}

\clearpage
\section{Computational Cost of Query Selection}
\label{sec:computational_cost}
In Table\,\ref{tab:time_cost}, we measured the wallclock time for various combinations of the algorithm, query selector, and labeling ratio.
We confirmed that as the percentage of labeled data increases, the time required to measure the importance score with the global model decreases due to the reduced amount of unlabeled data.
Conversly, the local-only model takes more time as it requires training on a larger number of labeled samples.
Our LoGo algorithm shows a comparable computational cost to the baselines that use the local-only model\,(L) for query selection.
Note that we used a simple Entropy sampling within LoGo algorithm to measure the uncertainty, and the only possible bottleneck is \textit{k}-means clustering in the Macro step.

\begin{table}[h!]
\centering
\addtolength{\tabcolsep}{-1pt}
\renewcommand*{\arraystretch}{1.05}
\resizebox{0.7\linewidth}{!}{
\begin{tabular}{l|cc|cc|cc|cc|cc|c}
\toprule
                   & \multicolumn{2}{c|}{\!\!\!Entropy\!\!\!} & \multicolumn{2}{c|}{Coreset} & \multicolumn{2}{c|}{BADGE} & \multicolumn{2}{c|}{GCNAL} & \multicolumn{2}{c|}{ALFA-Mix} & LoGo \\
 \cmidrule(l{2pt}r{2pt}){2-3} \cmidrule(l{2pt}r{2pt}){4-5} \cmidrule(l{2pt}r{2pt}){6-7} \cmidrule(l{2pt}r{2pt}){8-9} \cmidrule(l{2pt}r{2pt}){10-11} \cmidrule(l{2pt}r{2pt}){12-12}
\multirow{-2.5}{*}{Query ratio} & G & L & G & L & G & L & G & L & G & L & G,\,L \\ \midrule
5\%\,$\rightarrow$\,10\% & 5.99 \!  & \! 8.85 & 7.32 & 10.24 & 14.43 \!  & \! 17.36 & 8.20 & 11.13 & 13.88 \! & 20.87 & 17.10 \\ \hline
40\%\,$\rightarrow$\,45\% & 4.17 \!  & \! 33.59 & 7.02 & 33.99 & 10.01 \!  & \! 39.11 & 8.11 & 35.46 & 11.94 \! & 41.99 & 37.42 \\ \hline
75\%\,$\rightarrow$\,80\% & 3.95 \!  & \! 59.57 & 6.72 & 58.98 & 3.95 \!  & \! 62.62 & 7.71 & 60.26 & 10.46 \! & 65.16 & 56.81 \\
\bottomrule
\end{tabular}
}
\caption{Computational cost on CIFAR-10 with 4 layers of CNN. We averaged the query selection time\,(sec.) of all 10 clients, measured on a RTX 3090 GPU.}
\label{tab:time_cost}
\end{table}

\section{LoGo with Various FL Methods}
\label{sec:various_fl_algo}
We have further experimented with two federated learning algorithms, FedProx\cite{fedprox} and SCAFFOLD\cite{scaffold}, in conjunction with AL strategies.
Specifically, we compared our LoGo with baselines that demonstrated Top-1 or Top-2 performance more than once in Table\,\ref{tab:acc_comparision}. The experimental configurations are same to those used in Table\,\ref{tab:acc_comparision}.
As summarized in Table\,\ref{tab:compare_fedprox_scaffold}, LoGo consistently outperforms the baselines for both federated learning algorithms.
This observation suggests that LoGo is an orthogonal selection algorithm that can be integrated with any federated learning algorithm, having potential to improve the performance in various applications.

\begin{table}[h]
    \small
    \centering
    \renewcommand*{\arraystretch}{0.97}
    \resizebox{0.7\linewidth}{!}{
    \begin{tabular}{l|l|c|ccc|ccc}
        \toprule
         & &  & \multicolumn{3}{c|}{CIFAR-10} & \multicolumn{3}{c}{SVHN} \\
         \cmidrule(l{2pt}r{2pt}){4-6} \cmidrule(l{2pt}r{2pt}){7-9}
        \multirow{-2.5}{*}{FL\,algo.} & \multirow{-2.5}{*}{Method} & \multirow{-2.5}{*}{Model} & 20\% & 40\% & 60\% & 20\% & 30\% & 40\% \\
        \midrule
         \multirow{7}{*}{FedProx} & & G & 62.89 & 67.52 & 70.38 & 82.22 & 84.34 & 85.42  \\
         & \multirow{-2}{*}{Entropy} & L  & \underline{65.72} & \underline{70.57} & \underline{72.42} &82.08 & 83.73 & 85.30  \\
         \cline{2-9}
         & & G & 64.16 & 68.62 & 70.82 & \underline{83.09} & \textbf{84.65} & 85.84  \\
         & \multirow{-2}{*}{BADGE} & L  & 65.54 & 70.56 & 72.30 &81.99 & 84.17 & 85.17  \\
         \cline{2-9}
         & & G & 63.77 & 68.34 & 70.78 & 82.63 & 84.48 & \underline{85.94}  \\
         & \multirow{-2}{*}{ALFA-Mix} & L  & 63.44 & 67.83 & 70.31 &80.71 & 82.81 & 84.22  \\
         \cline{2-9}
          & \textbf{LoGo} & G, L & \textbf{65.79} & \textbf{70.61} & \textbf{72.61} & \textbf{83.12} & \underline{84.61} & \textbf{86.09}  \\
         \midrule
         \multirow{7}{*}{SCAFFOLD} & & G & 65.58 & 70.37 & 72.52 & 82.75 & 85.69 & 86.48  \\
         & \multirow{-2}{*}{Entropy} & L  & 67.96 & \underline{72.67} & \underline{74.06} & 83.24 & 84.30 & 85.82  \\
         \cline{2-9}
         & & G & 66.33 & 70.68 & 72.79 & 83.80 & 84.72 & \textbf{86.93}  \\
         & \multirow{-2}{*}{BADGE} & L  & \underline{68.27} & 72.52 & 73.79 & 83.40 & 84.61 & 86.16  \\
         \cline{2-9}
         & & G & 66.11 & 70.50 & 72.55 & \underline{84.11} & \textbf{85.72} & 86.14  \\
         & \multirow{-2}{*}{ALFA-Mix} & L  & 66.11 & 70.00 & 71.91 & 82.15 & 82.89 & 84.74  \\
         \cline{2-9}
          & \textbf{LoGo} & G, L & \textbf{68.33} & \textbf{72.77} & \textbf{74.48} & \textbf{84.29} & \underline{85.70} & \underline{86.73}  \\
         \bottomrule
    \end{tabular}}
    \caption{Classification accuracy on two benchmarks with FedProx\,($\mu$\,=\,0.01) and SCAFFOLD. We compared to three overwhelming baselines and averaged three random seeds. \textbf{Bold} and \underline{underline} mean Top-1 and Top-2, respectively.}
    \label{tab:compare_fedprox_scaffold}
    \vspace{-0.37cm}
\end{table}

\clearpage
\section{Detailed Experimental Results}
\label{sec:exp_detail_results}

\begin{figure*}[htb]
\centering
    \begin{subfigure}[b]{0.3\linewidth}
    \raggedleft
    \includegraphics[width=\linewidth]{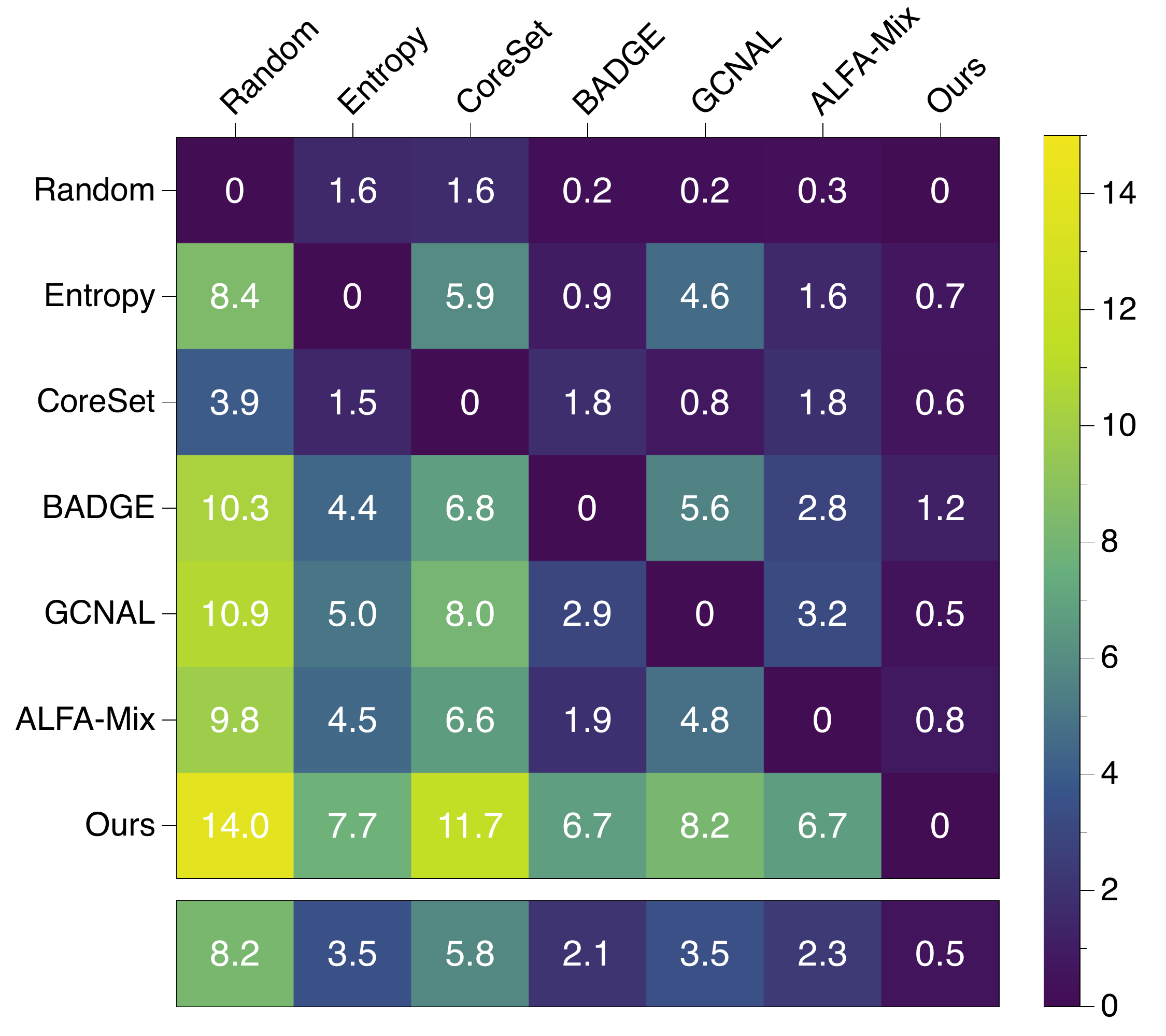}
    \caption{Global}
    \end{subfigure}
    \hspace{10pt}
    \begin{subfigure}[b]{0.3\linewidth}
    \raggedright
    \includegraphics[width=\linewidth]{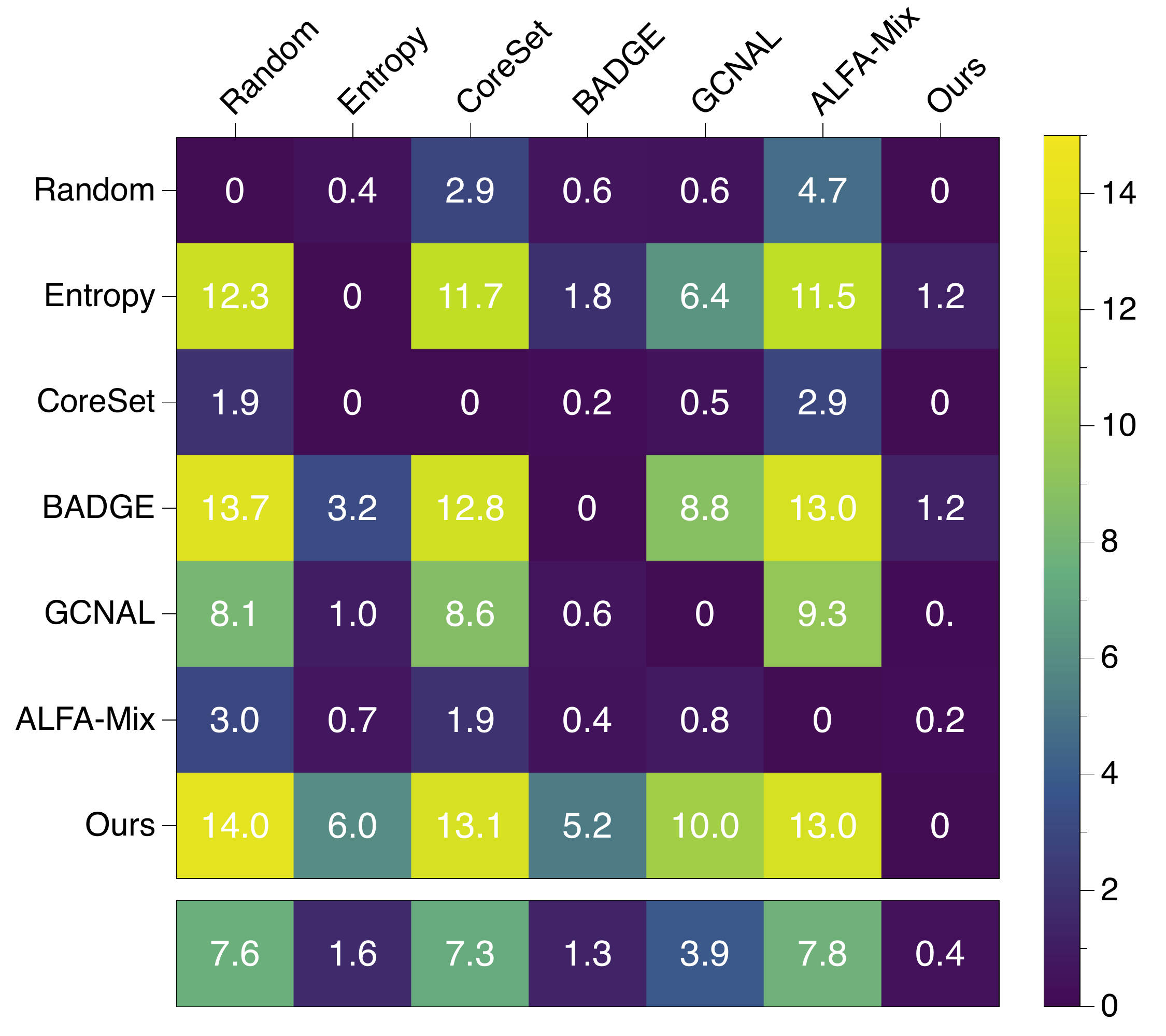}
    \caption{Local-only}
    \end{subfigure}
    \caption{Pairwise penalty matrix for a query selector category. The maximum value of both matrices is 19.}
    \label{fig:comp_selector}
\end{figure*}

\begin{figure*}[htb]
    \centering
    \begin{subfigure}[b]{0.3\linewidth}
    \includegraphics[width=\linewidth]{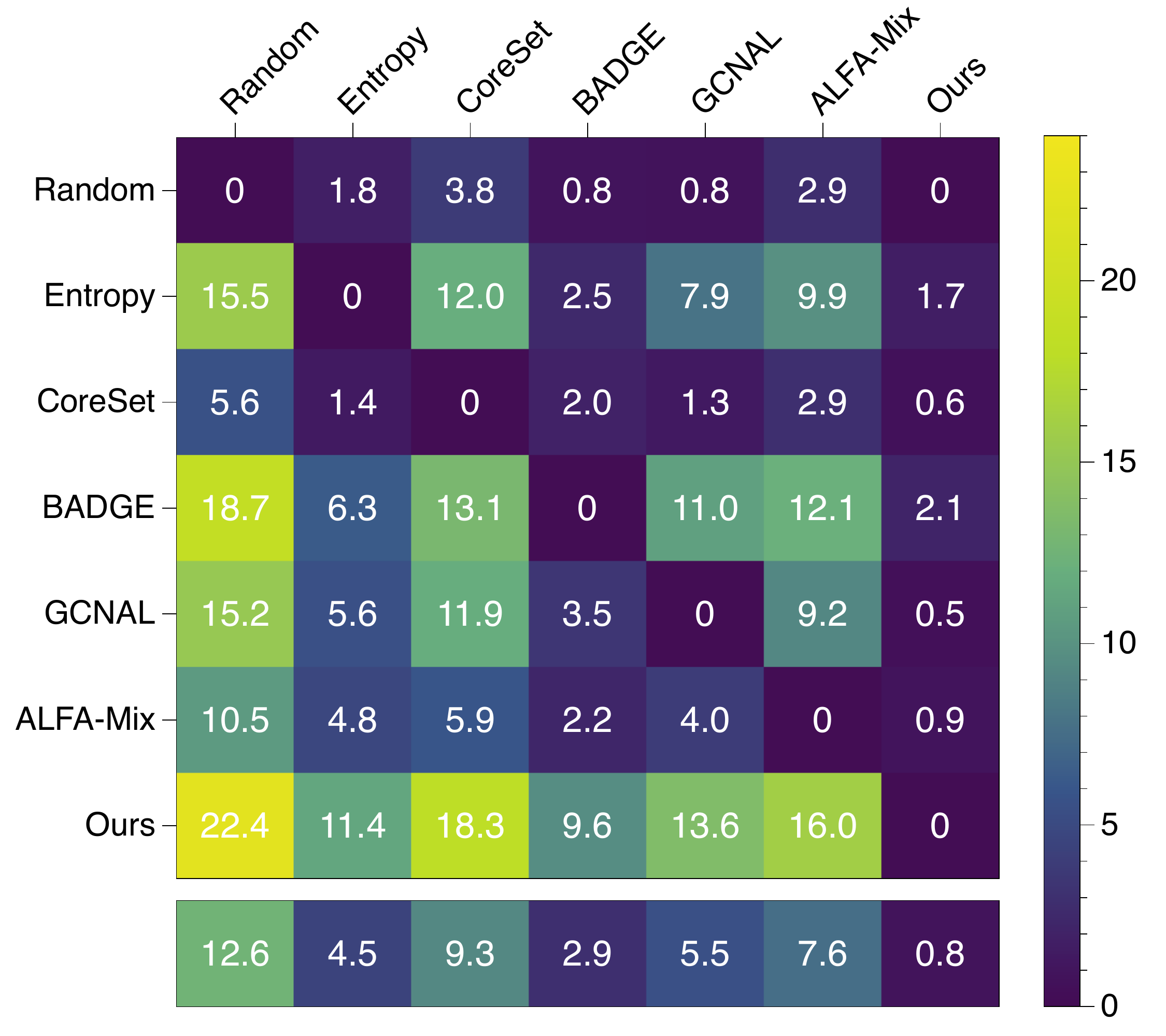}
    \caption{$\alpha = 0.1$}
    \end{subfigure}
    \hfill
    \begin{subfigure}[b]{0.3\linewidth}
    \includegraphics[width=\linewidth]{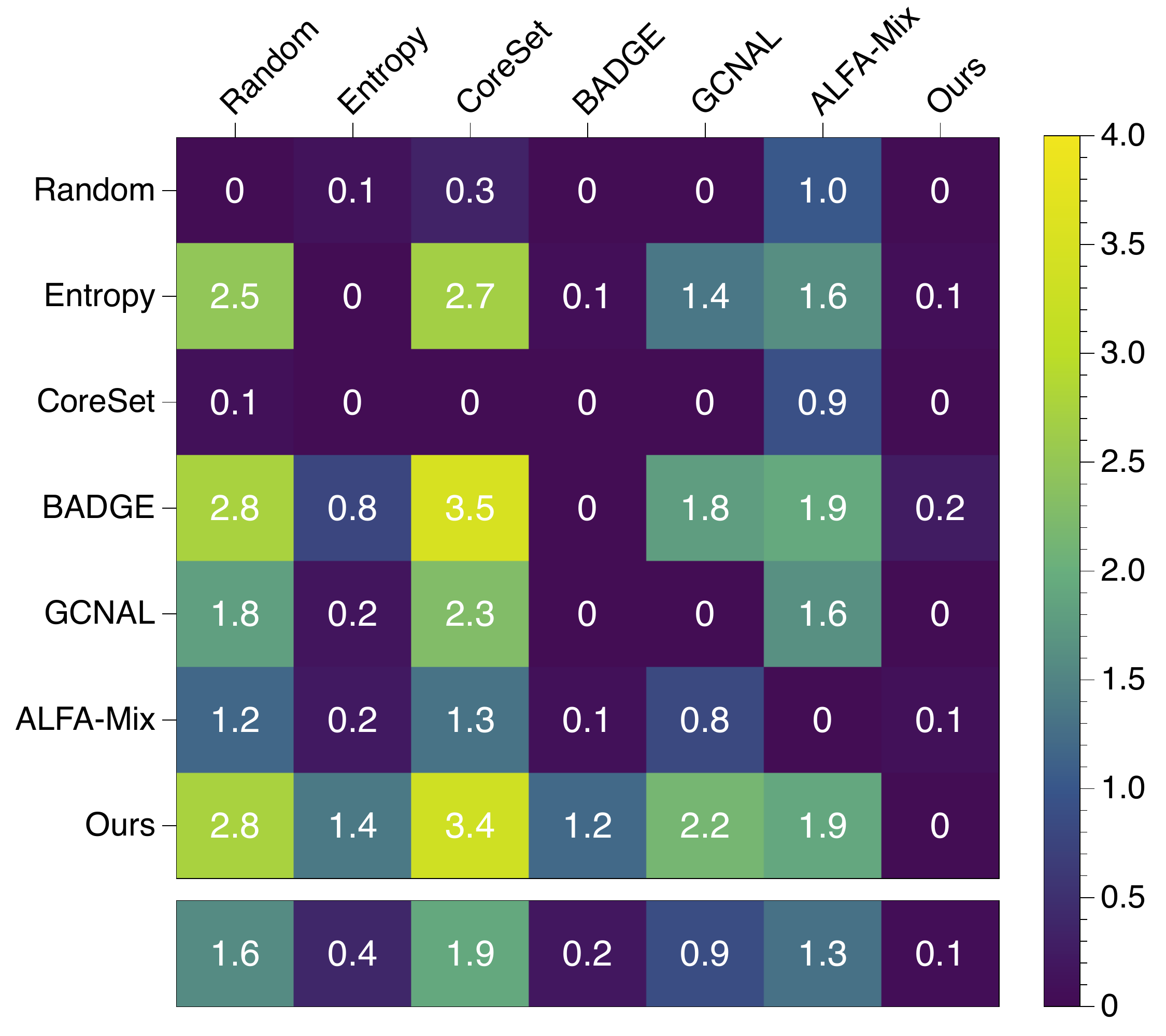}
    \caption{$\alpha = 1.0$}
    \end{subfigure}
    \hfill
    \begin{subfigure}[b]{0.3\linewidth}
    \includegraphics[width=\linewidth]{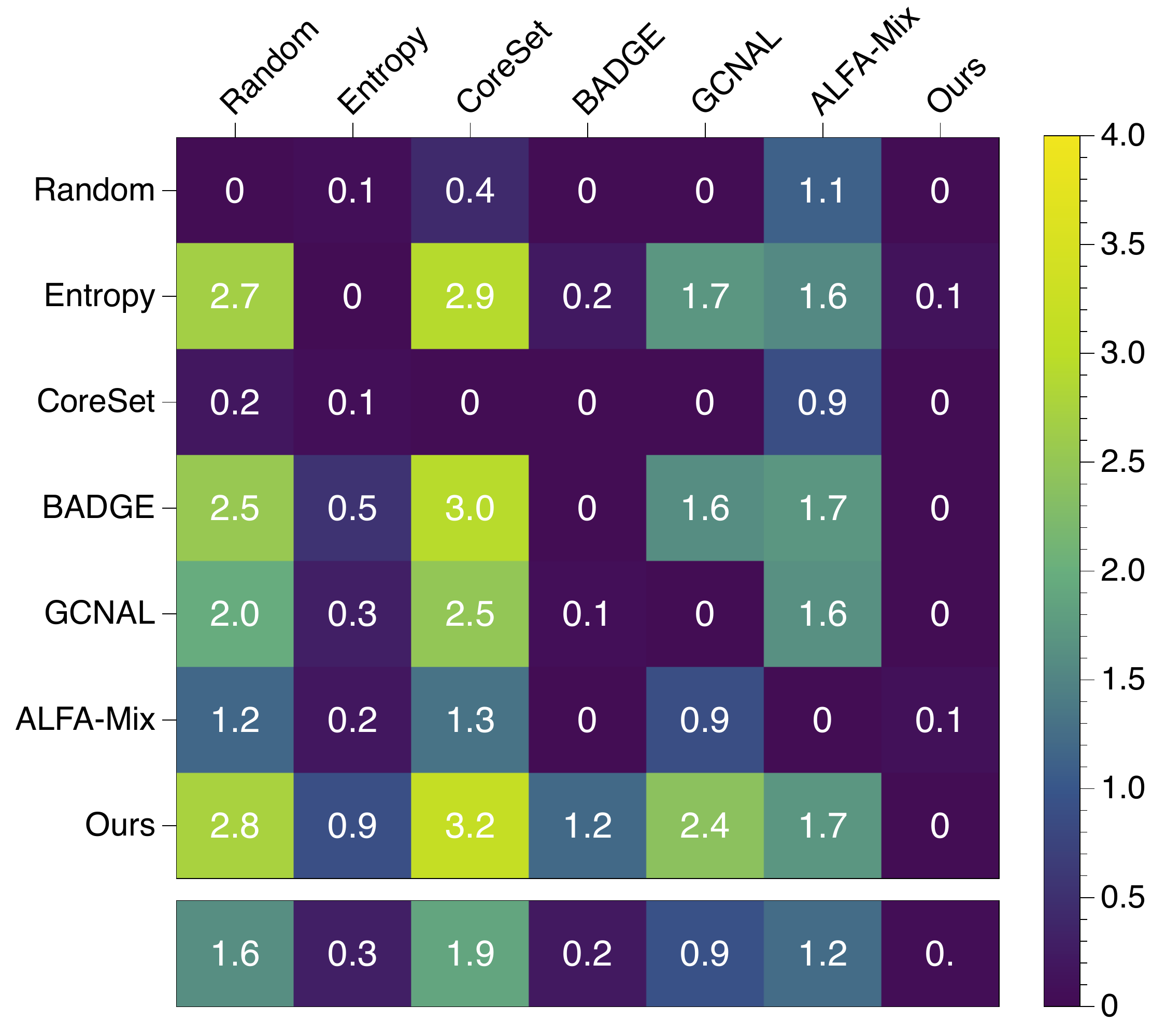}
    \caption{$\alpha = \infty$}
    \end{subfigure}
    \caption{Pairwise penalty matrix for a heterogeneity level category. The maximum value of three matrices is 30, 4, and 4.}
    \label{fig:comp_hetero}
\end{figure*}

\begin{figure*}[htb]
    \centering
    \begin{subfigure}[b]{0.3\linewidth}
    \raggedleft
    \includegraphics[width=\linewidth]{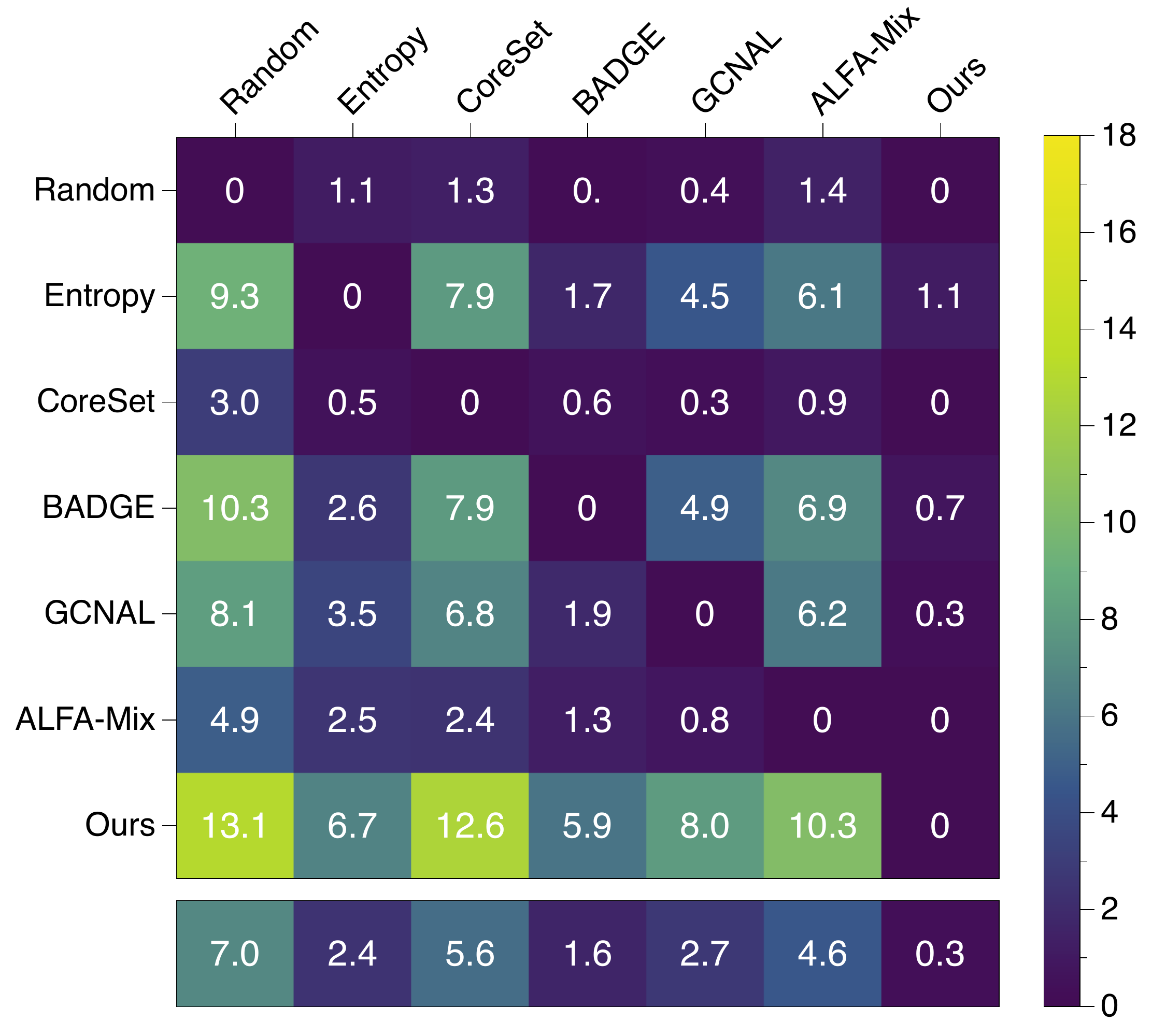}
    \caption{$\rho <$ 2}
    \end{subfigure}
    \hspace{5pt}
    \begin{subfigure}[b]{0.3\linewidth}
    \raggedright
    \includegraphics[width=\linewidth]{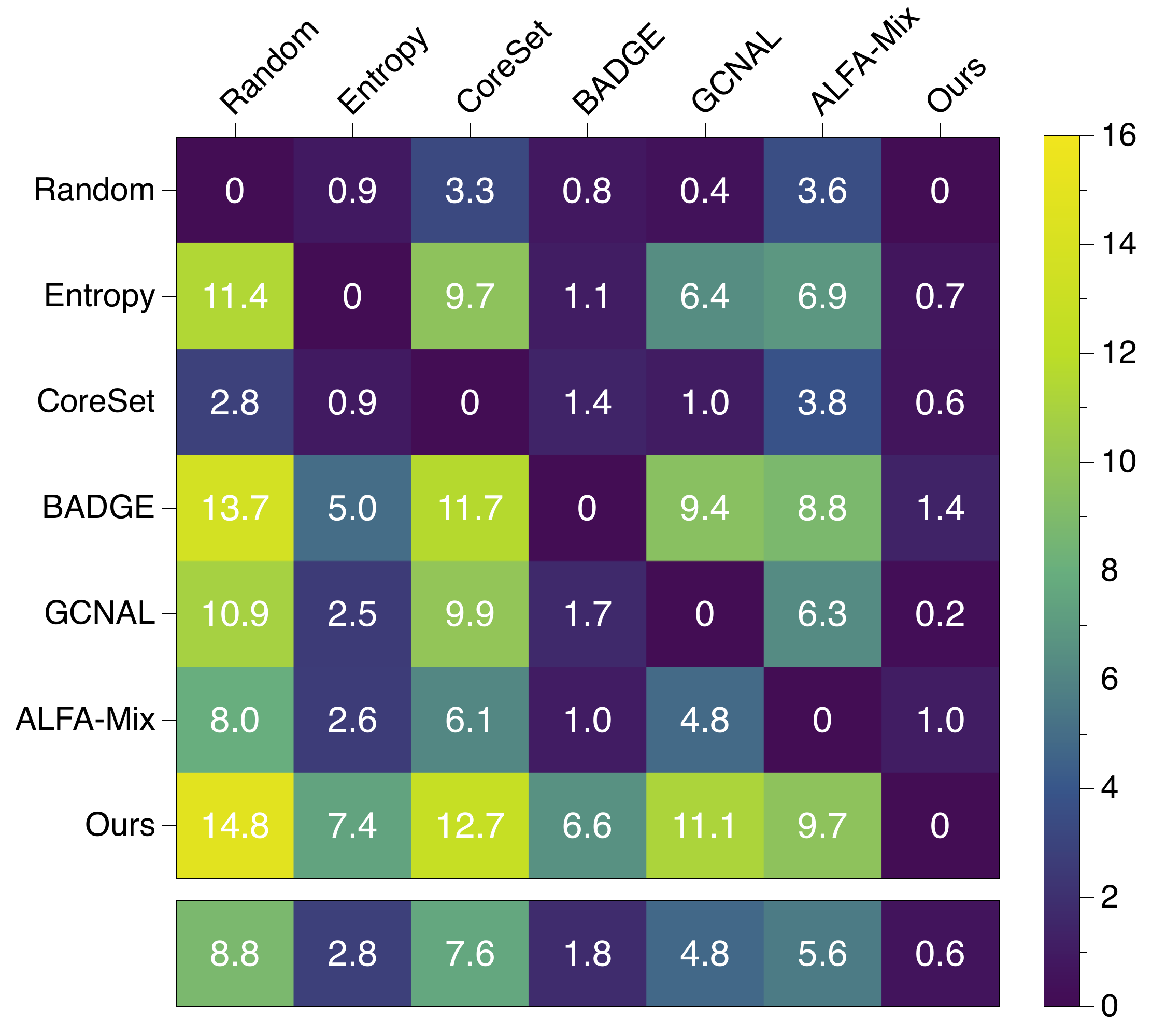}
    \caption{$\rho \ge$ 2}
    \end{subfigure}
    \caption{Pairwise penalty matrix for imbalance ratio category. The maximum value of two matrices is 18 and 20, respectively.}
    \label{fig:comp_data_type}
\end{figure*}

\begin{figure*}[!t]
    \centering
    \begin{subfigure}[b]{0.3\linewidth}
    \includegraphics[width=\linewidth]{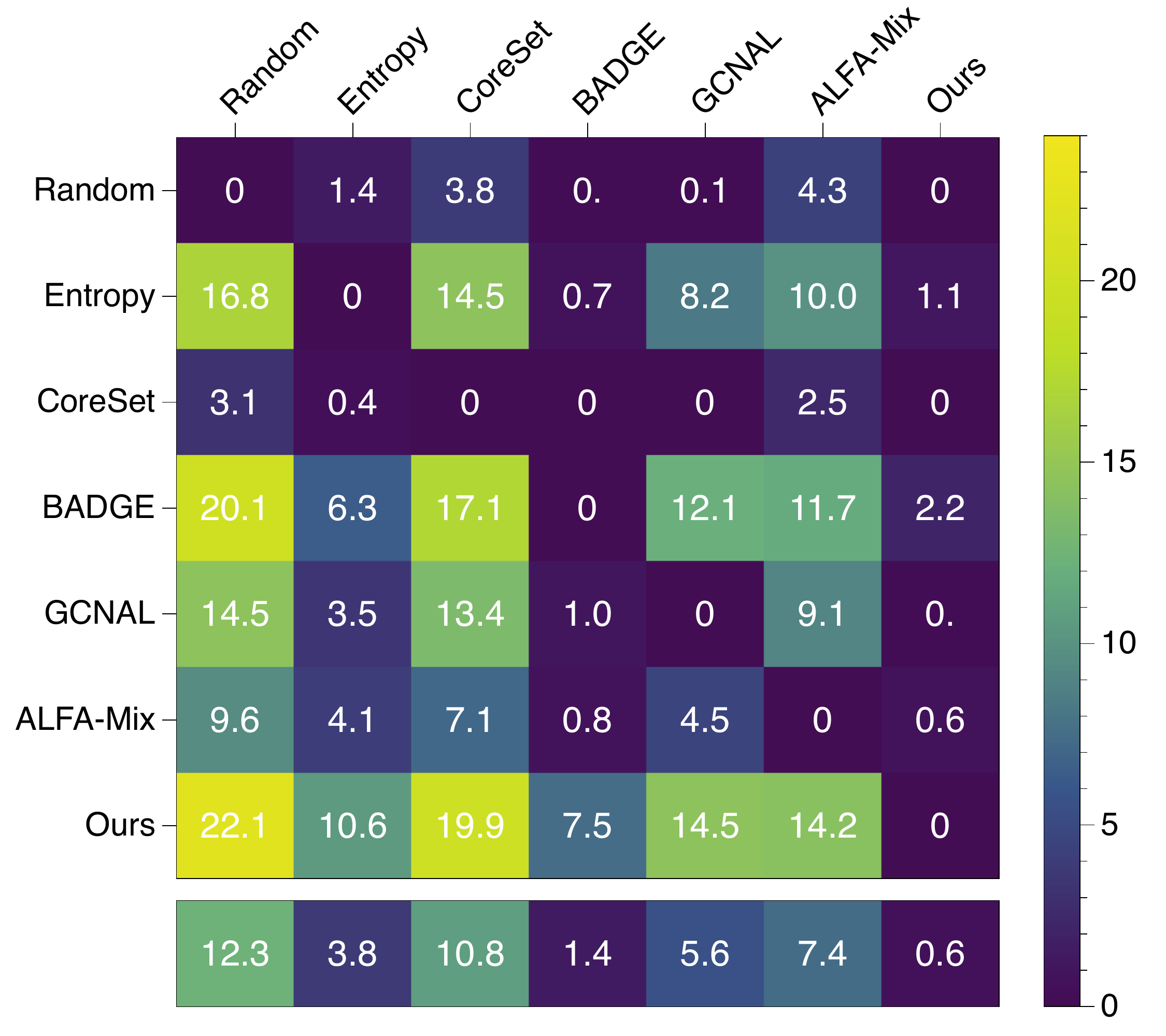}
    \caption{Four Convolutional Neural Network}
    \end{subfigure}
    \hfill
    \begin{subfigure}[b]{0.3\linewidth}
    \includegraphics[width=\linewidth]{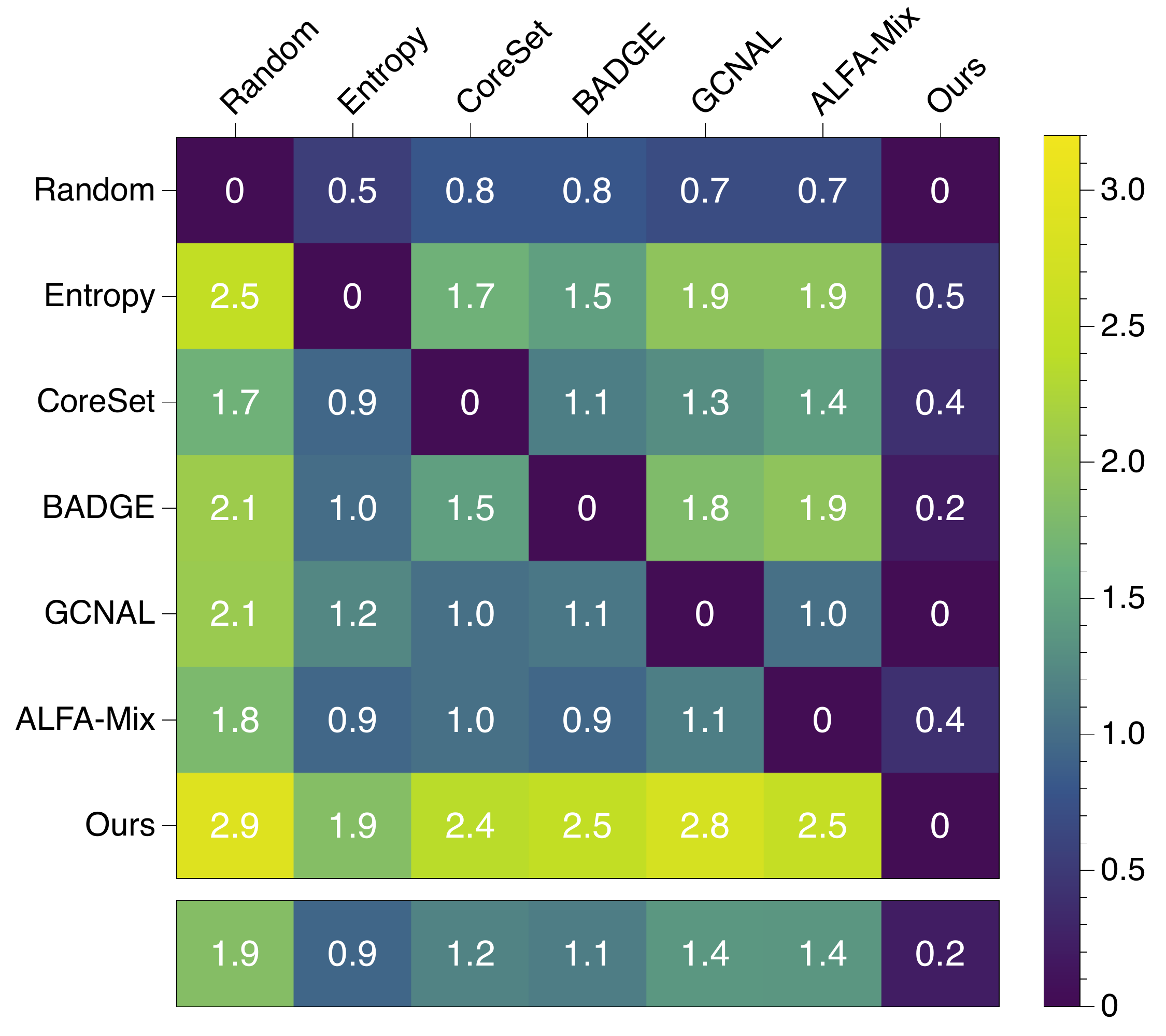}
    \caption{ResNet-18}
    \end{subfigure}
    \hfill
    \begin{subfigure}[b]{0.3\linewidth}
    \includegraphics[width=\linewidth]{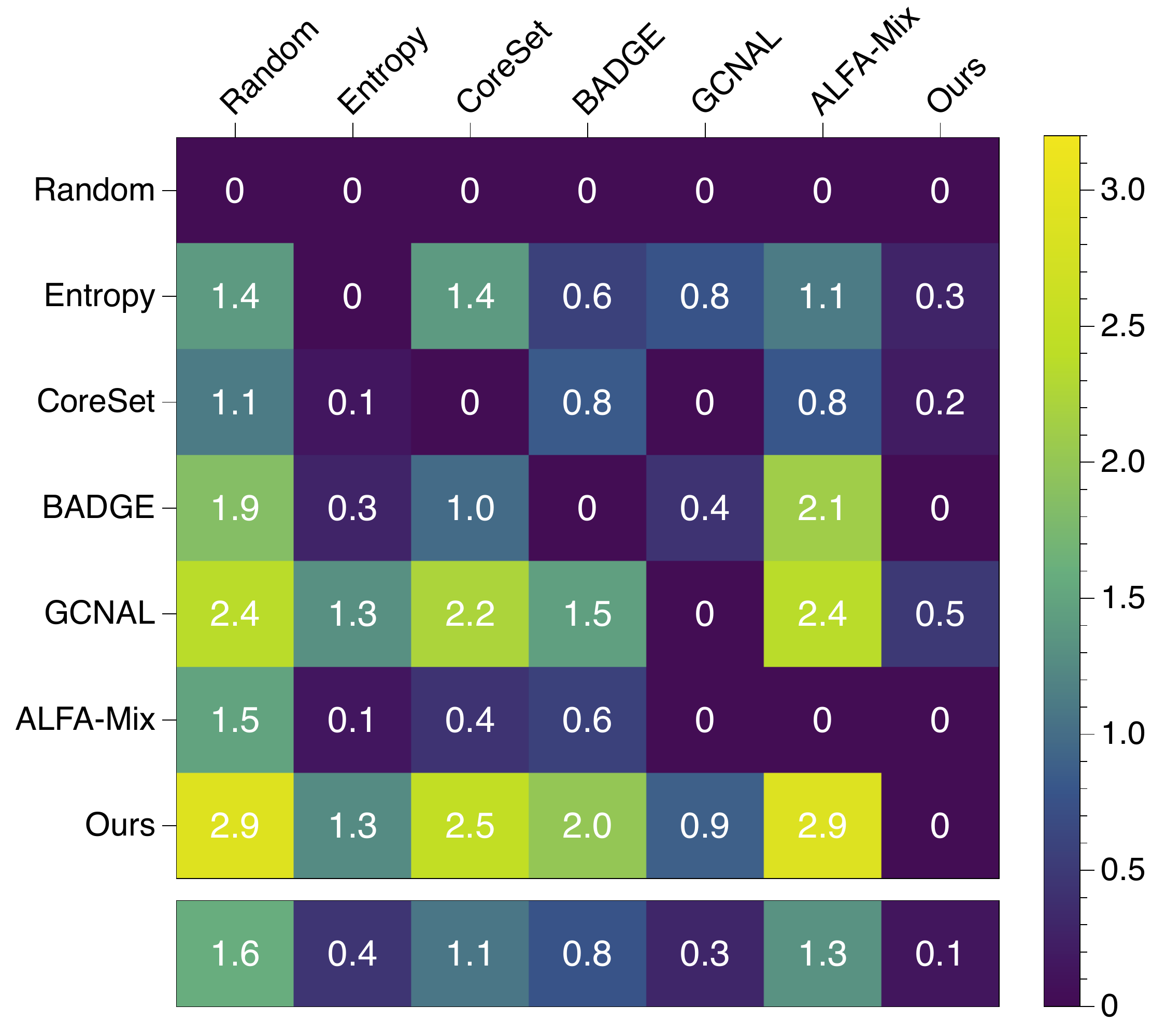}
    \caption{MobileNet}
    \end{subfigure}
    \caption{Pairwise penalty matrix for a model architecture category. The maximum value of three matrices is 30, 4, and 4.}
    \label{fig:comp_model_arch}
\end{figure*}

\begin{figure*}[!t]
    \centering
    \begin{subfigure}[b]{0.3\linewidth}
    \includegraphics[width=\linewidth]{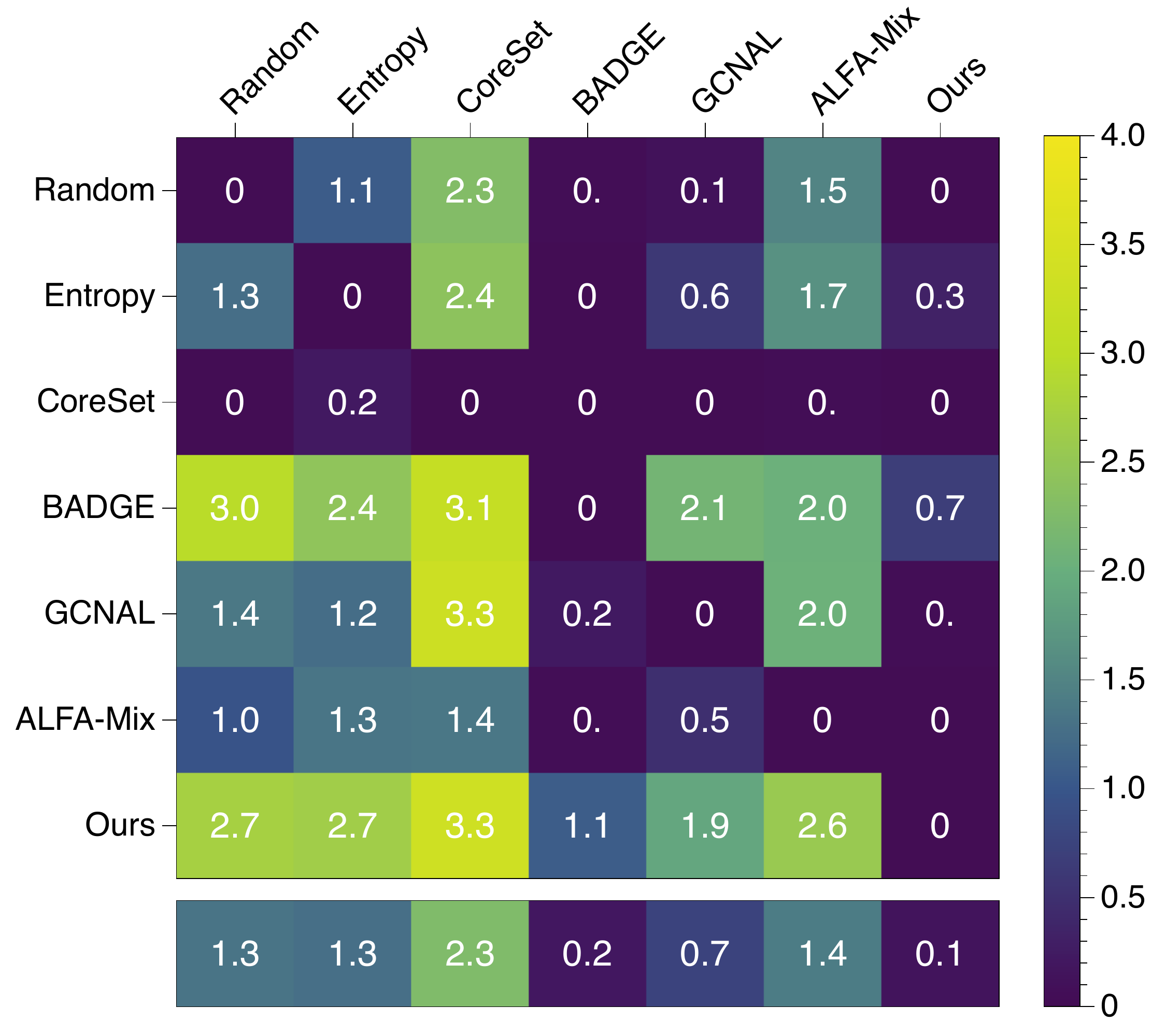}
    \caption{Budget 1\%}
    \end{subfigure}
    \hfill
    \begin{subfigure}[b]{0.3\linewidth}
    \includegraphics[width=\linewidth]{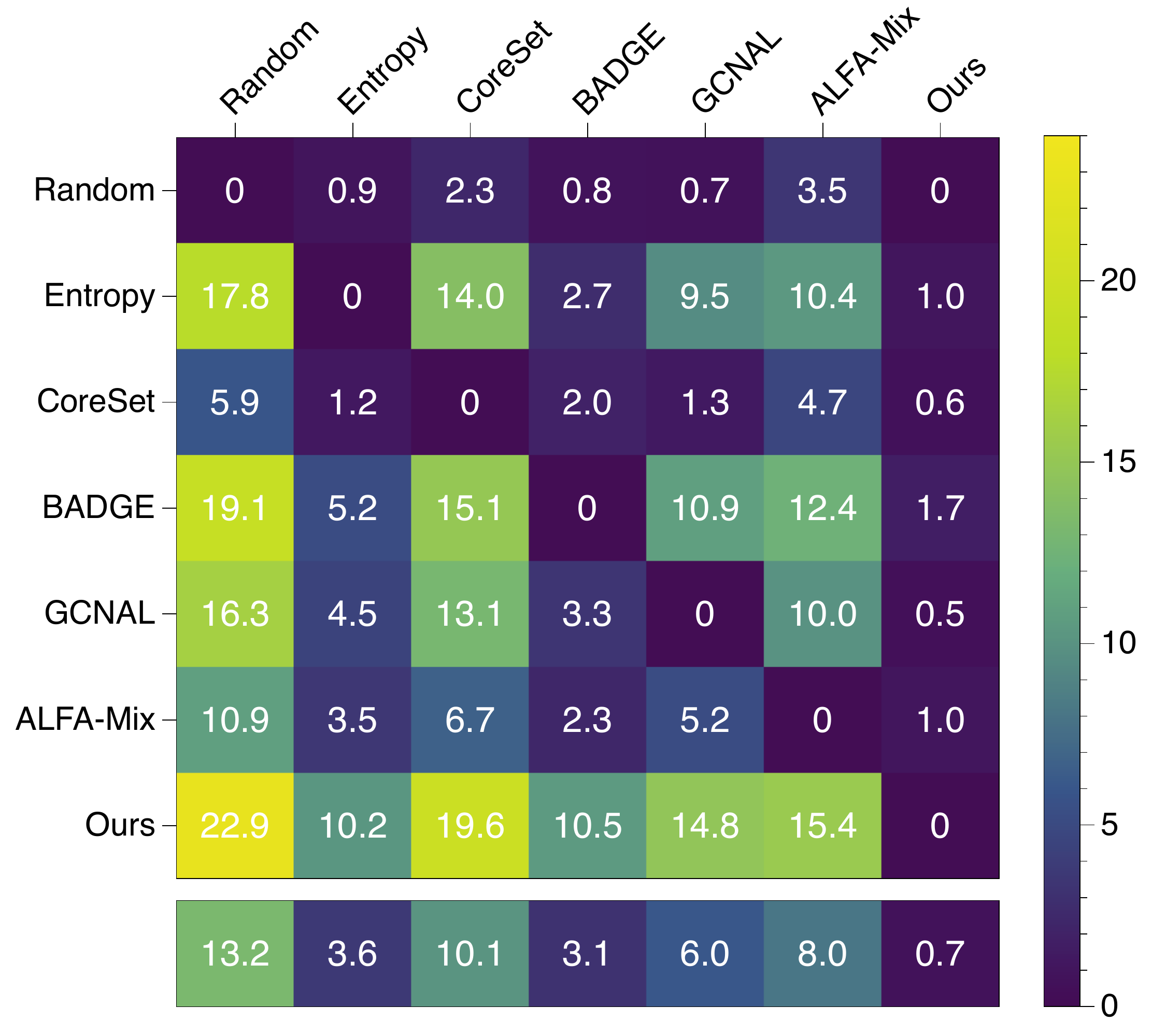}
    \caption{Budget 5\%}
    \end{subfigure}
    \hfill
    \begin{subfigure}[b]{0.3\linewidth}
    \includegraphics[width=\linewidth]{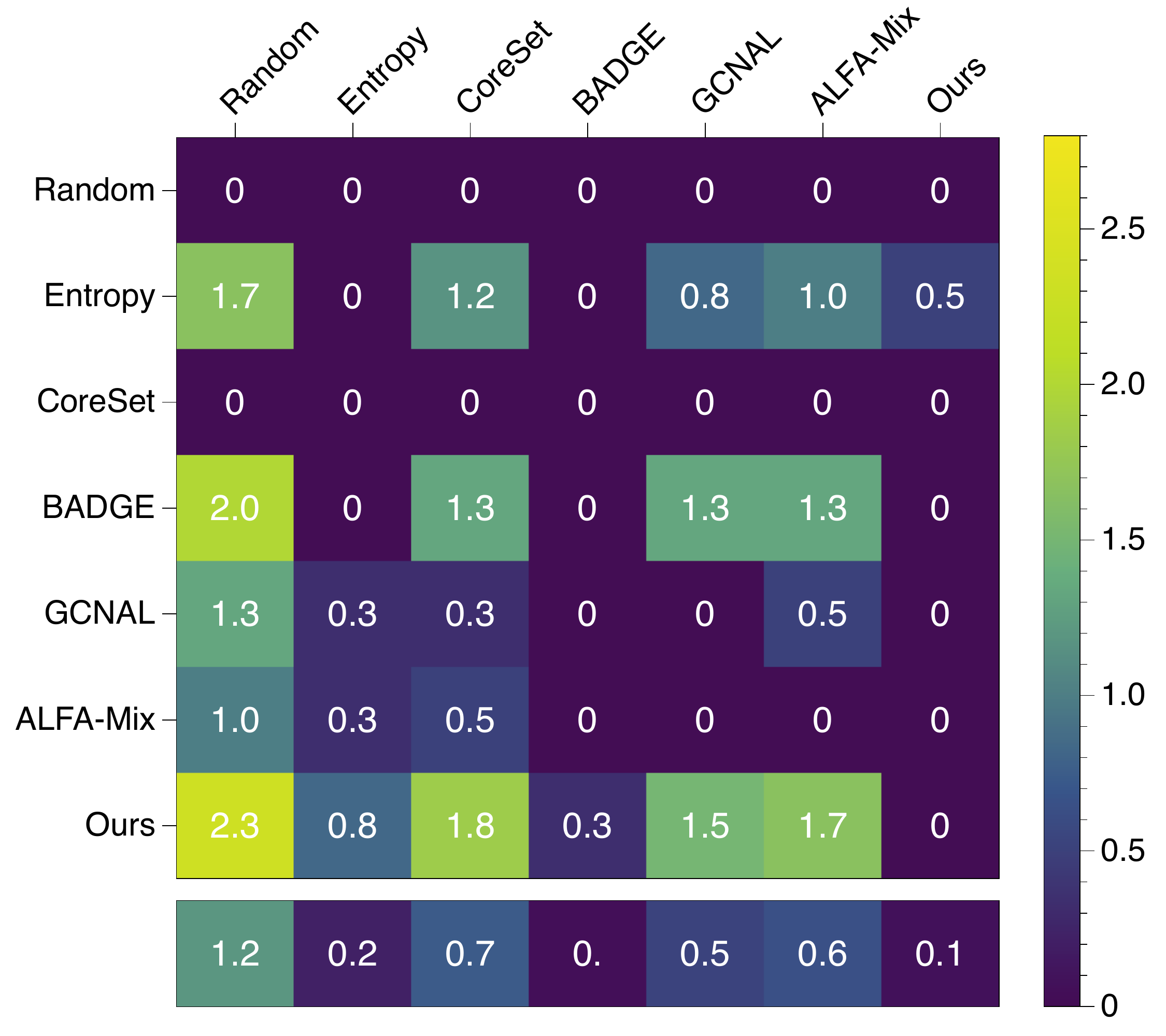}
    \caption{Budget 20\%}
    \end{subfigure}
    \caption{Pairwise penalty matrix for a budget size category. The maximum value of three matrices is 4, 30, and 4, respectively.}
    \label{fig:comp_budget_size}
\end{figure*}

\begin{figure*}[!t]
    \centering
    \begin{subfigure}[b]{0.3\linewidth}
    \raggedleft
    \includegraphics[width=\linewidth]{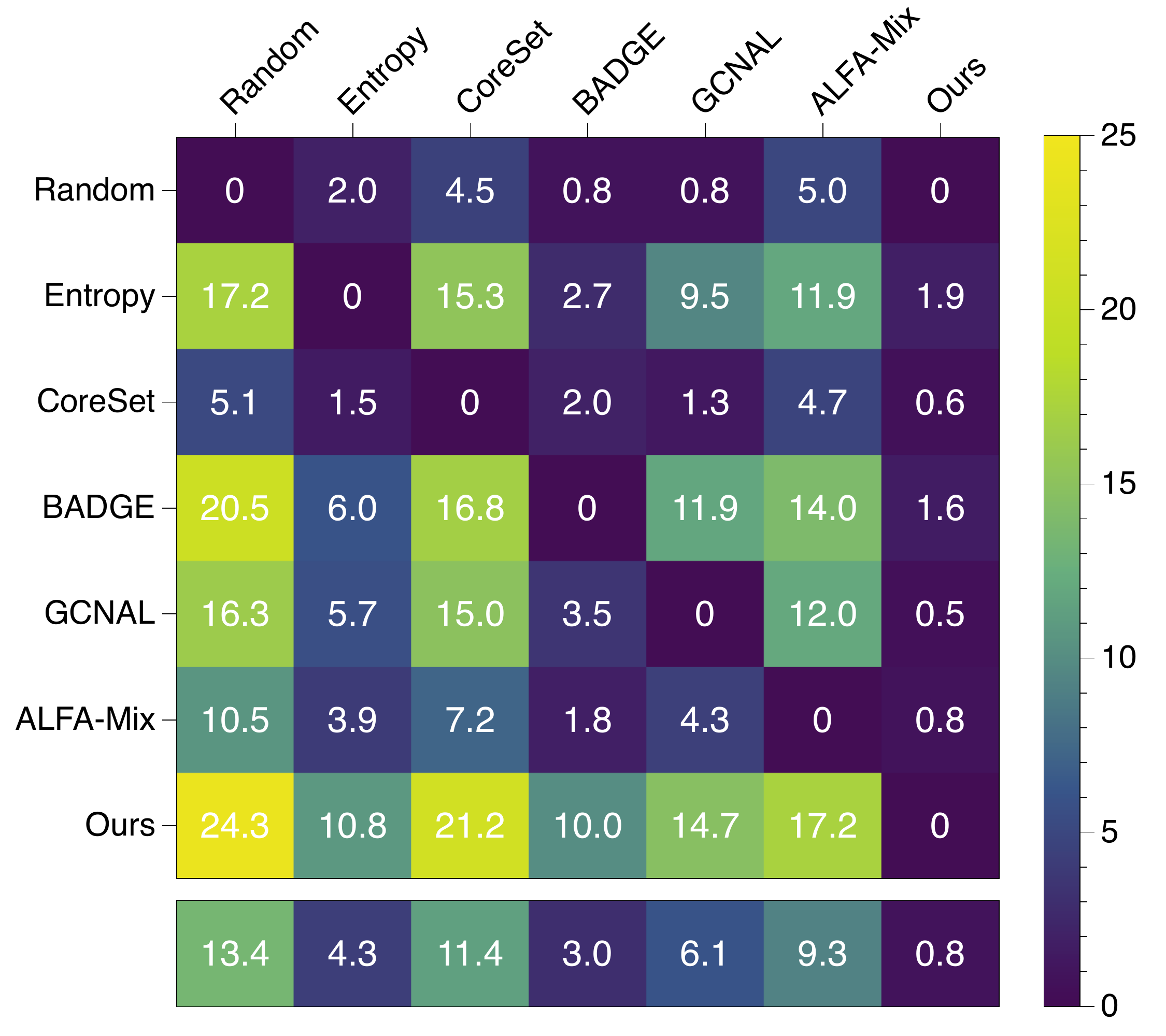}
    \caption{Random initialization}
    \end{subfigure}
    \hspace{10pt}
    \begin{subfigure}[b]{0.3\linewidth}
    \raggedright
    \includegraphics[width=\linewidth]{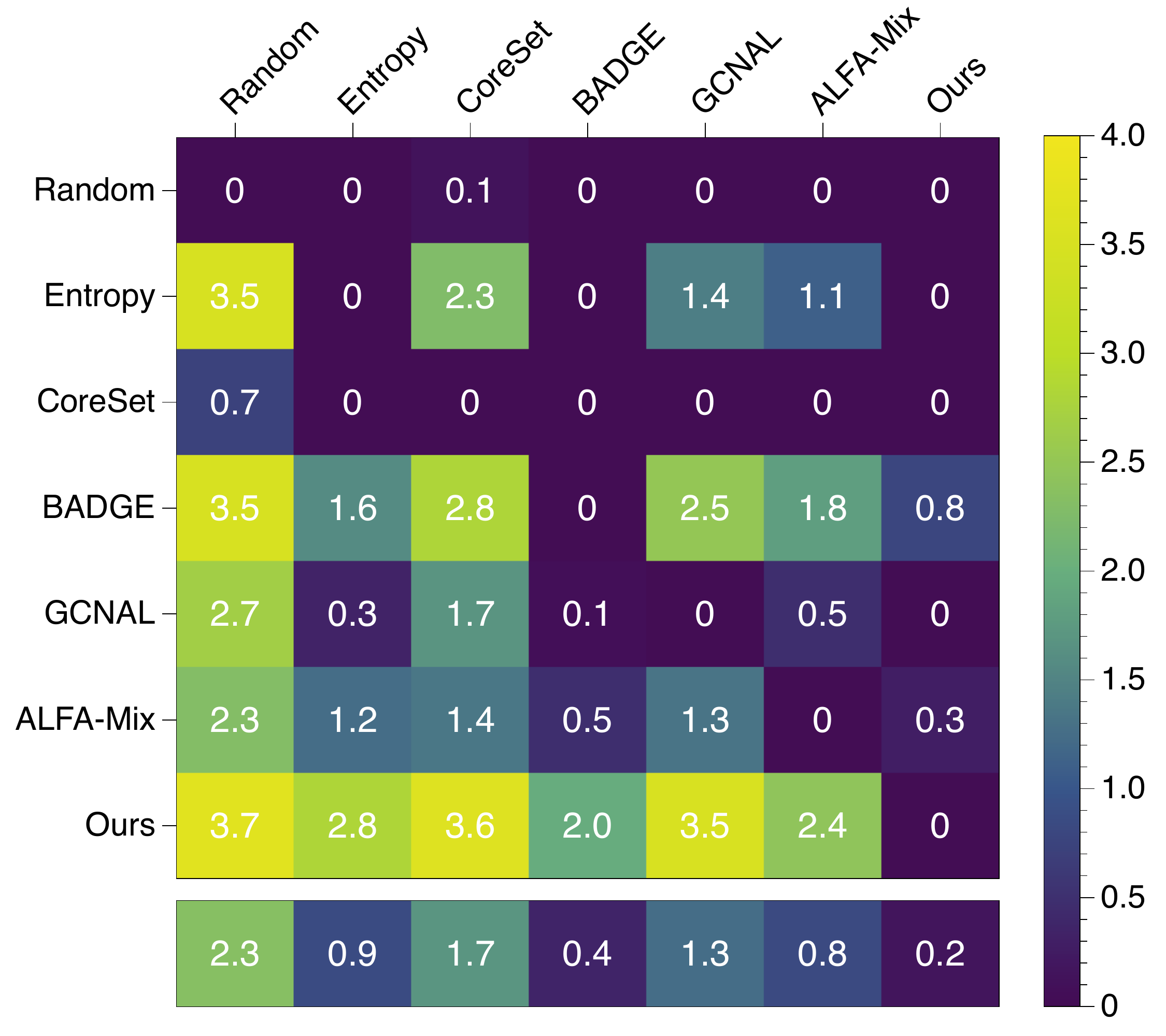}
    \caption{Continue initialization}
    \end{subfigure}
    \caption{Pairwise penalty matrix for a model initialization category. The maximum value of two matrices is 34 and 4.}
    \label{fig:comp_model_init}
\end{figure*}

In this Section, \ref{sec:detail_comp_matrix} summarizes all the comparison matrices results based on six categories: query selector, heterogeneity level, imbalance ratio, model architecture, budget size, and model initialization in Figure\,\ref{fig:bar_graph}. Figure\,\ref{fig:comp_selector}--\ref{fig:comp_model_init} are breakdowns of the matrix in Figure\,\ref{fig:comparision_matrix} into six categories.
\ref{sec:detail_comp_performance} provides comprehensive line plots for 38 experimental settings.
It can be seen that \algname{} overwhelms the baselines in most cases at both each category and detailed experimental setting level.

\subsection{Detailed Penalty Comparision Matrix}
\label{sec:detail_comp_matrix}

A maximum value of each matrix corresponds to Table\,\ref{tab:setting_summary}, and the bar plots in Figure\,\ref{fig:bar_graph} are calculated from these matrices.

\vspace{-15pt}

\vspace{-15pt}

\vspace{-15pt}

\vspace{-15pt}

\vspace{-15pt}

\clearpage

\subsection{Detailed Performance Comparision}
\label{sec:detail_comp_performance}

For the line plots, we note that `Random' and `Ours' are independent of the query selector type.

\begin{figure*}[h!]
    \centering
    \includegraphics[width=0.8\linewidth]{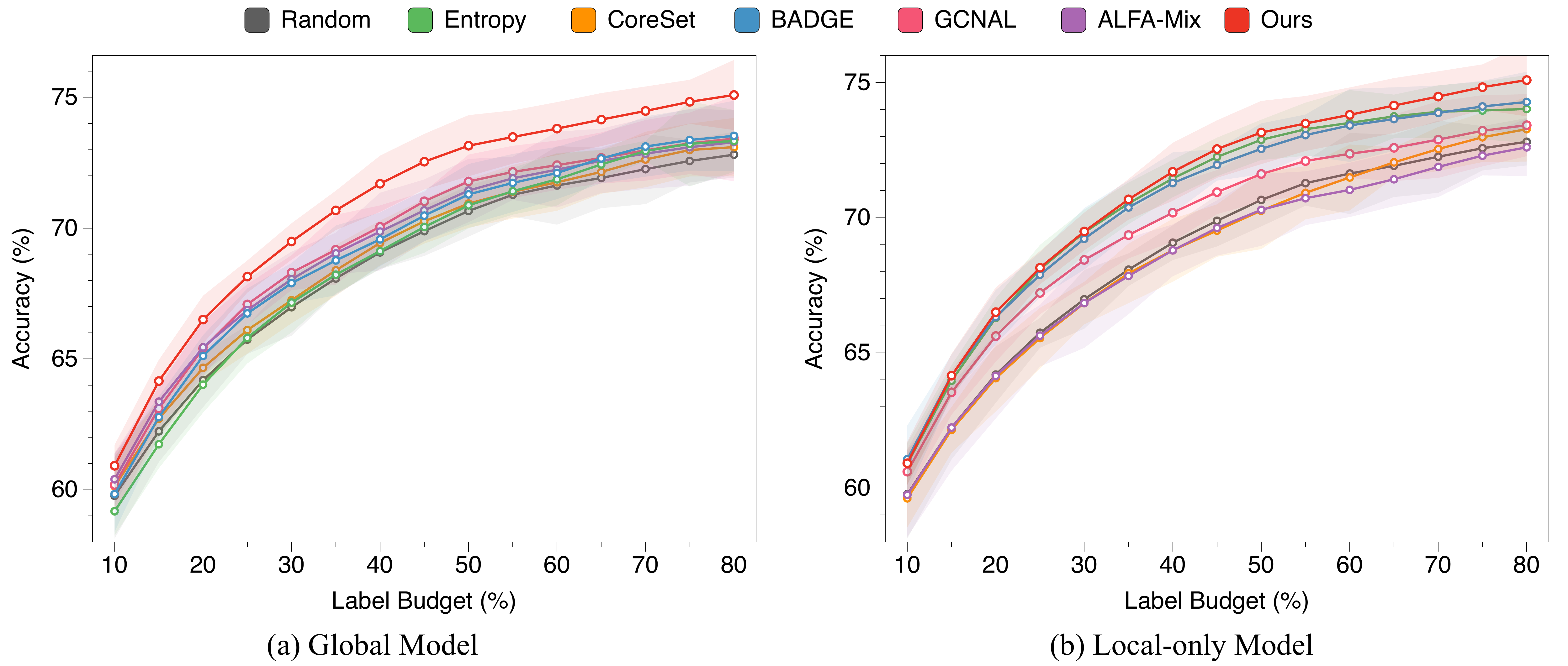}
    \caption{Test accuracy on CIFAR-10, four layers of CNN, $\alpha=0.1$,  medium budget size\,(5\%), and random initialization. }
    \label{fig:app_cifar10}
\end{figure*}

\vspace{-15pt}
\begin{figure*}[h!]
    \centering
    \includegraphics[width=0.8\linewidth]{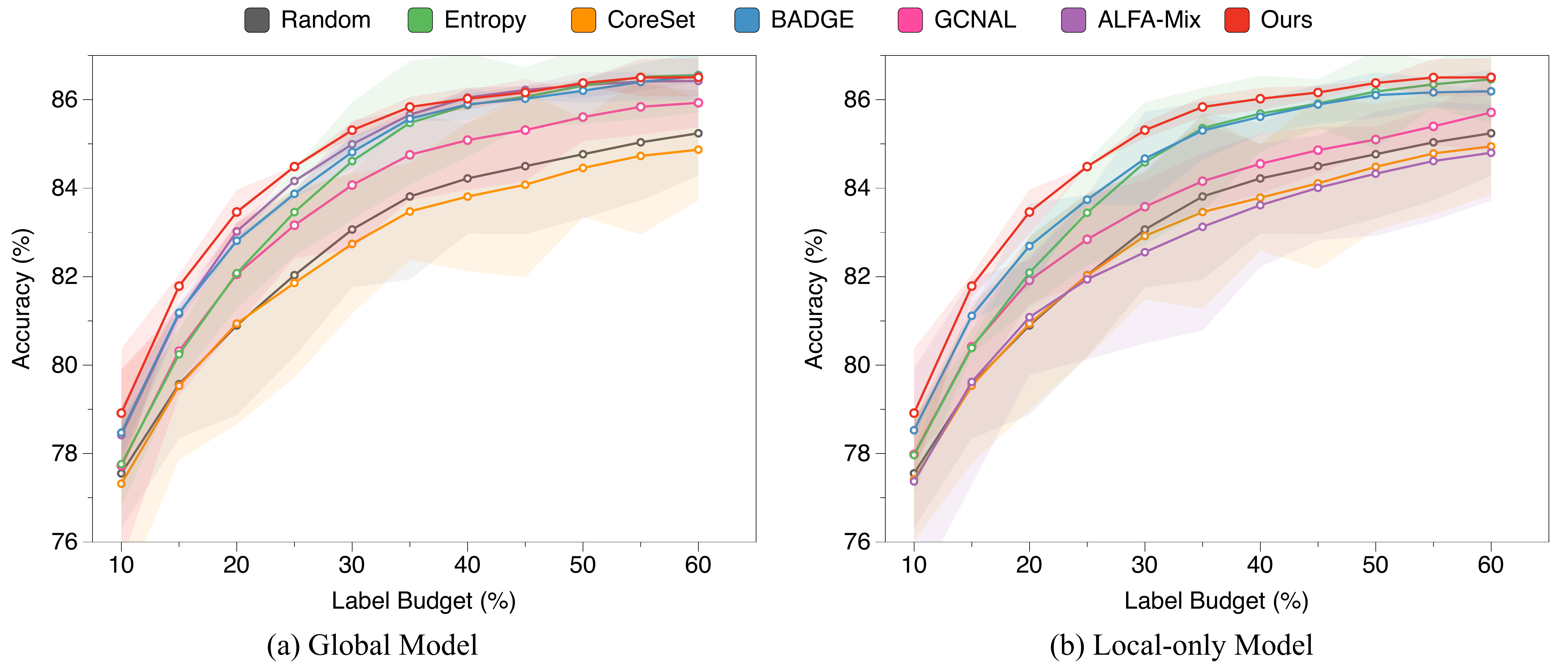}
    \caption{Test accuracy on SVHN, four layers of CNN, $\alpha=0.1$,  medium budget size\,(5\%), and random initialization.}
    \label{fig:app_svhn}
\end{figure*}

\vspace{-15pt}
\begin{figure*}[!h]
    \centering
    \includegraphics[width=0.8\linewidth]{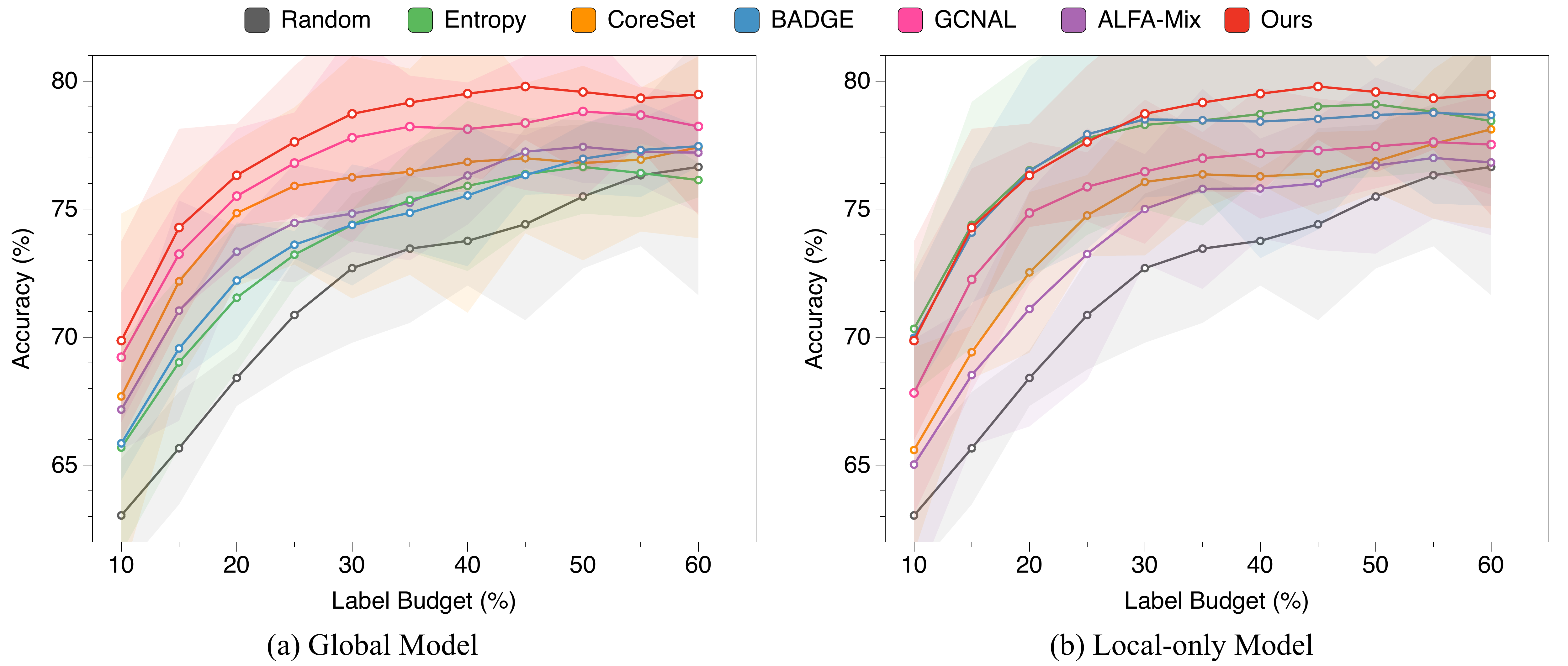}
    \caption{Test accuracy on PathMNIST, four layers of CNN, $\alpha=0.1$,  medium budget size\,(5\%), and random initialization.}
    \label{fig:app_pathmnist}
\end{figure*}

\vspace{-15pt}
\begin{figure*}[!h]
    \centering
    \includegraphics[width=0.8\linewidth]{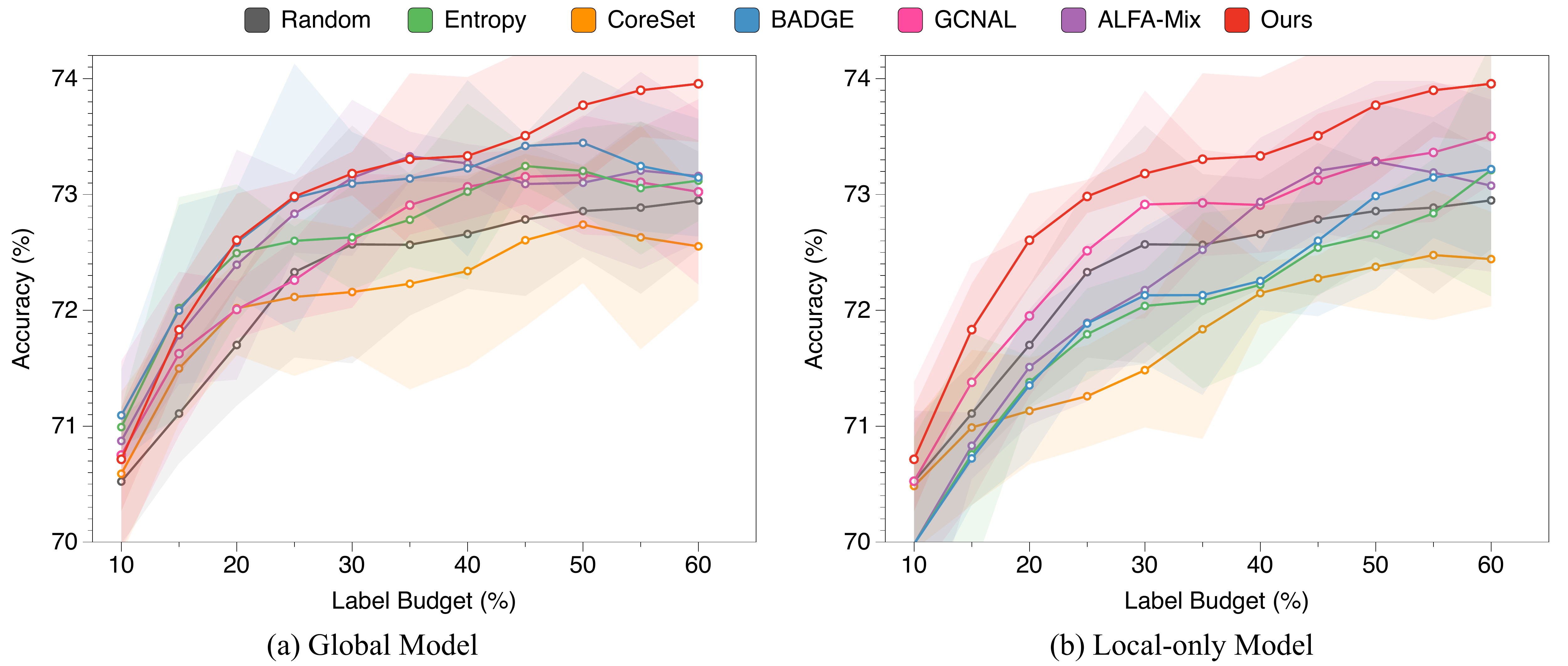}
    \caption{Test accuracy on DermaMNIST, four layers of CNN, $\alpha=0.1$,  medium budget size\,(5\%), and random initialization.}
    \label{fig:app_dermamnist}
\end{figure*}

\vspace{-15pt}
\begin{figure*}[!h]
    \centering
    \includegraphics[width=0.8\linewidth]{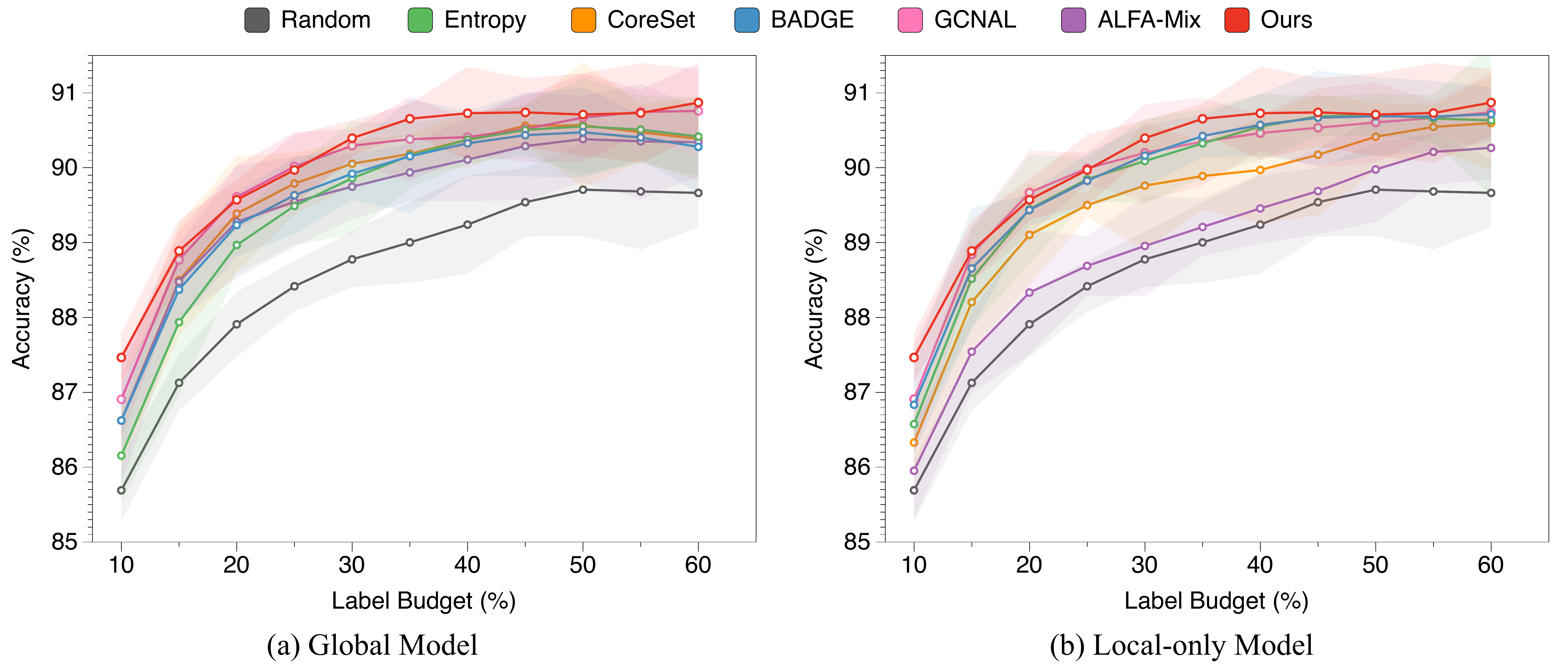}
    \caption{Test accuracy on OrganAMNIST, four layers of CNN, $\alpha=0.1$,  medium budget size\,(5\%), and random initialization.}
    \label{fig:app_organmnist}
\end{figure*}

\vspace{-15pt}
\begin{figure*}[!h]
    \centering
    \includegraphics[width=0.8\linewidth]{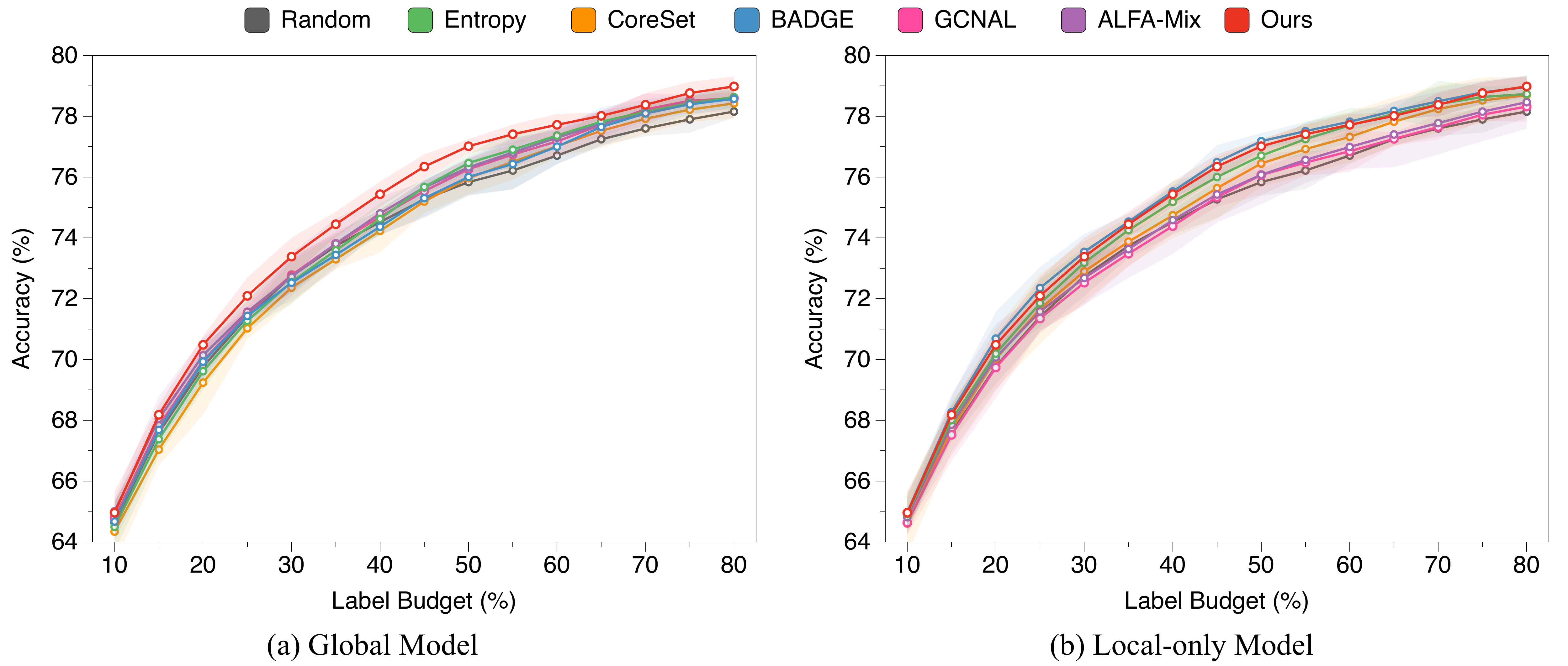}
    \caption{Test accuracy on CIFAR-10, four layers of CNN, \textbf{$\alpha=1.0$},  medium budget size\,(5\%), and random initialization.}
    \label{fig:app_cifar_dir1}
\end{figure*}

\vspace{-15pt}
\begin{figure*}[!h]
    \centering
    \includegraphics[width=0.8\linewidth]{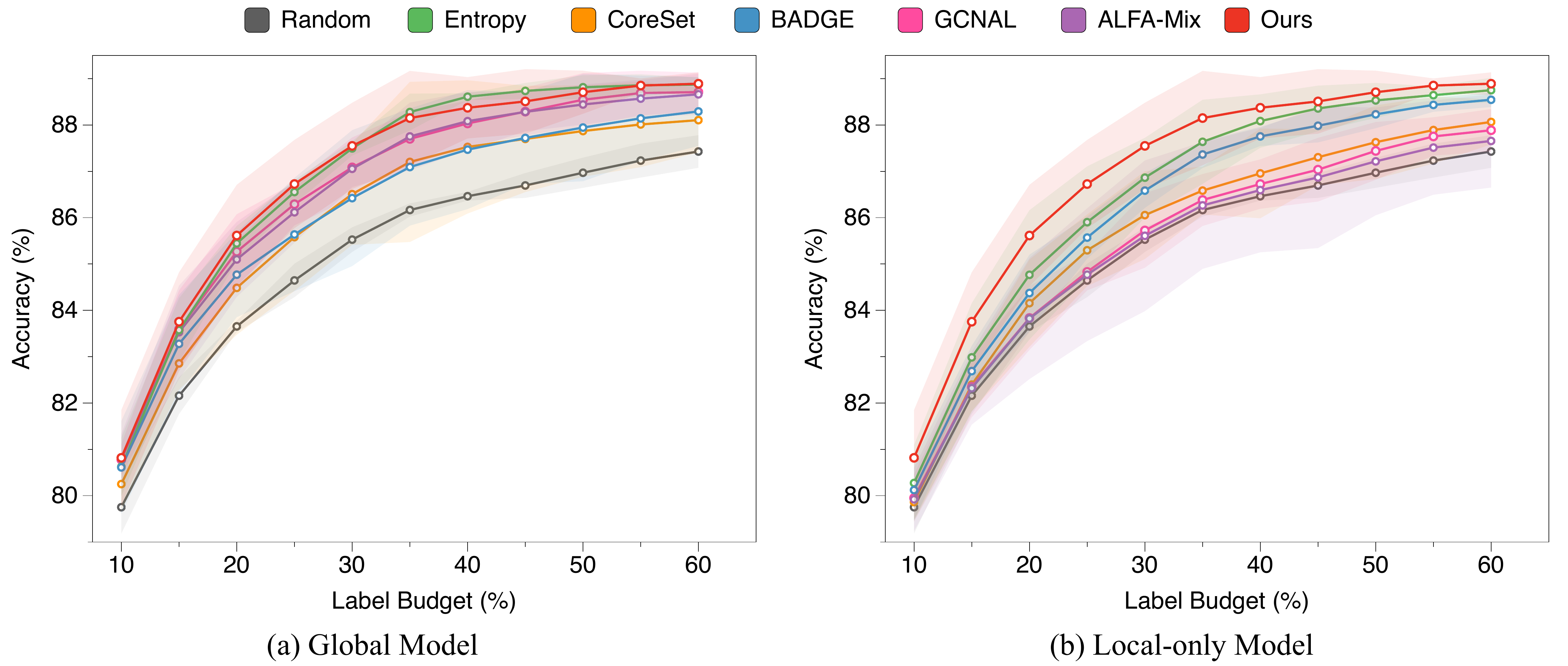}
    \caption{Test accuracy on SVHN, four layers of CNN, \textbf{$\alpha=1.0$},  medium budget size\,(5\%), and random initialization.}
    \label{fig:app_svhn_dir1}
\end{figure*}

\vspace{-15pt}
\begin{figure*}[!h]
    \centering
    \includegraphics[width=0.8\linewidth]{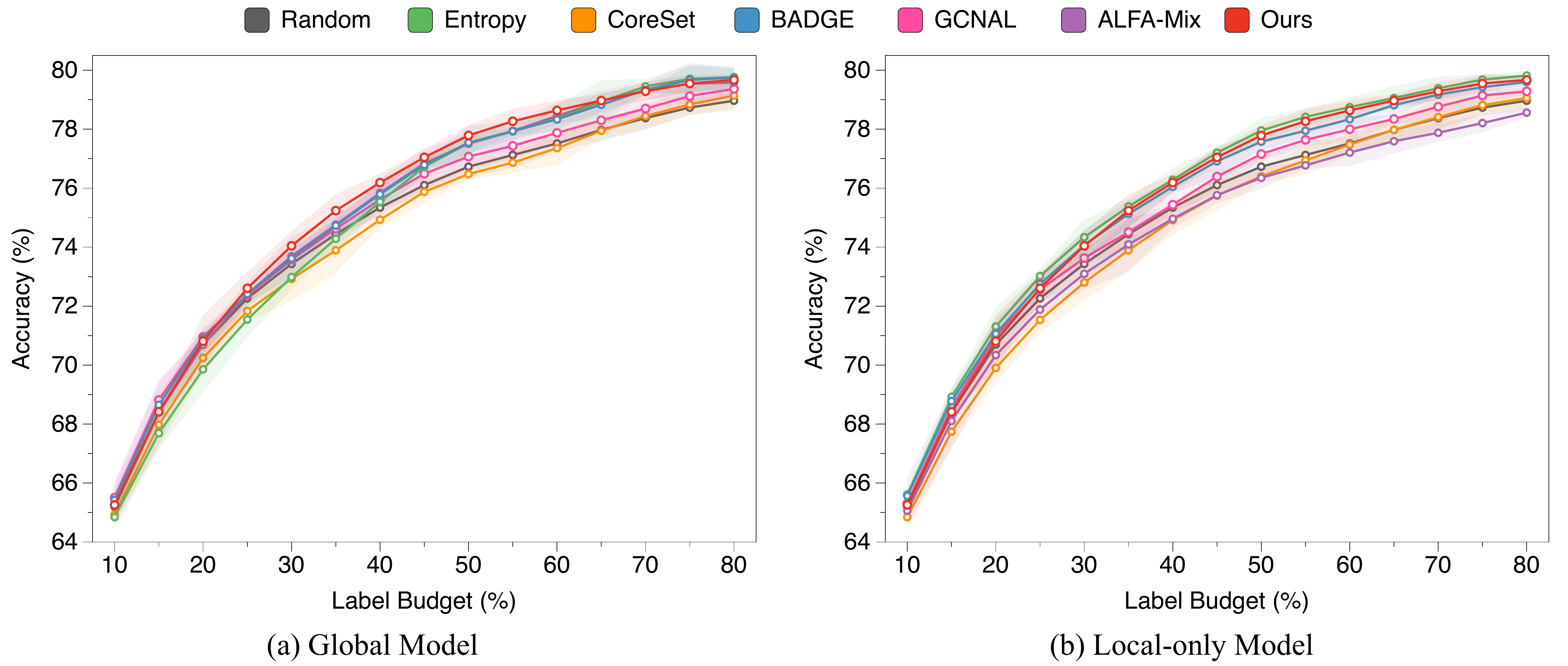}
    \caption{Test accuracy on CIFAR-10, four layers of CNN, \textbf{$\alpha=\infty$},  medium budget size\,(5\%), and random initialization.}
    \label{fig:app_cifar_dir10}
\end{figure*}

\vspace{-15pt}
\begin{figure*}[!h]
    \centering
    \includegraphics[width=0.8\linewidth]{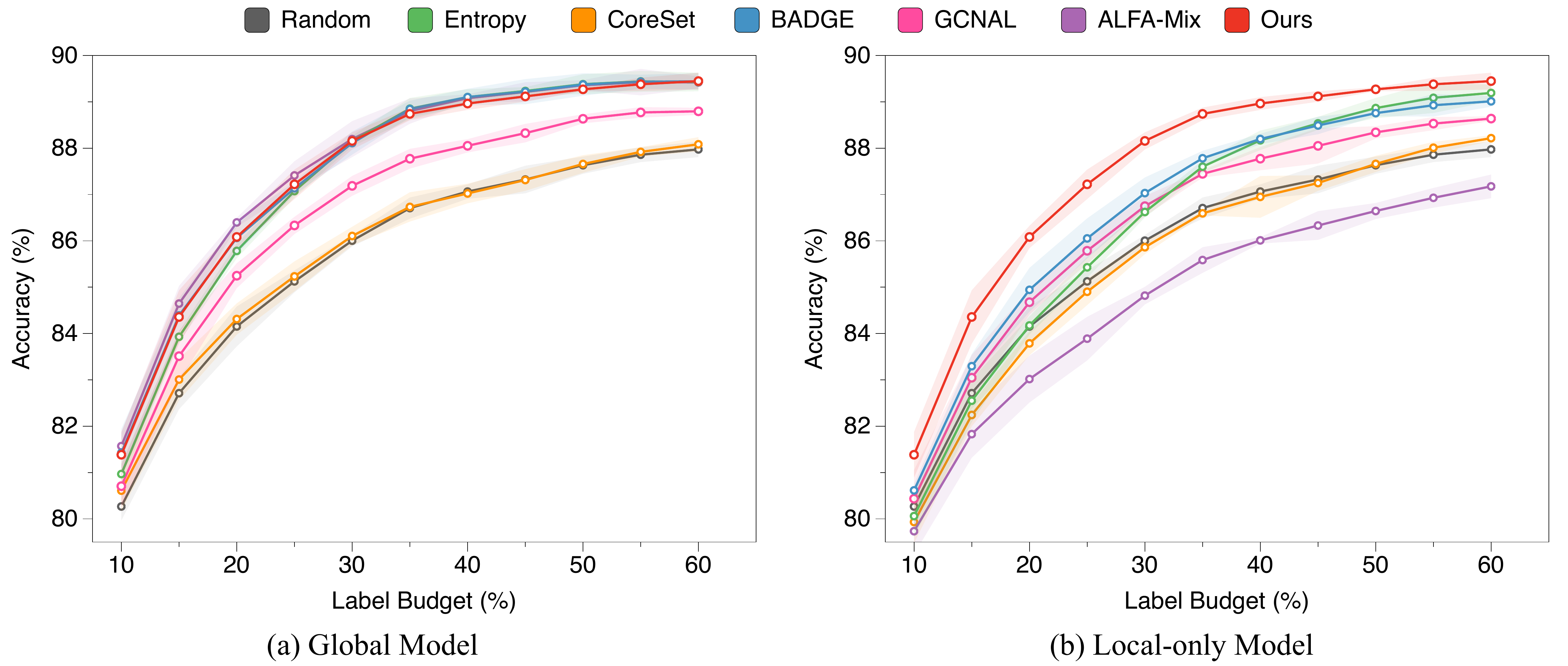}
    \caption{Test accuracy on SVHN, four layers of CNN, \textbf{$\alpha=\infty$},  medium budget size\,(5\%), and random initialization.}
    \label{fig:app_svhn_dir10}
\end{figure*}

\vspace{-15pt}
\begin{figure*}[!h]
    \centering
    \includegraphics[width=0.8\linewidth]{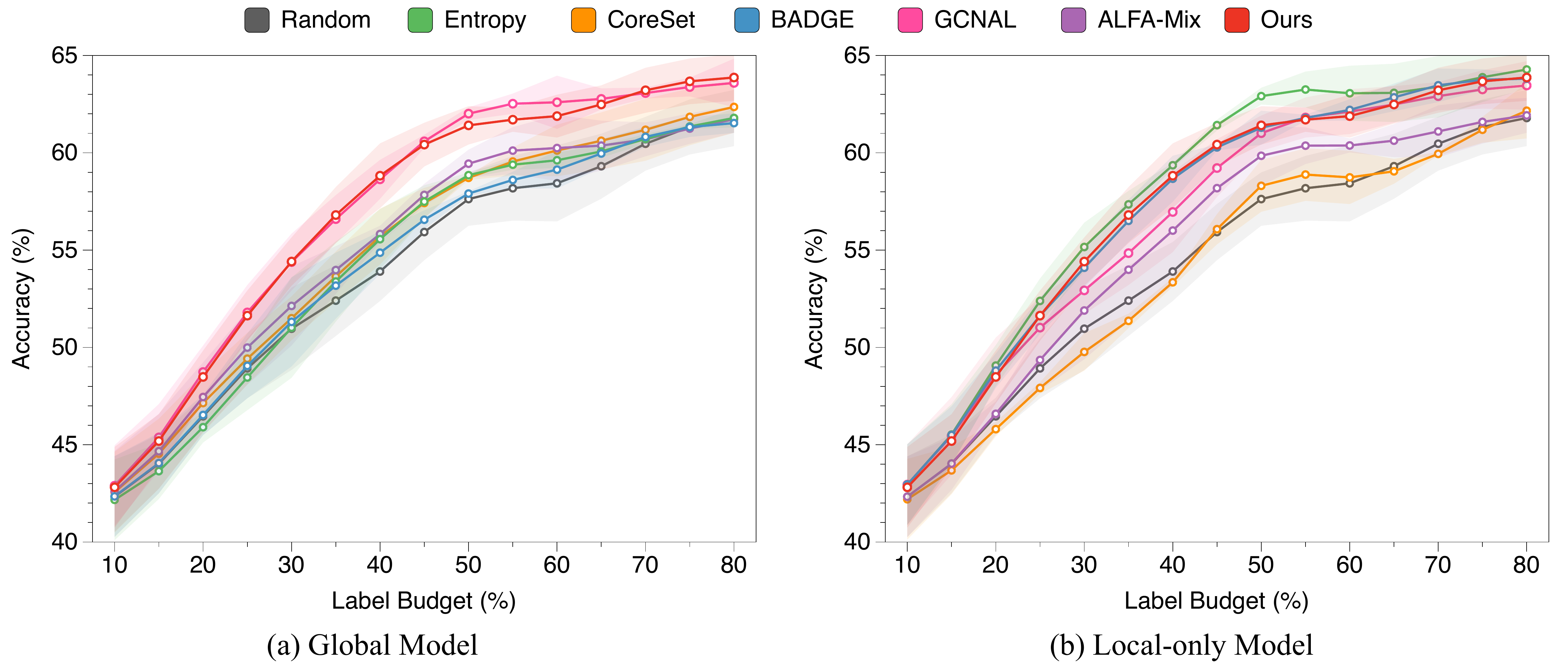}
    \caption{Test accuracy on CIFAR-10, MobileNet, $\alpha=0.1$,  medium budget size\,(5\%), and random initialization.}
    \label{fig:app_cifar10_mobile}
\end{figure*}

\vspace{-15pt}
\begin{figure*}[!h]
    \centering
    \includegraphics[width=0.8\linewidth]{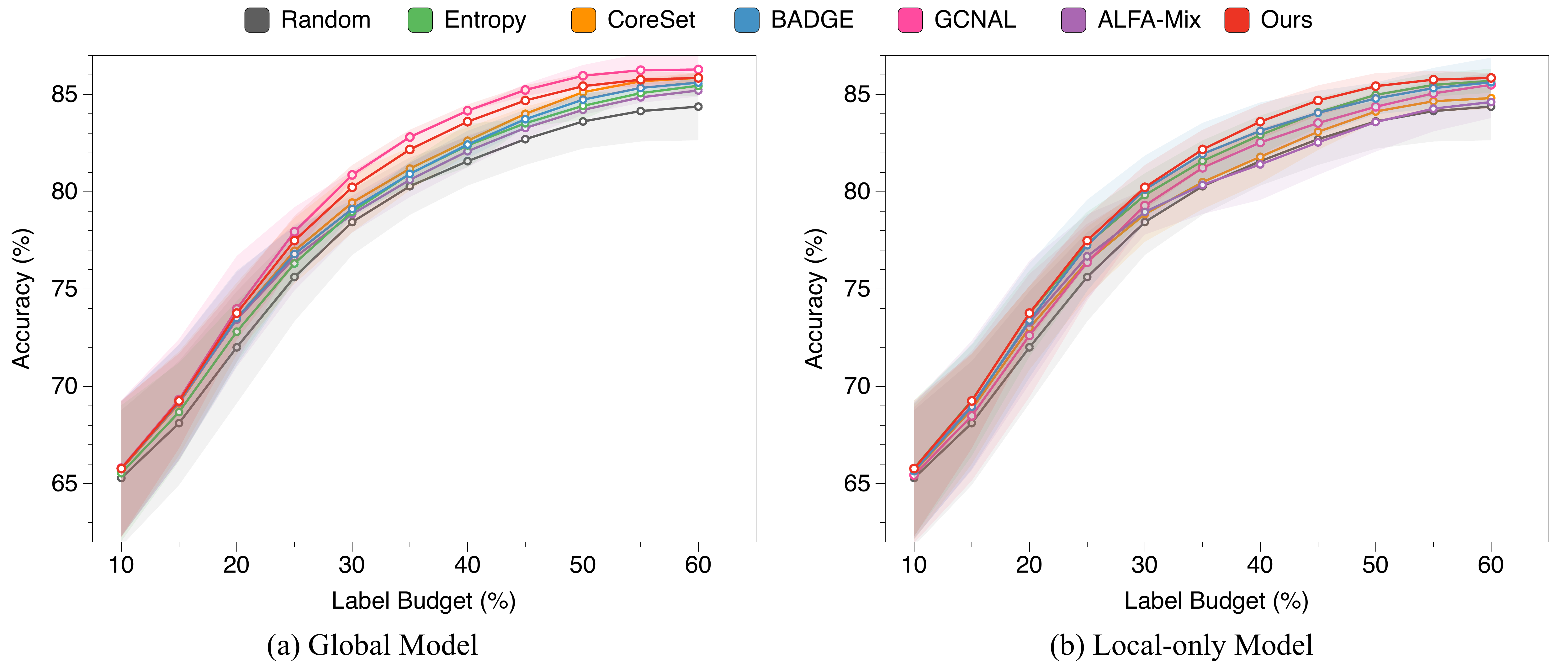}
    \caption{Test accuracy on SVHN, MobileNet, $\alpha=0.1$,  medium budget size\,(5\%), and random initialization.}
    \label{fig:app_svhn_mobile}
\end{figure*}

\vspace{-15pt}
\begin{figure*}[!h]
    \centering
    \includegraphics[width=0.8\linewidth]{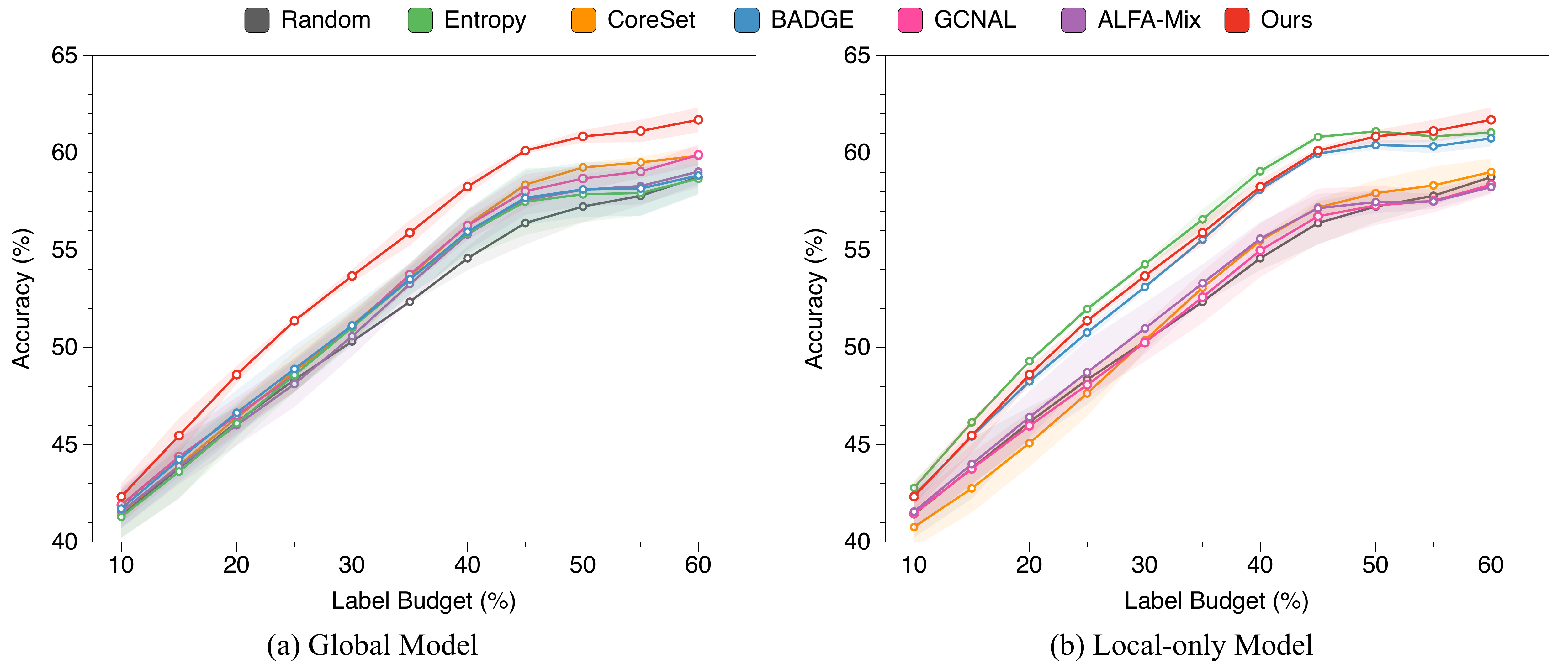}
    \caption{Test accuracy on CIFAR-10, ResNet-18, $\alpha=0.1$,  medium budget size\,(5\%), and random initialization.}
    \label{fig:app_cifar10_resnet}
\end{figure*}

\vspace{-15pt}
\begin{figure*}[!h]
    \centering
    \includegraphics[width=0.8\linewidth]{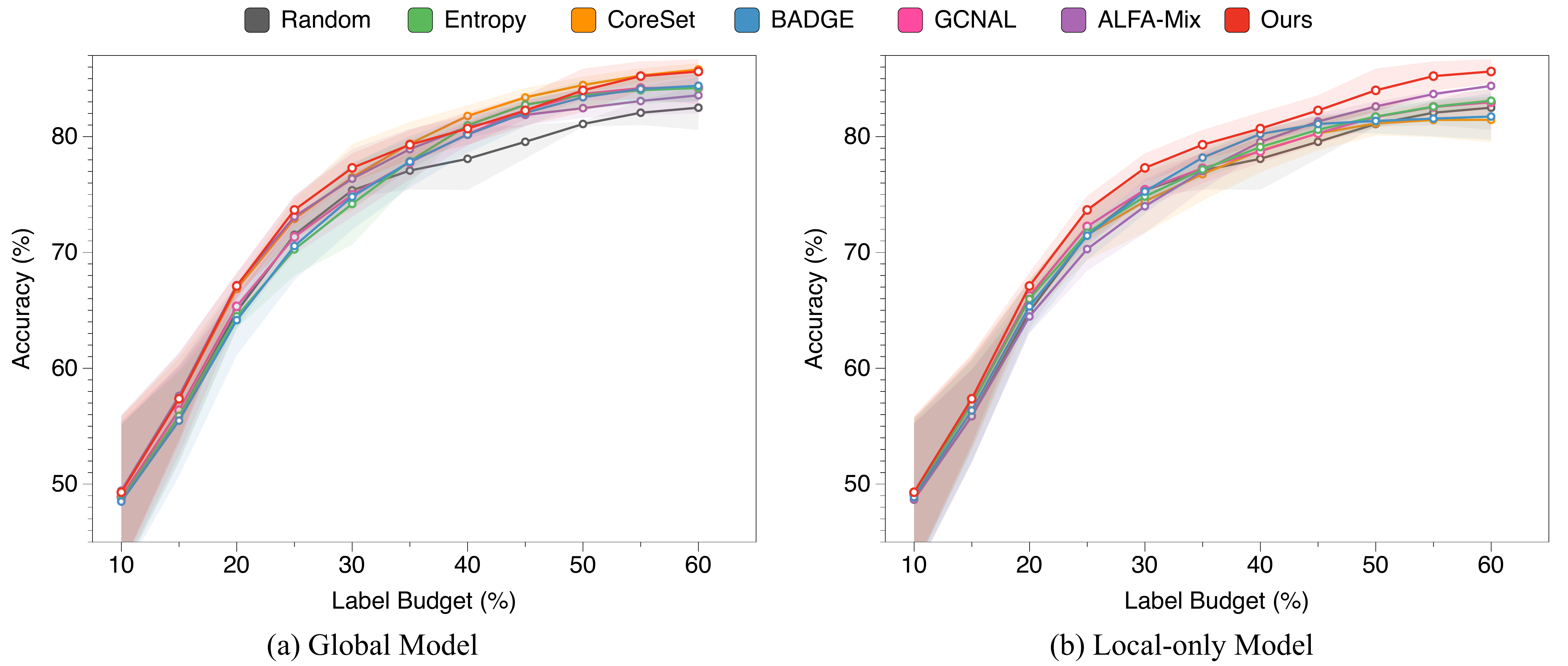}
    \caption{Test accuracy on SVHN, ResNet-18, $\alpha=0.1$,  medium budget size\,(5\%), and random initialization.}
    \label{fig:app_svhn_resnet}
\end{figure*}

\vspace{-15pt}
\begin{figure*}[!h]
    \centering
    \includegraphics[width=0.8\linewidth]{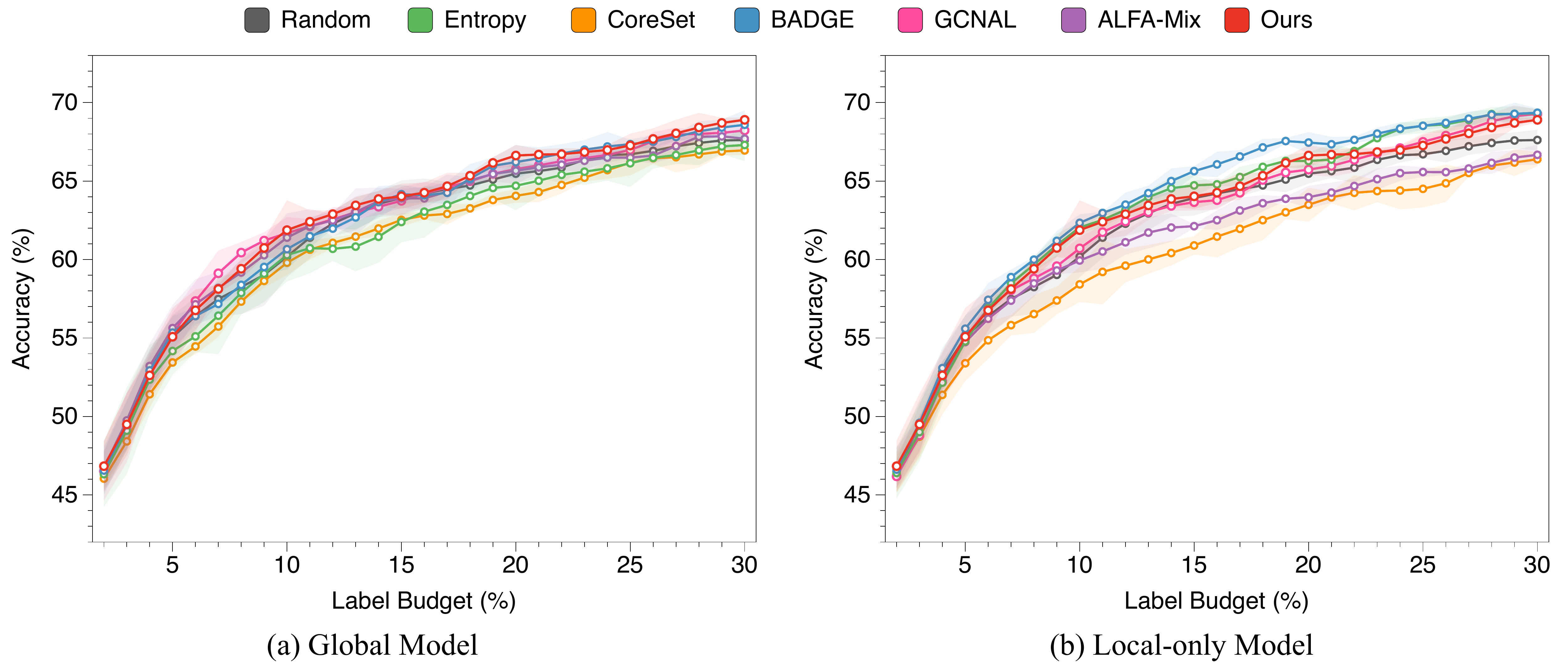}
    \caption{Test accuracy on CIFAR-10, four layers of CNN, $\alpha=0.1$,  small budget size\,(1\%), and random initialization.}
    \label{fig:app_cifar10_budget1}
\end{figure*}

\vspace{-15pt}
\begin{figure*}[!h]
    \centering
    \includegraphics[width=0.8\linewidth]{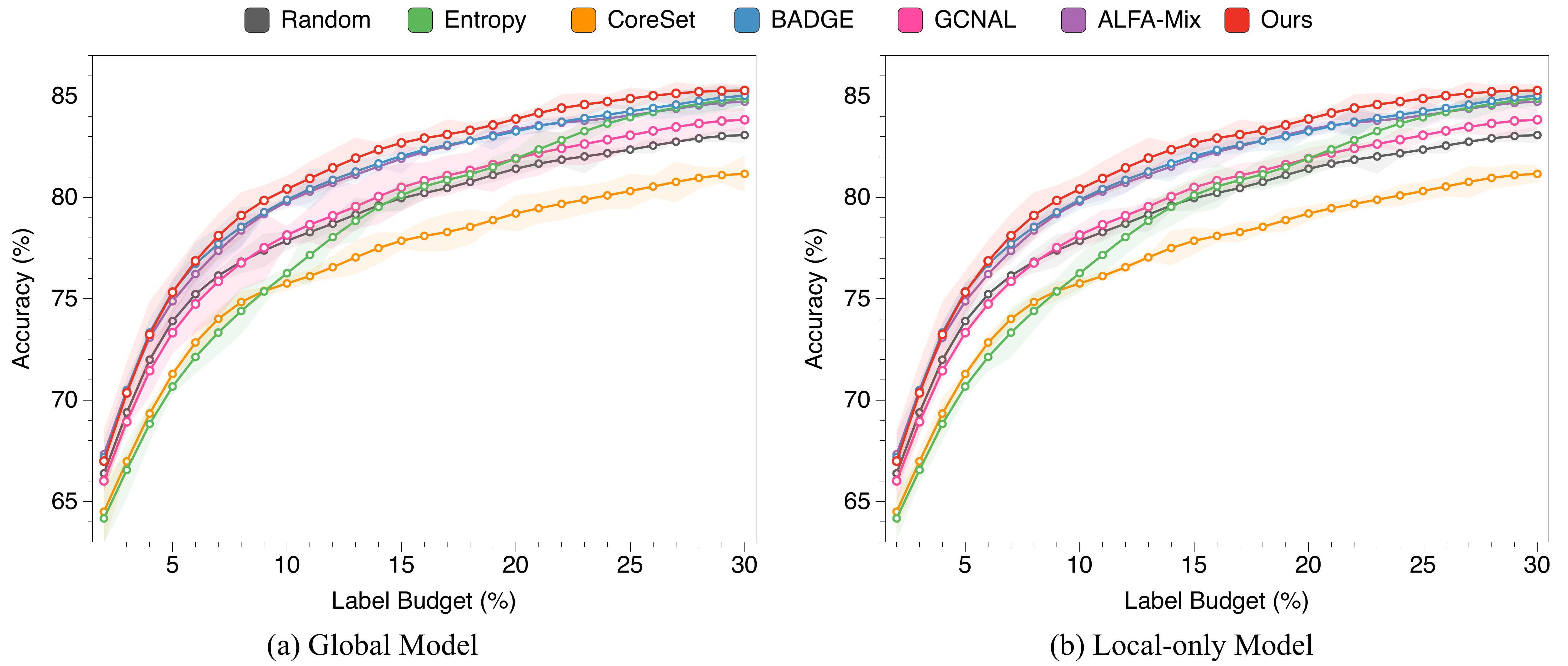}
    \caption{Test accuracy on SVHN, four layers of CNN, $\alpha=0.1$,  small budget size\,(1\%), and random initialization.}
    \label{fig:app_svhn_budget1}
\end{figure*}

\vspace{-15pt}
\begin{figure*}[!h]
    \centering
    \includegraphics[width=0.8\linewidth]{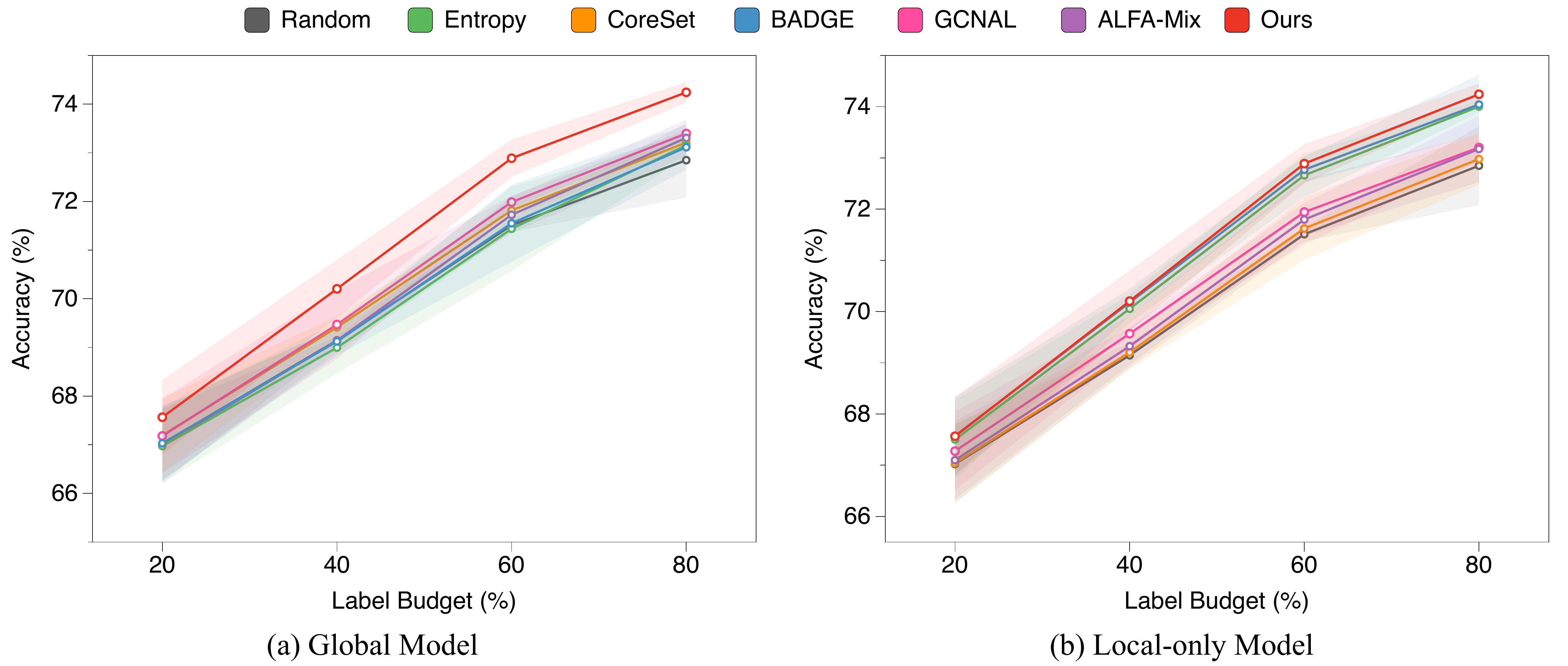}
    \caption{Test accuracy on CIFAR-10, four layers of CNN, $\alpha=0.1$,  large budget size\,(20\%), and random initialization.}
    \label{fig:app_cifar10_budget20}
\end{figure*}

\vspace{-15pt}
\begin{figure*}[!h]
    \centering
    \includegraphics[width=0.8\linewidth]{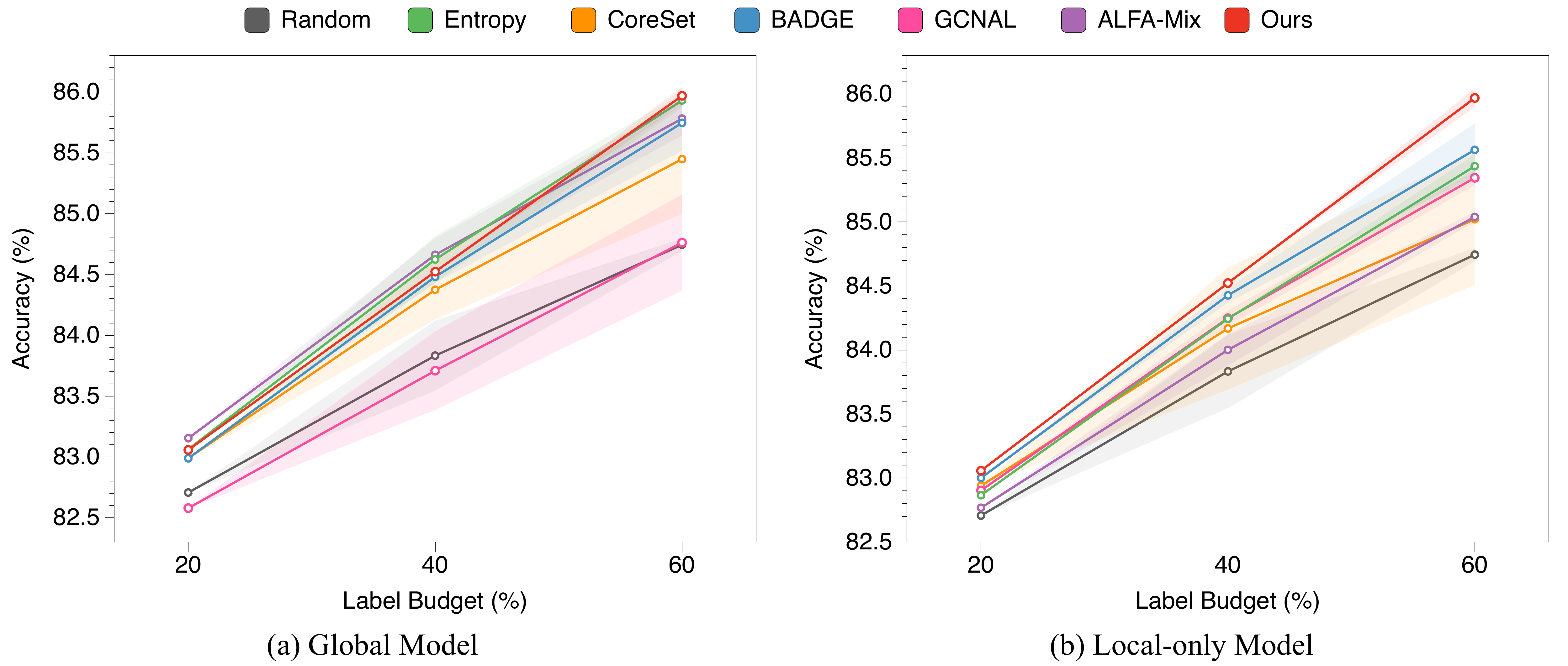}
    \caption{Test accuracy on SVHN, four layers of CNN, $\alpha=0.1$,  large budget size\,(20\%), and random initialization.}
    \label{fig:app_svhn_budget20}
\end{figure*}

\newpage
\vspace{-15pt}
\begin{figure*}[!h]
    \centering
    \includegraphics[width=0.8\linewidth]{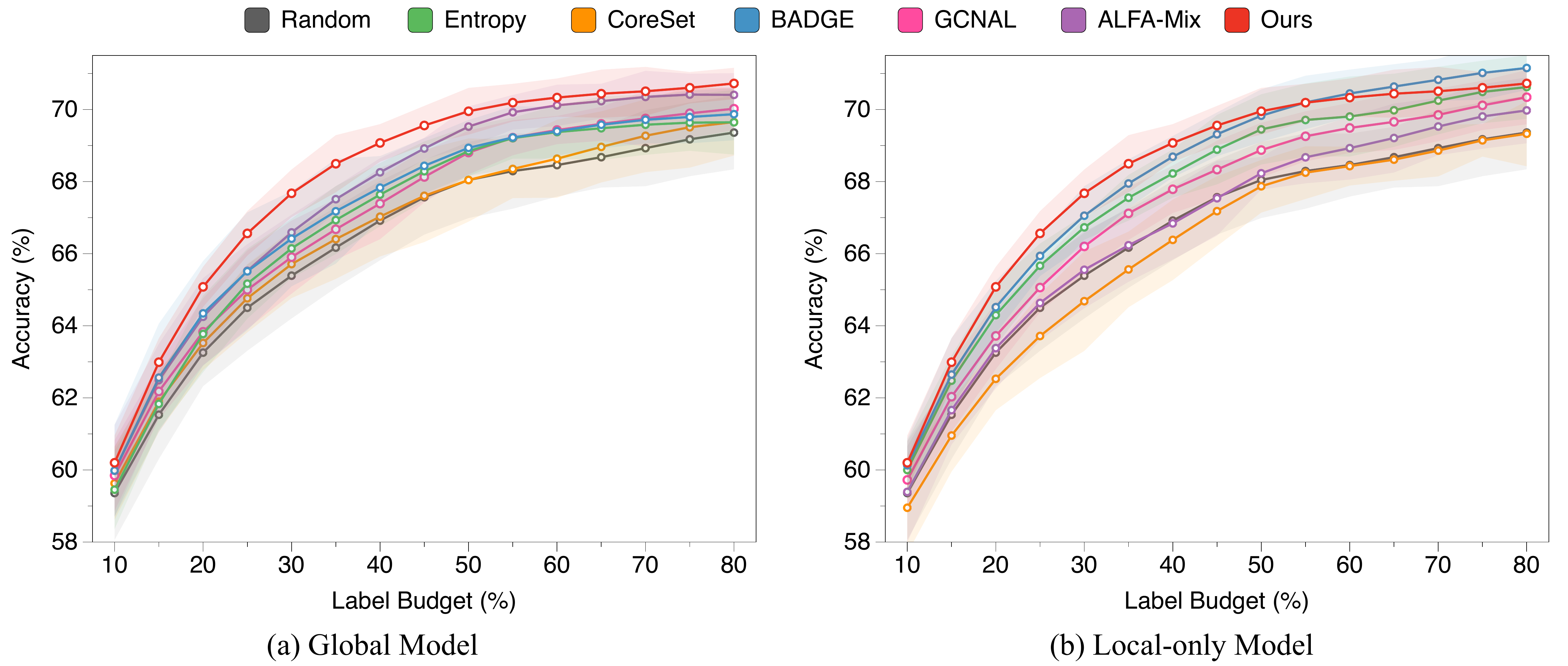}
    \caption{Test accuracy on CIFAR-10, four layers of CNN, $\alpha=0.1$,  medium budget size\,(5\%), and continue initialization.}
    \label{fig:app_cifar10_cont}
\end{figure*}

\vspace{-15pt}
\begin{figure*}[t]
    \centering
    \includegraphics[width=0.8\linewidth]{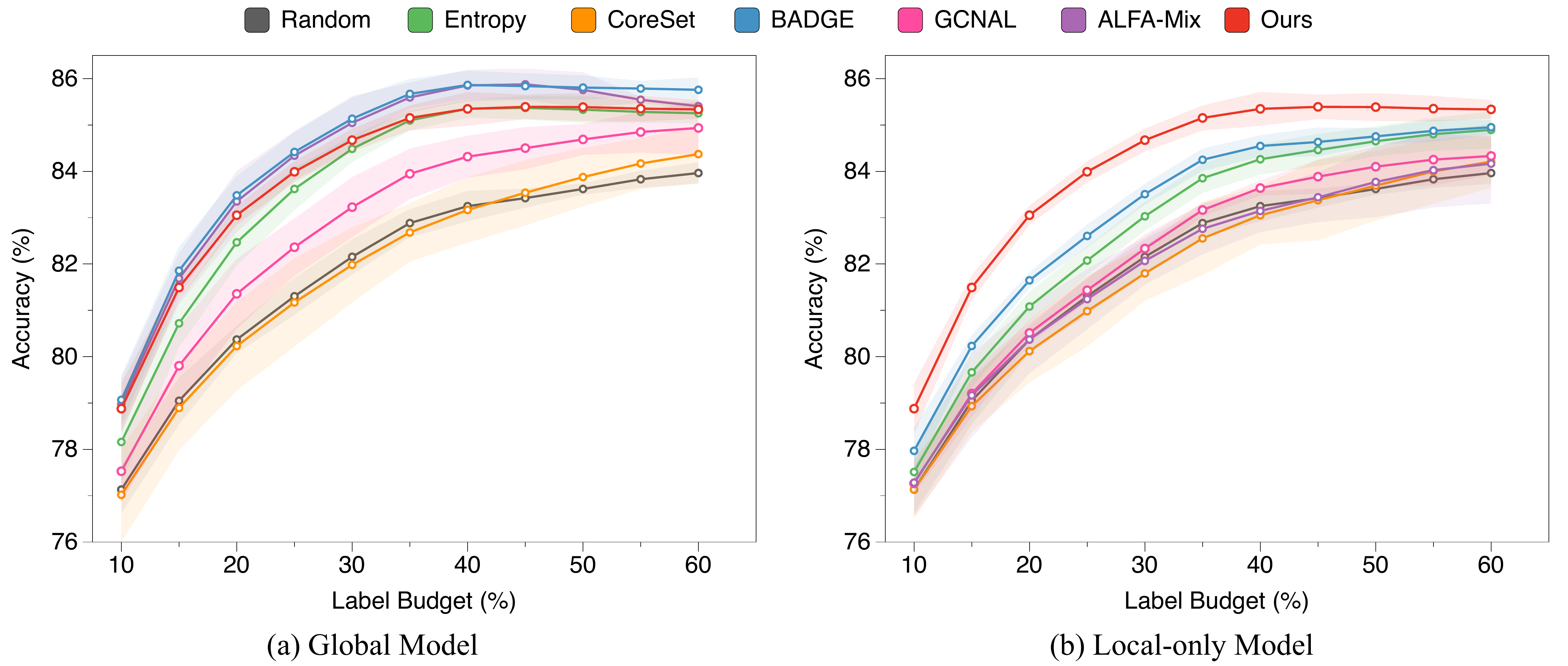}
    \caption{Test accuracy on SVHN, four layers of CNN, $\alpha=0.1$,  medium budget size\,(5\%), and continue initialization.}
    \label{fig:app_svhn_cont}
\end{figure*}

\end{document}